\def\year{2021}\relax
\documentclass[letterpaper]{article} 
\usepackage{aaai21}  
\usepackage{times}  
\usepackage{helvet} 
\usepackage{courier}  
\usepackage[hyphens]{url}  
\usepackage{graphicx} 
\usepackage{natbib}  
\usepackage{caption} 
\usepackage{xcolor}

\usepackage{amsmath,amsfonts,bm}

\newcommand{\Egauss}{\E}
\newcommand{\Qmc}{\mQ}

\newcommand{\UAPG}{\textsc{UAPG}\xspace}

\newcommand{\DBQPG}{\textsc{DBQPG}\xspace}
\newcommand{\PG}{\textsc{PG}\xspace}
\newcommand{\BQ}{\textsc{BQ}\xspace}
\newcommand{\TRPO}{\textsc{TRPO}\xspace}

\newcommand{\BAC}{\textsc{BAC}\xspace}
\newcommand{\GP}{\textsc{GP}\xspace}
\newcommand{\GPTD}{\textsc{GPTD}\xspace}
\newcommand{\NPG}{\textsc{NPG}\xspace}

\newcommand{\RL}{\textsc{RL}\xspace}
\newcommand{\MC}{\textsc{MC}\xspace}

\def\eqref#1{equation~\ref{#1}}

\def\1{\bm{1}}

\def\eps{{\epsilon}}

\def\vtheta{{\bm{\theta}}}

\def\vh{{\bm{h}}}

\def\vk{{\bm{k}}}

\def\vu{{\bm{u}}}
\def\vv{{\bm{v}}}

\def\mC{{\bm{C}}}

\def\mG{{\bm{G}}}

\def\mI{{\bm{I}}}

\def\mK{{\bm{K}}}
\def\mL{{\bm{L}}}

\def\mP{{\bm{P}}}
\def\mQ{{\bm{Q}}}
\def\mR{{\bm{R}}}

\def\mU{{\bm{U}}}

\def\mW{{\bm{W}}}

\def\mLambda{{\bm{\Lambda}}}

\DeclareMathAlphabet{\mathsfit}{\encodingdefault}{\sfdefault}{m}{sl}
\SetMathAlphabet{\mathsfit}{bold}{\encodingdefault}{\sfdefault}{bx}{n}

\newcommand{\E}{\mathbb{E}}

\newcommand{\R}{\mathbb{R}}

\newcommand{\Cov}{\mathrm{Cov}}

\usepackage{url}
\usepackage{natbib}
\usepackage{subfigure}
\usepackage{amssymb}

\usepackage{adjustbox,xspace}
\usepackage{algorithm}
\usepackage{algpseudocode,amsmath,mathtools}
\usepackage[normalem]{ulem}
\usepackage[]{multirow}

\renewcommand{\cite}[1]{\citep{#1}}

\title{Deep Bayesian Quadrature Policy Optimization}

\author {
    Akella Ravi Tej\textsuperscript{\rm $\dagger$} \quad
    Kamyar Azizzadenesheli\textsuperscript{\rm $\dagger$} \quad
    Mohammad Ghavamzadeh\textsuperscript{\rm $\ddagger$} \\
    Anima Anandkumar\textsuperscript{\rm $\mathsection$} \quad
    Yisong Yue\textsuperscript{\rm $\mathsection$}\\
}
\affiliations {
    \textsuperscript{\rm $\dagger$}Purdue University \quad \textsuperscript{\rm $\ddagger$}Google Research \quad \textsuperscript{\rm $\mathsection$}Caltech \\
    \texttt{\{rakella,kamyar\}@purdue.edu, ghavamza@google.com, \{anima,yyue\}@caltech.edu}
}

\begin{document}

\maketitle

\begin{abstract}
We study the problem of obtaining accurate policy gradient estimates using a finite number of samples. Monte-Carlo methods have been the default choice for policy gradient estimation, despite suffering from high variance in the gradient estimates. On the other hand, more sample efficient alternatives like Bayesian quadrature methods have received little attention due to their high computational complexity.
In this work, we propose deep Bayesian quadrature policy gradient (\DBQPG), a computationally efficient high-dimensional generalization of Bayesian quadrature, for policy gradient estimation. We show that \DBQPG can substitute Monte-Carlo estimation in policy gradient methods, and demonstrate its effectiveness on a set of continuous control benchmarks. In comparison to Monte-Carlo estimation, \DBQPG provides (i) more accurate gradient estimates with a significantly lower variance, (ii) a consistent improvement in the sample complexity and average return for several deep policy gradient algorithms, and, (iii) the uncertainty in gradient estimation that can be incorporated to further improve the performance.
\end{abstract}

\section{Introduction}
Policy gradient (\PG) is a reinforcement learning (\RL) approach that directly optimizes the agent's policies by operating on the gradient of their expected return \citep{Sutton_1999,baxter2000direct}. The use of deep neural networks for the policy class has recently demonstrated a series of success for \PG methods~\citep{lillicrap2015continuous,TRPO} on high-dimensional continuous control benchmarks, such as MuJoCo~\citep{todorov2012mujoco}. However, the derivation and analysis of the aforementioned methods mainly rely on access to the expected return and its true gradient. In general, \RL agents do not have access to the true gradient of the expected return, i.e., the gradient of integration over returns; instead, they have access to its empirical estimate from sampled trajectories. Monte-Carlo (\MC) sampling~\citep{metropolis1949monte} is a widely used point estimation method for approximating this integration~\citep{Williams_1992}. 
%
%
However, \MC estimation returns high variance gradient estimates that are undesirably inaccurate, imposing a high sample complexity requirement for \PG methods~\citep{Rubinstein69,accuracy_variance_PG}.

An alternate approach to approximate integrals in probabilistic numerics is Bayesian Quadrature (\BQ)~\citep{O'Hagan1991}. Under mild regularity assumptions, \BQ offers impressive empirical advances and strictly faster convergence rates~\citep{kanagawa2016convergence}. Typically, the integrand in \BQ is modeled using a Gaussian process (\GP), such that the linearity of the integral operator provides a Gaussian posterior over the integral. In addition to a point estimate of the integral, \BQ also quantifies the uncertainty in its estimation, a missing piece in \MC methods. 
%
%
%
%
%

In \RL, the \BQ machinery can be used to obtain a Gaussian approximation of the \PG integral, by placing a \GP prior over the action-value function. However, estimating the moments of this Gaussian approximation, i.e., the \PG mean and covariance, requires the integration of \GP's kernel function, which, in general, does not have an analytical solution. Interestingly, \citet{BAC_1} showed that the \PG integral can be solved analytically when the \GP kernel is an additive combination of an arbitrary state kernel and a fixed Fisher kernel.
%
While the authors demonstrate a superior performance of \BQ over \MC using a small policy network on simple environments,
their approach does not scale to large non-linear policies ($> 1000$ trainable parameters) and high-dimensional domains.

\textbf{Contribution 1:} We propose deep Bayesian quadrature policy gradient (\DBQPG), a \BQ-based \PG framework that extends \citet{BAC_1} to high-dimensional settings, thus placing it in the context of contemporary deep \PG algorithms. The proposed framework uses a \GP to implicitly model the action-value function, and without explicitly constructing the action-value function, returns a Gaussian approximation of the policy gradient, represented by a mean vector (gradient estimate) and covariance (gradient uncertainty). Consequently, this framework can be used with an explicit critic network (different from implicit \GP critic\footnote{The \GP critic in \BQ is implicit because it is used to solve \PG integral analytically rather than explicitly predicting the $Q$-values.}) to leverage the orthogonal benefits of actor-critic frameworks \cite{GAE}.


%
%

The statistical efficiency of \BQ, relative to \MC estimation, depends on the compatibility between the \GP kernel and the MDP's action-value function. To make \DBQPG robust to a diverse range of target MDPs, we choose a base kernel capable of universal approximation (e.g. RBF kernel), and augment its expressive power using deep kernel learning \cite{dkl}. Empirically, \DBQPG estimates gradients that are both much closer to the true gradient, and with much lower variance, when compared to MC estimation. Moreover, \DBQPG is a linear-time program that leverages the recent advances in structured kernel interpolation~\citep{kissgp} for computational efficiency and GPU acceleration for fast kernel learning~\citep{gpytorch}. Therefore, \DBQPG can favorably substitute \MC estimation subroutine in a variety of \PG algorithms. Specifically, we show that replacing the \MC estimation subroutine with \DBQPG provides a significant improvement in the sample complexity and average return for vanilla \PG \citep{Sutton_1999}, natural policy gradient (\NPG) \citep{Kakade_2001_NPG} and trust region policy optimization (\TRPO) \citep{TRPO} algorithms on $7$ diverse MuJoCo domains.

\textbf{Contribution 2:} We propose uncertainty aware \PG (\UAPG), a novel policy gradient method that utilizes the uncertainty in the gradient estimation for computing reliable policy updates. A majority of \PG algorithms~\citep{Kakade_2001_NPG,TRPO} are derived assuming access to true gradients and therefore do not account the stochasticity in gradient estimation due to finite sample size. 
However, one can obtain more reliable policy updates by lowering the stepsize along the directions of high gradient uncertainty. \UAPG captures this intuition by utilizing \DBQPG's uncertainty to bring different components of a stochastic gradient estimate to the same scale. \UAPG does this by normalizing the step-size of the gradient components by their respective uncertainties, returning a new gradient estimate with uniform uncertainty, i.e., with new gradient covariance as the identity matrix.
\UAPG further provides a superior sample complexity and average return in comparison to \DBQPG.
%
%
%
Our implementation of DBQPG and UAPG methods is available online: {\color{blue}{\url{https://github.com/Akella17/Deep-Bayesian-Quadrature-Policy-Optimization}}}.
%
%

\section{Preliminaries}
Consider a Markov decision process (MDP) $\langle\mathcal{S}, \mathcal{A}, P, r, \rho_0, \gamma\rangle$, where $\mathcal{S}$ is the state-space, $\mathcal{A}$ is the action-space, $P : \mathcal{S} \times \mathcal{A} \rightarrow \Delta_\mathcal{S}$ is the transition kernel that maps each state-action pair to a distribution over the states $\Delta_\mathcal{S}$, $r: \mathcal{S} \times \mathcal{A} \rightarrow \R$ is the reward kernel, $\rho_0: \Delta_\mathcal{S}$ is the initial state distribution, and $\gamma \in [0,1)$ is the discount factor. We denote by $\pi_{\vtheta}: \mathcal{S} \rightarrow \Delta_\mathcal{A}$ a stochastic parameterized policy with parameters $\theta\in\Theta$. The MDP controlled by the policy $\pi_{\vtheta}$ induces a Markov chain over state-action pairs $z = (s,a) \in \mathcal{Z} = \mathcal{S} \times \mathcal{A}$, with an initial density $\rho^{\pi_{\vtheta}}_0(z_0) = \pi_{\vtheta}(a_0|s_0) \rho_0(s_0)$ and a transition probability distribution $P^{\pi_{\vtheta}}(z_t|z_{t-1}) = \pi_{\vtheta}(a_t|s_t) P(s_t|z_{t-1})$.
%
We use the standard definitions for action-value function $Q_{\pi_{\vtheta}}$ and expected return $J(\vtheta)$ under $\pi_{\vtheta}$:

\begin{align}
\label{eqn:q_func_definitions}
Q_{\pi_{\vtheta}} (z_t) = \E\Bigg[ & \sum_{\tau=0}^{\infty} \gamma^\tau r(z_{t+\tau}) ~\Big|~ z_{t+\tau+1} \sim P^{\pi_{\vtheta}}(.|z_{t+\tau})\Bigg],\nonumber
\\
&J(\vtheta) = \E_{z_0\sim\rho^{\pi_{\vtheta}}_0}\left[Q_{\pi_{\vtheta}}(z_0)\right].
\end{align}
However, the gradient of $J(\vtheta)$ with respect to policy parameters $\vtheta$ cannot be directly computed from this formulation.
The policy gradient theorem~\citep{Sutton_1999,konda_2000} provides an analytical expression for the gradient of the expected return $J(\vtheta)$, as:
%
\begin{align}
\label{eqn:policy_gradient_theorem}
\displaystyle \nabla_{\vtheta} J(\vtheta) &= \int_{\mathcal{Z}} dz \rho^{\pi_\vtheta}(z) \vu(z) Q_{\pi_\vtheta}(z)\nonumber
\\
&= \E_{z \sim \rho^{\pi_\vtheta}}\big[\vu(z) Q_{\pi_\vtheta}(z)\big],
\end{align}
where $\vu(z) = \nabla_{\vtheta} \log \pi_{\vtheta} (a|s)$ is the score function and $\rho^{\pi_{\vtheta}}$ is the discounted state-action visitation frequency defined as:
%
\begin{align}
\label{eqn:state_visitation}
\rho^{\pi_{\vtheta}}(z) &= \sum_{t=0}^{\infty} \gamma^t P^{\pi_{\vtheta}}_t(z), \text{where}
\\
P^{\pi_{\vtheta}}_t(z_t) = \int_{\mathcal{Z}^t} dz_0 ... & dz_{t-1} P^{\pi_{\vtheta}}_0(z_0) \prod_{\tau=1}^t P^{\pi_{\vtheta}}(z_\tau|z_{\tau-1})\nonumber.
\end{align}
%
%

\section{Policy Gradient Evaluation}
\label{sec:Policy_Gradient_Evaluation}
The exact evaluation of the \PG integral in Eq.~\ref{eqn:policy_gradient_theorem} is often intractable for MDPs with a large (or continuous) state or action space. We discuss two prominent approaches to approximate this integral using a finite set of samples $\{z_i\}_{i=1}^{n} \sim \rho^{\pi_\vtheta}$: (i) Monte-Carlo method and (ii) Bayesian Quadrature.

\textbf{Monte-Carlo (\MC)} method\footnote{Here, \MC refers to the Monte-Carlo numerical integration method, and not the TD(1) (a.k.a Monte-Carlo) $Q$-estimates.} approximates the integral in Eq.~\ref{eqn:policy_gradient_theorem} by the finite sum:
\begin{equation}
\label{eqn:Monte-Carlo_PG}
\displaystyle \mL_{\vtheta}^{MC} = \frac{1}{n}\sum_{i=1}^{n} Q_{\pi_\vtheta}(z_i) \vu(z_i) = \frac{1}{n} \mU \Qmc,
\end{equation}
where $\vu(z)$ is a $|\Theta|$ dimensional vector ($|\Theta|$ is the number of policy parameters). For conciseness, the function evaluations at sample locations $\{z_i\}_{i=1}^{n} \sim \rho^{\pi_\vtheta}$ are grouped into $\mU = [ \vu(z_1), ..., \vu(z_n)]$, a $|\Theta| \times n$ dimensional matrix, and $\Qmc = [Q_{\pi_\vtheta}(z_1),...,Q_{\pi_\vtheta}(z_n)]$\footnote{$Q_{\pi_\vtheta}(z)$ is computed using TD(1) (a.k.a Monte-Carlo) method or an explicit critic module (different from \BQ's implicit \GP critic)}, an $n$ dimensional vector.
\MC method returns the gradient mean evaluated at sample locations, which, according to the central limit theorem (CLT), is an unbiased estimate of the true gradient. However, CLT also suggests that the \MC estimates suffer from a slow convergence rate ($n^{-1/2}$) and high variance, necessitating a large sample size $n$ for reliable \PG estimation \citep{accuracy_variance_PG}.
Yet, \MC methods are more computationally efficient than their sample-efficient alternatives (e.g. \BQ), making them ubiquitous in \PG algorithms.

\textbf{Bayesian quadrature (\BQ)} \citep{O'Hagan1991} is an approach from \textit{probabilistic numerics} \cite{HenOsbGir15} that converts numerical integration into a Bayesian inference problem. The first step in \BQ is to formulate a prior stochastic model over the integrand. This is done by placing a Gaussian process (\GP) prior on the $Q_{\pi_\vtheta}$ function, i.e., a mean zero \GP $\Egauss\left[ Q_{\pi_\vtheta}(z) \right] = 0$, with a covariance function $k(z_p,z_q) = \Cov [Q_{\pi_\vtheta}(z_p),Q_{\pi_\vtheta}(z_q)]$, and observation noise $\sigma$.
Next, the \GP prior on  $Q_{\pi_\vtheta}$ is conditioned (Bayes rule) on the sampled data $\mathcal{D} = \{z_i\}_{i=1}^{n}$ to obtain the posterior moments
$\Egauss\left[ Q_{\pi_\vtheta}(z)| \mathcal{D} \right]$ and $\mathcal{C}_Q(z_1,z_2) = \Cov\left[ Q_{\pi_\vtheta}(z_1), Q_{\pi_\vtheta}(z_2)| \mathcal{D} \right]$.
Since the transformation from $Q_{\pi_\vtheta}(z)$ to $\nabla_{\vtheta} J(\vtheta)$ happens through a linear integral operator (in Eq.~\ref{eqn:policy_gradient_theorem}), $\nabla_{\vtheta} J(\vtheta)$ also follows a Gaussian distribution:
\begin{align}
\label{eqn:BayesianQuadratureIntegral}
    \displaystyle \mL_{\vtheta}^{BQ} &= \Egauss \left[ \nabla_{\vtheta} J(\vtheta) | \mathcal{D} \right] 
    =\displaystyle \int_{\mathcal{Z}} dz \rho^{\pi_\vtheta}(z) \vu(z) \Egauss \left[ Q_{\pi_\vtheta}(z) | \mathcal{D} \right] \nonumber\\
    \displaystyle \mC_{\vtheta}^{BQ} &= \Cov [ \nabla_{\vtheta} J(\vtheta) | \mathcal{D}] \\
    =& \displaystyle \int_{\mathcal{Z}^2} dz_1 dz_2 \rho^{\pi_\vtheta}(z_1) \rho^{\pi_\vtheta}(z_2) \vu(z_1) \mathcal{C}_Q(z_1,z_2) \vu(z_2)^\top,\nonumber
\end{align}
where the mean vector $\mL_{\vtheta}^{BQ}$ is the \PG estimate and the covariance $\mC_{\vtheta}^{BQ}$ is its uncertainty estimation.
While the integrals in Eq.~\ref{eqn:BayesianQuadratureIntegral} are still intractable for an arbitrary \GP kernel $k$,
they have closed-form solutions when $k$ is the additive composition of a state kernel $k_s$ (arbitrary) and the Fisher kernel $k_f$ (indispensable) \citep{BAC_1}:
\begin{align}
\label{eqn:kernel_definitions}
    k(z_1,z_2) &= c_1 k_s(s_1,s_2) + c_2 k_f(z_1,z_2),\nonumber\\
    ~~ \textit{with}~ k_f & (z_1,z_2) = \vu(z_1)^\top \mG^{-1} \vu(z_2),
\end{align}
where $c_1, c_2$ are hyperparameters\footnote{$c_1, c_2$ are redundant hyperparameters that are simply introduced to offer a better explanation of \BQ-\PG.} and $\mG$ is the Fisher information matrix of $\pi_\vtheta$. Using the matrices,
\begin{align}
\label{eqn:kernel_equations}
\mK_f = \mU^\top \mG^{-1} \mU, \quad & \mK = c_1 \mK_s + c_2 \mK_f,\nonumber\\ 
\mG = \E_{z \sim \rho^{\pi_\vtheta}} [\vu(z) & \vu(z)^\top] \approx \frac{1}{n} \mU\mU^\top,
\end{align}
the \PG mean $\mL_{\vtheta}^{BQ}$ and covariance $\mC_{\vtheta}^{BQ}$ can be computed analytically (proof is deferred to the supplement Sec.~\ref{appendix_bq_pg_closedform}),
\begin{align}\label{eq:GP_grad}
    \displaystyle \mL_{\vtheta}^{BQ} &= \displaystyle c_2 \mU (\mK +\sigma^2\mI)^{-1}\Qmc,\nonumber\\
    \displaystyle \mC_{\vtheta}^{BQ} &= c_2 \mG - c_2^2 \mU \left( \mK + \sigma^2 \mI \right)^{-1} \mU^\top.
\end{align}
Here, $\Qmc$ is same as in the \MC method.
Note that removing the Fisher kernel $k_f$ (setting $c_2 = 0$ in Eq.~\ref{eq:GP_grad}) causes $\mL_{\vtheta}^{BQ} = 0$ and $\mC_{\vtheta}^{BQ} = 0$. Further, \citet{BAC_1} showed that \BQ-\PG $\mL_{\vtheta}^{BQ}$ reduces to \MC-\PG $\mL_{\vtheta}^{MC}$ when the state kernel $k_s$ is removed ($c_1 = 0$).\newline
\textbf{To summarize}, (i) the presence of Fisher kernel ($c_2 \neq 0$) is essential for obtaining an analytical solution for \BQ-\PG, and (ii) with the Fisher kernel fixed, the choice of state kernel alone determines the convergence rate of \BQ-\PG relative to the \MC baseline (equivalently \BQ-\PG with $c_1 = 0$).

\section{Deep Bayesian Quadrature Policy Gradient}
\label{sec:DBQPG_algorithm}
Here, we introduce the steps needed for obtaining a practical \BQ-\PG method, that (i) is more sample-efficient than \MC baseline, and (ii) easily scales to  high-dimensional settings.

\textbf{Kernel Selection:}
In the previous section, it was highlighted that the flexibility of kernel selection is limited to the choice of the state kernel $k_s$. To understand the role of $k_s$, we first highlight the implication of the kernel composition proposed in \citet{BAC_1}. Specifically, the additive composition of the state kernel $k_s$ and the Fisher kernel $k_f$ implicitly divides the $Q_{\pi_\vtheta}$ function into state-value function and advantage function, separately modeled by $k_s$ and $k_f$ respectively (supplement, Sec.~\ref{appendix_state_value_advantage}). Thus, removing the Fisher kernel $k_f$ ($c_2 = 0$) results in a non-existent advantage function, which explains $\mL_{\vtheta}^{BQ} = 0$ and $\mC_{\vtheta}^{BQ} = 0$. More interestingly, removing the state kernel $k_s$ ($c_1 = 0$) results in a non-existent state-value function, which also reduces $\mL_{\vtheta}^{BQ}$ to $\mL_{\vtheta}^{MC}$. In other words, \MC-\PG is a limiting case of \BQ-\PG where the state-value function is suppressed to $0$ (more details in the supplement Sec.~\ref{appendix:MCPG_degenerateofBQ}).

Thus, \BQ-\PG can offer more accurate gradient estimates than \MC-\PG, along with well-calibrated uncertainty estimates, when the state kernel $k_s$ is a better prior than the trivial $k_s = 0$, for the MDP's state-value function. To make the choice of $k_s$ robust to a diverse range of target MDPs, we suggest (i) a base kernel capable of universal approximation (e.g. RBF kernel), followed by (ii) increasing the base kernel's expressive power by using a deep neural network (DNN) to transform its inputs \cite{dkl}. Deep kernels combine the non-parametric flexibility of \GP{}s with the structural properties of NNs to obtain more expressive kernel functions when compared to their base kernels. The kernel parameters $\bm{\phi}$ (DNN parameters + base kernel hyperparameters) are tuned using the gradient of \GP's log marginal likelihood $J_{GP}$~\citep{rasmussen_05_book_gpml},
\begin{align}
\label{eqn:GP_nll}
    & J_{GP}(\bm{\phi}|\mathcal{D}) \propto \log |\mK| - \Qmc^\top \mK^{-1} \Qmc,\\
    \nabla_{\bm{\phi}} J_{GP} &= \Qmc^\top \mK^{-1} (\nabla_{\bm{\phi}}\mK) \mK^{-1} \Qmc + Tr(\mK^{-1} \nabla_{\bm{\phi}}\mK).\nonumber
\end{align}

\begin{figure*}[!ht]
	\centering
    \begin{subfigure}
  	\centering
  	\includegraphics[scale=0.53]{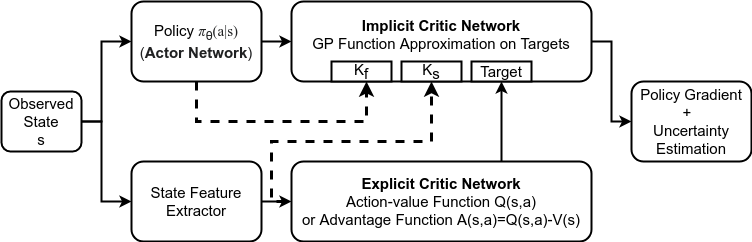}
    \end{subfigure}
    \hspace{1.5em}
    \begin{subfigure}
  	\centering
  	\includegraphics[scale=0.43]{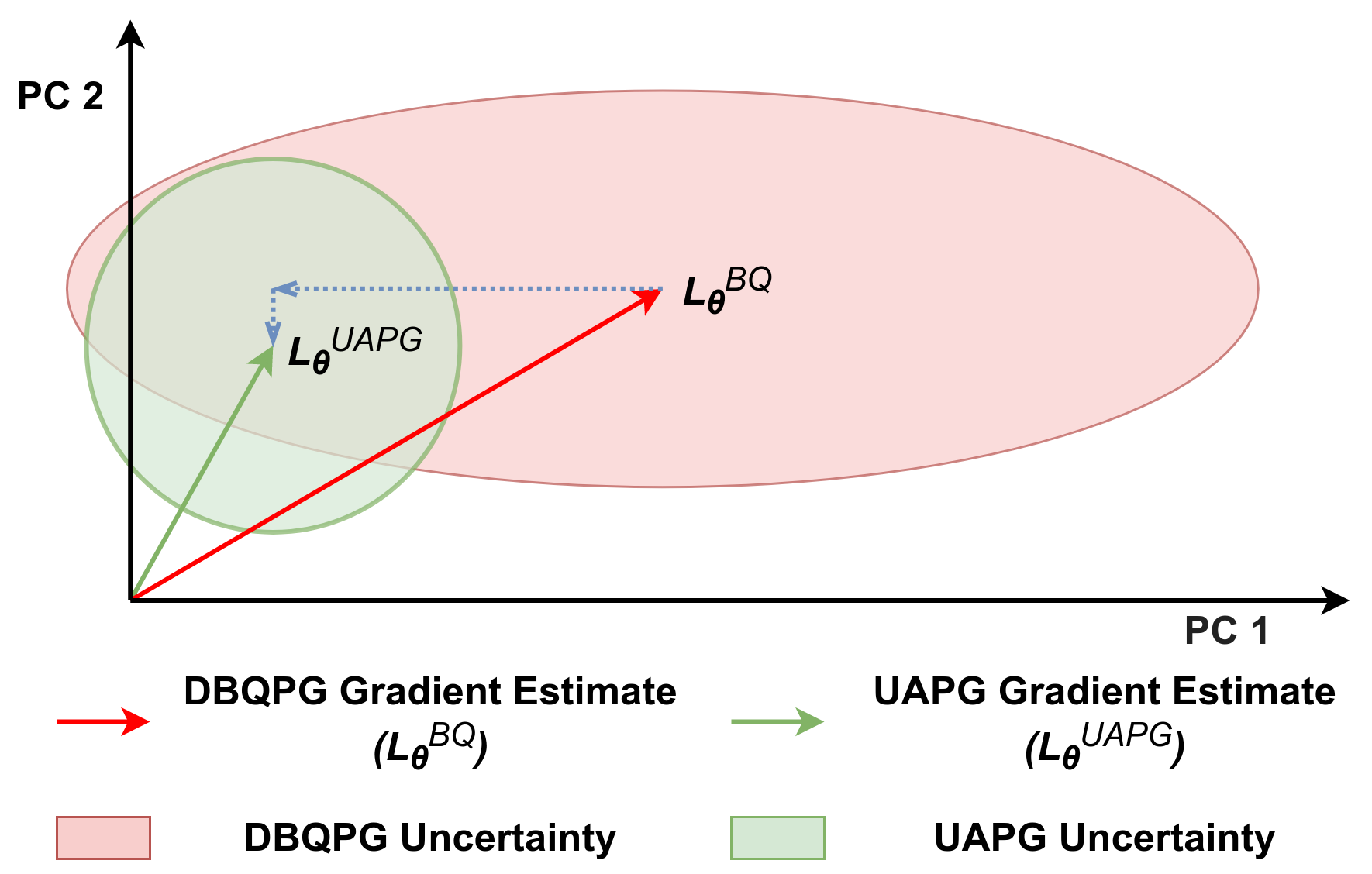}
    \end{subfigure}
    \caption{(Left) Overview of the \DBQPG method, (Right) Illustration of \DBQPG and \UAPG updates along the first two principal components (PCs) of the gradient covariance matrix.}
  \label{fig:DBQPG_UAPG_diagram}
\end{figure*}

\textbf{Scaling \BQ to large sample sizes $n$:}
The complexity of estimating $\mL_{\vtheta}^{BQ}$ (Eq.~\ref{eq:GP_grad}) is largely influenced by the matrix-inversion operation $(\mK +\sigma^2\mI)^{-1}$, whose exact computation scales with an $\mathcal{O}(n^3)$ time and $\mathcal{O}(n^2)$ space complexity. Our first step is to shift the focus from an expensive matrix-inversion operation to an approximate inverse matrix-vector multiplication (i-MVM). In particular, we use the conjugate gradient (CG) method to compute the i-MVM $(\mK +\sigma^2\mI)^{-1} \vv$ within machine precision using $p \ll n$ iterations of MVM operations $\mK \vv' = c_1 \mK_s \vv' + c_2 \mK_f \vv'$.
%
%

While the CG method nearly reduces the time complexity by an order of $n$, the MVM operation with a dense matrix still incurs a prohibitive $\mathcal{O}(n^2)$ computational cost. Fortunately, the matrix factorization of $\mK_f$ (Eq.~\ref{eqn:kernel_equations}) allows for computing the Fisher kernel MVM $\mK_f \vv$ in linear-time through automatic differentiation, without the explicit creation or storage of the $\mK_f$ matrix (more details in Sec.~\ref{sec:appendix_Fisher_mvm} of the supplement).
On the other hand, since the choice of $k_s$ is arbitrary, we deploy \textit{structured kernel interpolation (SKI)} \citep{kissgp}, a general kernel sparsification strategy to efficiently compute the state kernel MVM $\mK_s \vv$.
SKI uses a set of $m \leq n$ ``inducing points'' $\{\hat{s}_i\}_{i=1}^{m}$ to approximate $\mK_s$ with a rank $m$ matrix $\hat{\mK}_s = \mW \mK_s^m \mW^\top$, where $\mK_s^m$ is an $m \times m$ Gram matrix with entries ${\mK_s^m}_{(p,q)} = k_s(\hat{s}_p, \hat{s}_q)$, and $\mW$ is an $n\times m$ interpolation matrix whose entries depend on the relative placement of sample points $\{s_i\}_{i=1}^{n}$ and inducing points $\{\hat{s}_i\}_{i=1}^{m}$. In practice, a sparse $\mW$ matrix that follows local cubic interpolation (only $4$ non-zero entries per row) provides a good approximation $\hat{\mK}_s$, and more importantly, offers $\hat{\mK}_s \vv$ MVM in $\mathcal{O}(n+m^2)$ time and storage complexity.

Further, the SKI framework also provides the flexibility to select the inducing point locations for exploiting the structure of specialized \GP kernels. For instance, 
one-dimensional stationary kernels, i.e., $k_s(x,y) = k_s(x-y)$, can additionally leverage the Toeplitz method \citep{toeplitz_turner:2010} by picking evenly-spaced inducing points. Since these matrices are constant along the diagonal, i.e. ${\mK_s}_{(x,y)} = {\mK_s}_{(x+1,y+1)}$, the Toeplitz method utilizes fast Fourier transform to attain an $\mathcal{O}(n + m\log m)$ time and $\mathcal{O}(n + m)$ storage for the MVM operation. Further, Toeplitz methods can be extended to multiple dimensions by assuming that the kernel decomposes as a sum of one-dimensional stationary kernels along each of the input dimensions. Compared to conventional inducing point methods that operate with $m \ll n$ inducing points, choosing a base kernel that conforms with the Toeplitz method enables the realization of larger $m$ values, thereby providing a more accurate approximation of $\mK_s$.

\textbf{Practical \DBQPG Algorithm:}
The \DBQPG algorithm is designed to compute the gradient from a batch of samples (parallelly) leveraging the automatic differentiation framework and fast kernel computation methods, thus placing \BQ-\PG in the context of contemporary \PG algorithms (see Fig.~\ref{fig:DBQPG_UAPG_diagram}(left)). In contrast, the original \BQ-\PG method \citep{BAC_1} was designed to process the samples sequentially (slow). Other major improvements in \DBQPG include (i) replacing a traditional inducing points method ($\mathcal{O}(m^2n+m^3)$ time and $\mathcal{O}(mn+m^2)$ storage) with SKI, a more efficient alternative ($\mathcal{O}(n+m^2)$ time and storage), and (ii) replacing a fixed base kernel for $k_s$ with the more expressive deep kernel, followed by learning the kernel parameters (Eq.~\ref{eqn:GP_nll}). The performance gains from deep kernel learning and SKI are documented in supplement Sec.~\ref{sec:appendix_BACvsDBQPG}.

For \DBQPG, our default choice for the prior state covariance function $k_s$ is a deep RBF kernel
which comprises of an RBF base kernel on top of a DNN feature extractor. Our choice of RBF as the base kernel is based on: (i) its compelling theoretical properties such as infinite basis expansion and universal function approximation \citep{micchelli_06_rbf} and (ii) its compatibility with the Toeplitz method. Thus, the overall computational complexity of estimating $\mL_{\vtheta}^{BQ}$ (Eq.~\ref{eq:GP_grad}) is $\mathcal{O}(p(n+Ym \log m))$ time and $\mathcal{O}(n+Ym)$ storage ($Y$: state dimensionality; $p$: CG iterations; $m$: number of inducing points in SKI+Toeplitz approximation for a deep RBF kernel; automatic differentiation for Fisher kernel).

Note that we intentionally left out kernel learning from the complexity analysis since a naive implementation of the gradient-based optimization step (Eq.~\ref{eqn:GP_nll}) incurs a cubic time complexity in sample size. Our implementation relies on the black-box matrix-matrix multiplication (BBMM) feature (a batched version of CG algorithm that effectively uses GPU hardware) offered by the \textit{GPyTorch library} \cite{gpytorch}, coupled with a SKI approximation over $\mK_s$. This combination offers a linear-time routine for kernel learning. In other words, \DBQPG with kernel learning is a linear-time program that leverages GPU acceleration to efficiently estimate the gradient of a large policy network, with a few thousands of parameters, on high-dimensional domains.

\section{Uncertainty Aware Policy Gradient}
We propose \UAPG, a novel uncertainty aware \PG method that utilizes the gradient uncertainty $\mC_{\vtheta}^{BQ}$ from \DBQPG to provide more reliable policy updates. Most classical \PG methods consider stochastic gradient estimates as the true gradient, without accounting the uncertainty in their gradient components, thus, occasionally taking large steps along the directions of high uncertainty. \UAPG provides more reliable policy updates by lowering the stepsize along the directions with high uncertainty. In particular, \UAPG uses $\mC_{\vtheta}^{BQ}$ to normalize the components of $\mL_{\vtheta}^{BQ}$ with their respective uncertainties, bringing them all to the same scale. Thus, \UAPG offers \PG estimates with a uniform uncertainty, i.e. their gradient covariance is the identity matrix. See Fig.~\ref{fig:DBQPG_UAPG_diagram} (right). In theory, the \UAPG update can be formulated as $\big( \mC_{\vtheta}^{BQ} \big)^{-\frac{1}{2}}\mL_{\vtheta}^{BQ}$.

\textbf{Practical \UAPG Algorithm:}
Empirical $\mC_{\vtheta}^{BQ}$ estimates are often ill-conditioned matrices (spectrum decays quickly) with a numerically unstable inversion.
Since $\mC_{\vtheta}^{BQ}$ only provides a faithful estimate of the top few principal directions of uncertainty, the \UAPG update is computed from a rank-$\delta$ singular value decomposition (SVD) approximation of $\mC_{\vtheta}^{BQ} \approx \nu_\delta \mI + \sum_{i=1}^{\delta} \vh_i (\nu_i - \nu_\delta) \vh_i^\top$ as follows:
\begin{align}
\label{eqn:UAPG_practical_vanilla}
   \mL_{\vtheta}^{UAPG} = \nu_{\delta}^{-\frac{1}{2}} \big(\mI + \sum_{i=1}^{\delta} \vh_i \big(\sqrt{{\nu_{\delta}}/{\nu_i}} - 1\big)\vh_i^\top \big) \mL_{\vtheta}^{BQ}.
\end{align}
The principal components (PCs) $\{\vh_i\}_{i=1}^{\delta}$ denote the top $\delta$ directions of uncertainty and the singular values $\{\nu_i\}_{i=1}^{\delta}$ denote their corresponding magnitudes of uncertainty. The rank-$\delta$ decomposition of $\mC_{\vtheta}^{BQ}$ can be computed in linear-time using the randomized SVD algorithm \citep{randomizedSVD}. The \UAPG update in Eq.~\ref{eqn:UAPG_practical_vanilla} dampens the stepsize of the top $\delta$ directions of uncertainty, relative to the stepsize of remaining gradient components.
Thus, in comparison to \DBQPG, \UAPG lowers the risk of taking large steps along the directions of high uncertainty, providing reliable policy updates.

For natural gradient $\mL_{\vtheta}^{NBQ} = \mG^{-1} \mL_{\vtheta}^{BQ}$, the gradient uncertainty can be computed as follows:
\begin{align}
\label{eqn:natural_bq_covariace}
    &\mC_{\vtheta}^{NBQ} = \mG^{-1} \mC_{\vtheta}^{BQ} \mG^{-1}\nonumber\\
    &= c_2 \mG^{-1} - c_2^2 \mG^{-1} \mU \left( c_1 \mK_s +c_2 \mK_f + \sigma^2 \mI \right)^{-1} \mU^\top \mG^{-1} \nonumber
    \\
    &= c_2 (\mG + c_2 \mU \left( c_1 \mK_s + \sigma^2 \mI \right)^{-1} \mU^\top)^{-1}.
\end{align}
Since $\mC_{\vtheta}^{NBQ}$ is the inverse of an ill-conditioned matrix, we instead apply a low-rank approximation on ${\mC_{\vtheta}^{NBQ}}^{-1} \approx \nu_\delta \mI + \sum_{i=1}^{\delta} \vh_i (\nu_i - \nu_\delta) \vh_i^\top$ for the \UAPG update of \NPG:
\begin{equation}
\label{eqn:UAPG_npg}
    \mL_{\vtheta}^{NUAPG} = \nu_{\delta}^{\frac{1}{2}} \Big(\mI + \sum_{i=1}^{\delta} \vh_i \big( \min \big(\sqrt{\frac{\nu_i}{\nu_{\delta}}}, \eps \big) - 1\big)\vh_i^\top \Big) \mG^{-1} \mL_{\vtheta}^{BQ},
\end{equation}
%
where, $\{\vh_i, \nu_i\}_{i=1}^\delta$ correspond to the top $\delta$ PCs of ${\mC_{\vtheta}^{NBQ}}^{-1}$ (equivalently the bottom $\delta$ PCs of $\mC_{\vtheta}^{NBQ}$), and $\eps > 1$ is a hyperparameter. We replace $\sqrt{{\nu_i}/{\nu_{\delta}}}$ with $\min(\sqrt{{\nu_i}/{\nu_\delta}},\eps)$ to avoid taking large steps along these directions, solely on the basis of their uncertainty estimates.
For $c_2 \ll 1$, it is interesting to note that (i) $\mC_{\vtheta}^{BQ} \approx c_2 \mG$ and $\mC_{\vtheta}^{NBQ} \approx c_2 \mG^{-1}$, which implies that the most uncertain gradient directions for vanilla \PG approximately correspond to the most confident directions for \NPG, and (ii) the ideal UAPG update for both vanilla \PG and \NPG converges along the $\mG^{-\frac{1}{2}}\mL_{\vtheta}^{BQ}$ direction. A more rigorous discussion on the relations between Vanilla \PG, \NPG and their \UAPG updates can be found in the supplement Sec.~\ref{sec:appendix_vpg_npg_uapg}.
%
%



\begin{algorithm}
  \caption{\BQ-\PG Estimator Subroutine}\label{alg:BDQPG}
  \begin{algorithmic}[1]
    \State \textbf{\BQ-\PG}($\vtheta, n$) 
      \Statex \textbullet~$\vtheta$: policy parameters
      \Statex \textbullet~$n$: sample size for \PG estimation
        \State Collect $n$ state-action pairs (samples) from running the policy $\pi_{\vtheta}$ in the environment.
        \State Compute action-values $\Qmc$ for these $n$ samples using \MC rollouts (TD(1) estimate) or an explicit critic network.
        \State Update kernel parameters and explicit critic's parameters using \GP MLL (Eq.~\ref{eqn:GP_nll}) and TD error respectively.
        \State Policy gradient estimation (\DBQPG):
         \Statex $
         \begin{small}\mL_{\vtheta} =
              \begin{cases}
                       \mL_{\vtheta}^{BQ} & \text{(Vanilla \PG)}\\
                       \mG^{-1} \mL_{\vtheta}^{BQ} & \text{(Natural \PG)}
              \end{cases}\end{small}$
        \State (\textbf{Optional}) Compute $\{\vh_i, \nu_i\}_{i=1}^{\delta}$ using fast SVD over ${\mC_{\vtheta}^{BQ}}$ (Eq.~\ref{eq:GP_grad}, vanilla \PG) or ${\mC_{\vtheta}^{NBQ}}^{-1}$ (Eq.~\ref{eqn:natural_bq_covariace}, \NPG).
        \State (\textbf{Optional}) Uncertainty-based adjustment (\UAPG):
         \Statex $
         \begin{small}\mL_{\vtheta} =
              \begin{cases}
                       \nu_{\delta}^{-\frac{1}{2}} \left(\mI + \sum\limits_{i=1}^{\delta} \vh_i \left(\sqrt{\frac{\nu_{\delta}}{\nu_i}} - 1\right)\vh_i^\top \right) \mL_{\vtheta}\\
                       \text{(Vanilla \PG)} \\
                       \nu_{\delta}^{\frac{1}{2}} \left(\mI + \sum\limits_{i=1}^{\delta} \vh_i \left( \min \left(\sqrt{\frac{\nu_i}{\nu_\delta}}, \eps \right) - 1\right)\vh_i^\top \right) \mG^{-1}\mL_{\vtheta} \\
                       \text{(Natural \PG)}
              \end{cases}\end{small}$
        
      \State \textbf{return} $\mL_{\vtheta}$
  \end{algorithmic}
\end{algorithm}

\section{Experiments}
\label{sec:experiments}
We study the behaviour of \BQ-\PG methods (Algorithm~\ref{alg:BDQPG}) on MuJoCo environments, using the \textit{mujoco-py} library of OpenAI Gym \citep{gym}.
In our experiments, we replace $\Qmc$ with generalized advantage estimates \citep{GAE} computed using an explicit state-value critic network, i.e., $V_{\pi_{\vtheta}}(s)$ is approximated using a linear layer on top of the state feature extractor in Fig.~\ref{fig:DBQPG_UAPG_diagram}(left).

\textbf{Quality of Gradient Estimation:} Inspired from the experimental setup of \citet{accuracy_variance_PG}, we evaluate the quality of \PG estimates obtained via \DBQPG and \MC estimation using two metrics: (i) \textbf{gradient accuracy} or the average cosine similarity of the obtained gradient estimates with respect to the true gradient estimates (estimated from $10^6$ state-action pairs) and (ii) \textbf{variance} in the gradient estimates (normalized by the norm of the mean gradient for scale invariance).
%
See Fig.~\ref{fig:pg_accuracy_variance}. We observe that \DBQPG provides more accurate gradient estimates with a considerably lower variance. Interestingly, \DBQPG and \MC estimates offer nearly the same quality gradients at the start of training. However, as the training progresses, and \DBQPG learns kernel bases,
we observe that \DBQPG returns superior quality gradient estimates. Moreover, as training progress from $0$ to $150$ iterations, the gradient norms of both \DBQPG and \MC estimates drop by a factor of $3$, while the ``unnormalized'' gradient variances increase by $5$ folds. The drop in the signal-to-noise ratio for gradient estimation explains the fall in accuracy over training time. Further, the wall-clock time of \DBQPG and \UAPG updates is comparable to \MC-\PG (Fig.~\ref{fig:appendix_wall_clock} in supplement). These results motivate substituting \MC with \BQ-based \PG estimates in deep \PG algorithms.
%
%
\newcommand{\lineW}{0.136}
\begin{figure*}[!ht]
	\centering
    \begin{subfigure}
  	\centering
  	\includegraphics[scale=\lineW]{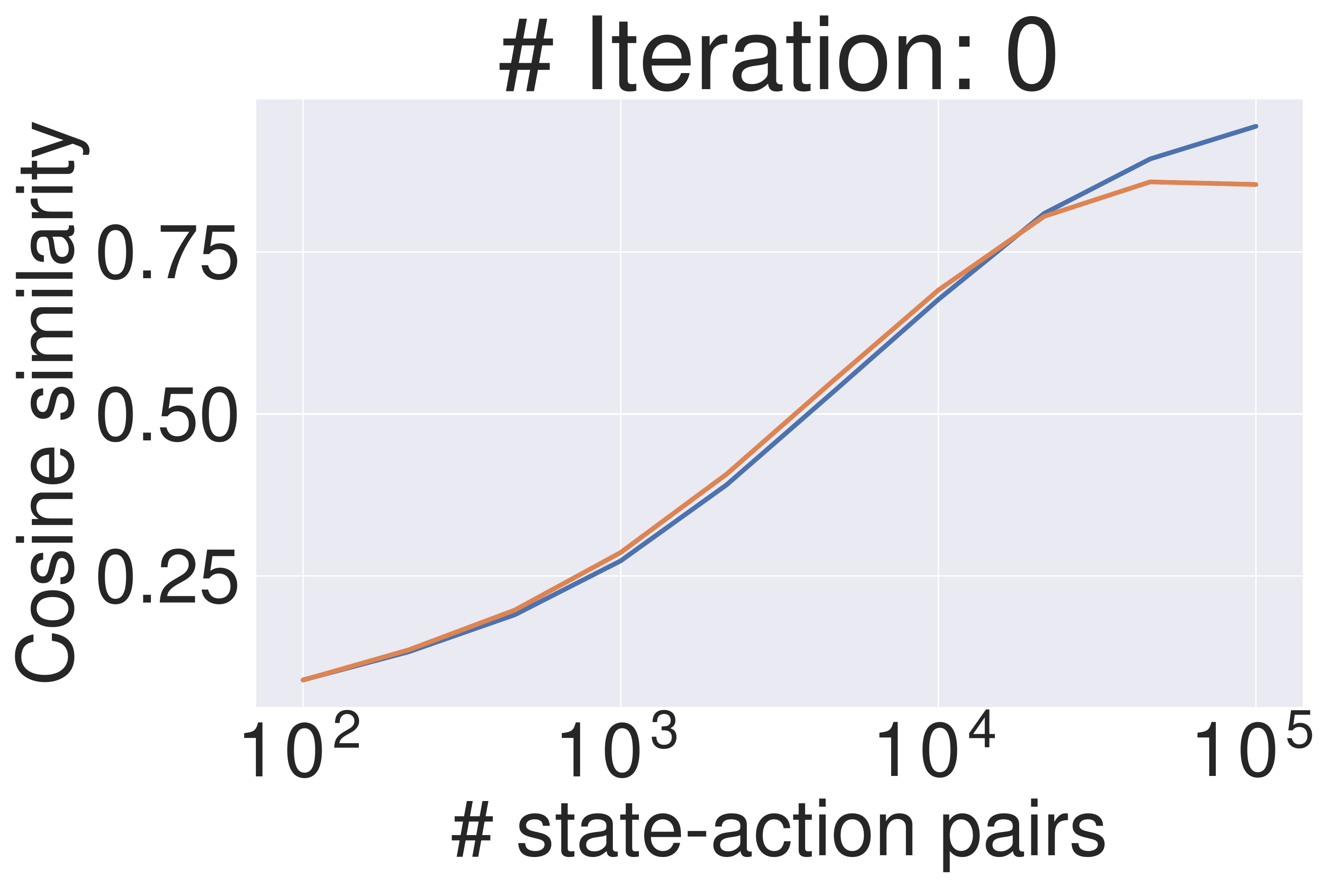}
    \end{subfigure}
    \hspace{-0.7em}
    \begin{subfigure}
  	\centering
  	\includegraphics[scale=\lineW]{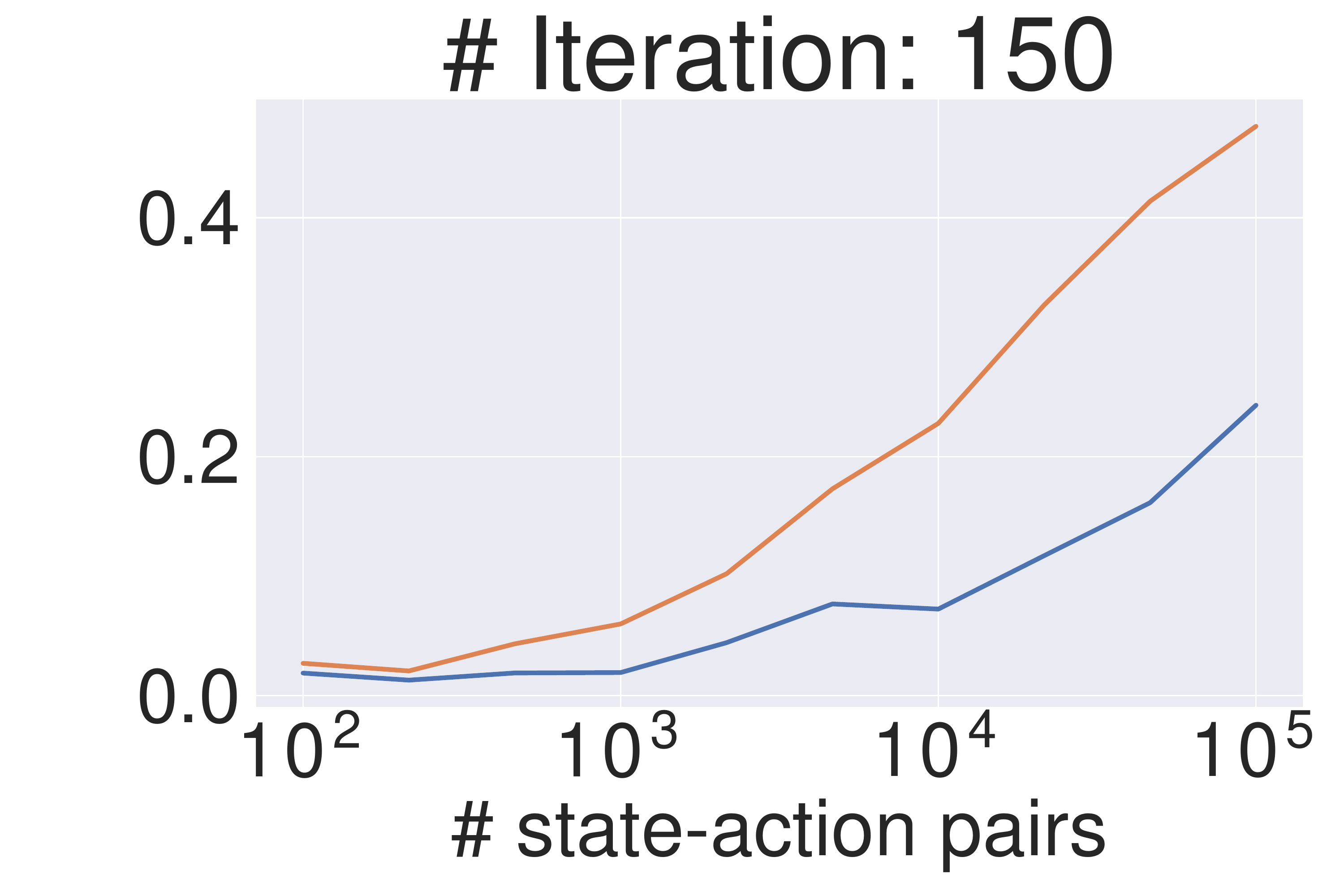}
    \end{subfigure}
    \hspace{-0.7em}
    \begin{subfigure}
  	\centering
  	\includegraphics[scale=\lineW]{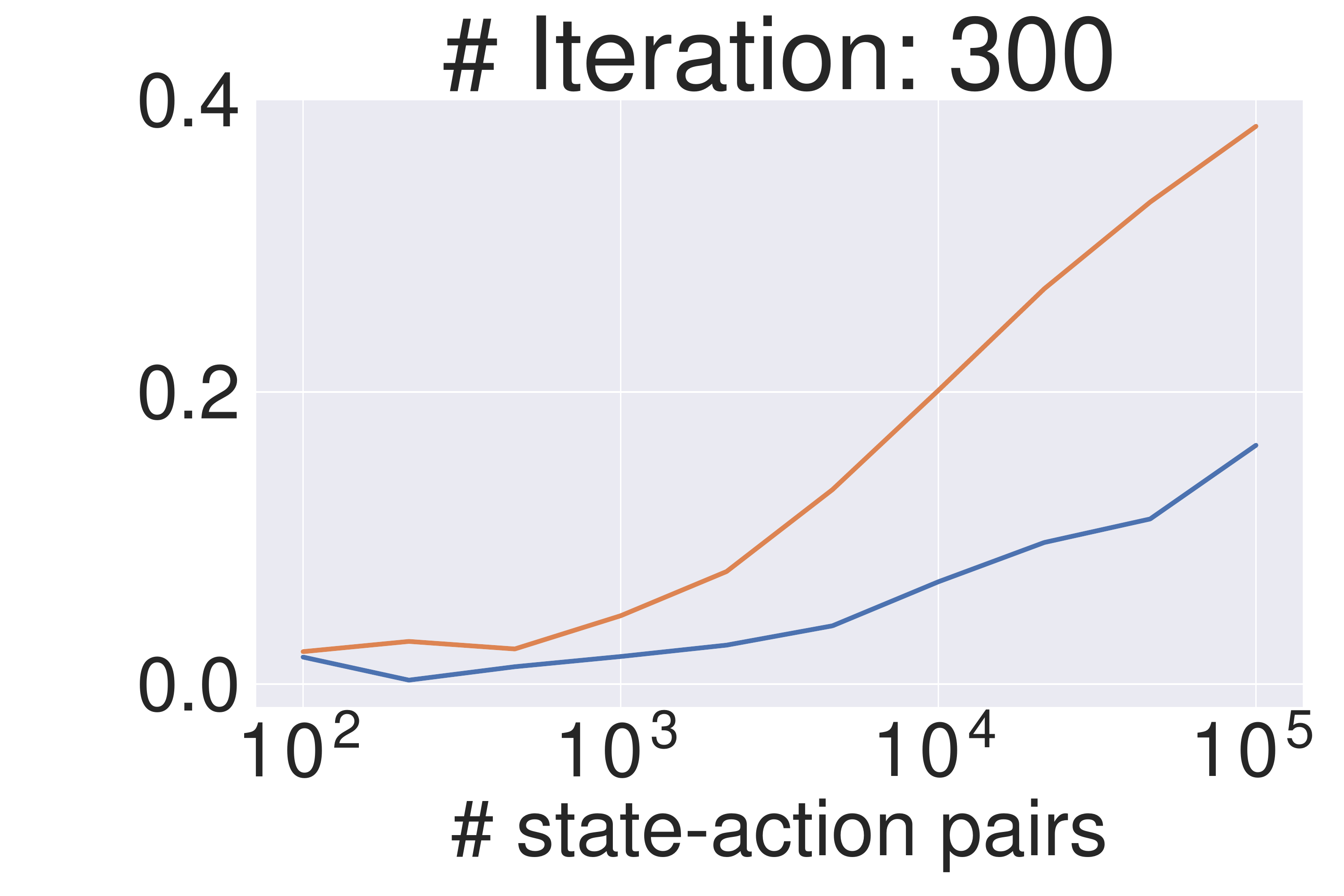}
    \end{subfigure}
    \hspace{-0.7em}
    \begin{subfigure}
  	\centering
  	\includegraphics[scale=\lineW]{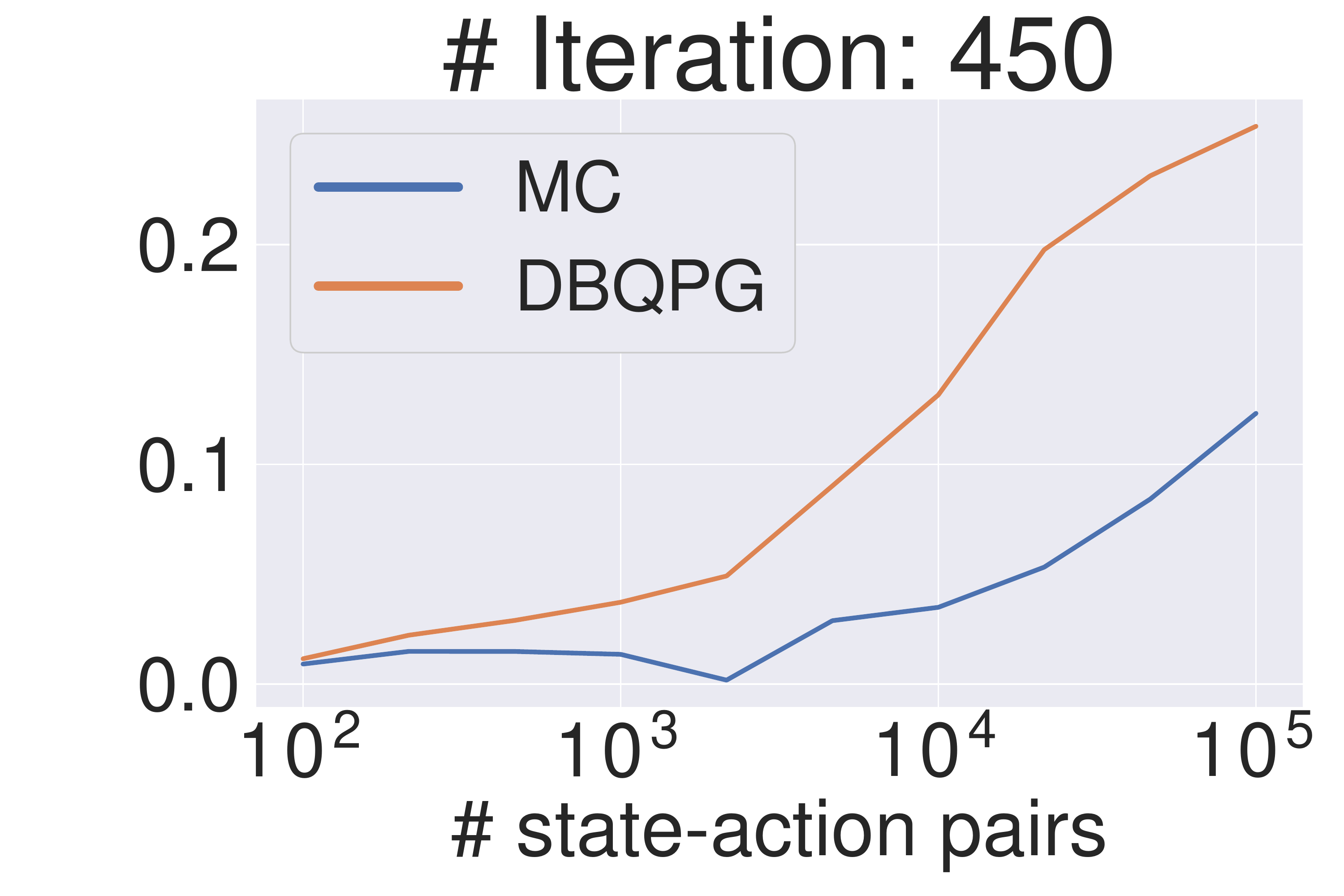}
    \end{subfigure}
    \begin{subfigure}
  	\centering
  	\hspace*{0.005cm}
  	\includegraphics[scale=\lineW]{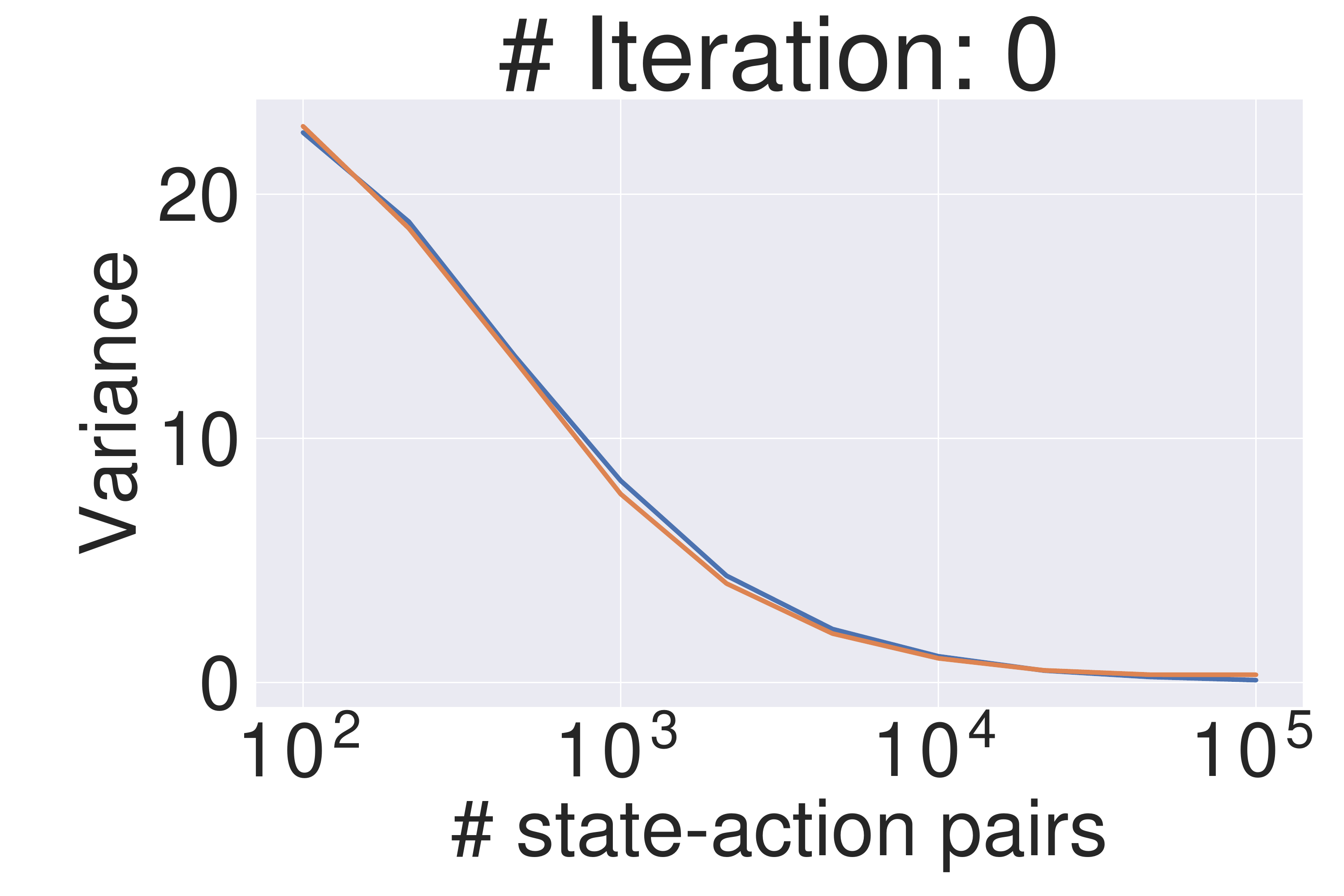}
    \end{subfigure}
    \hspace{-0.7em}
    \begin{subfigure}
  	\centering
  	\includegraphics[scale=\lineW]{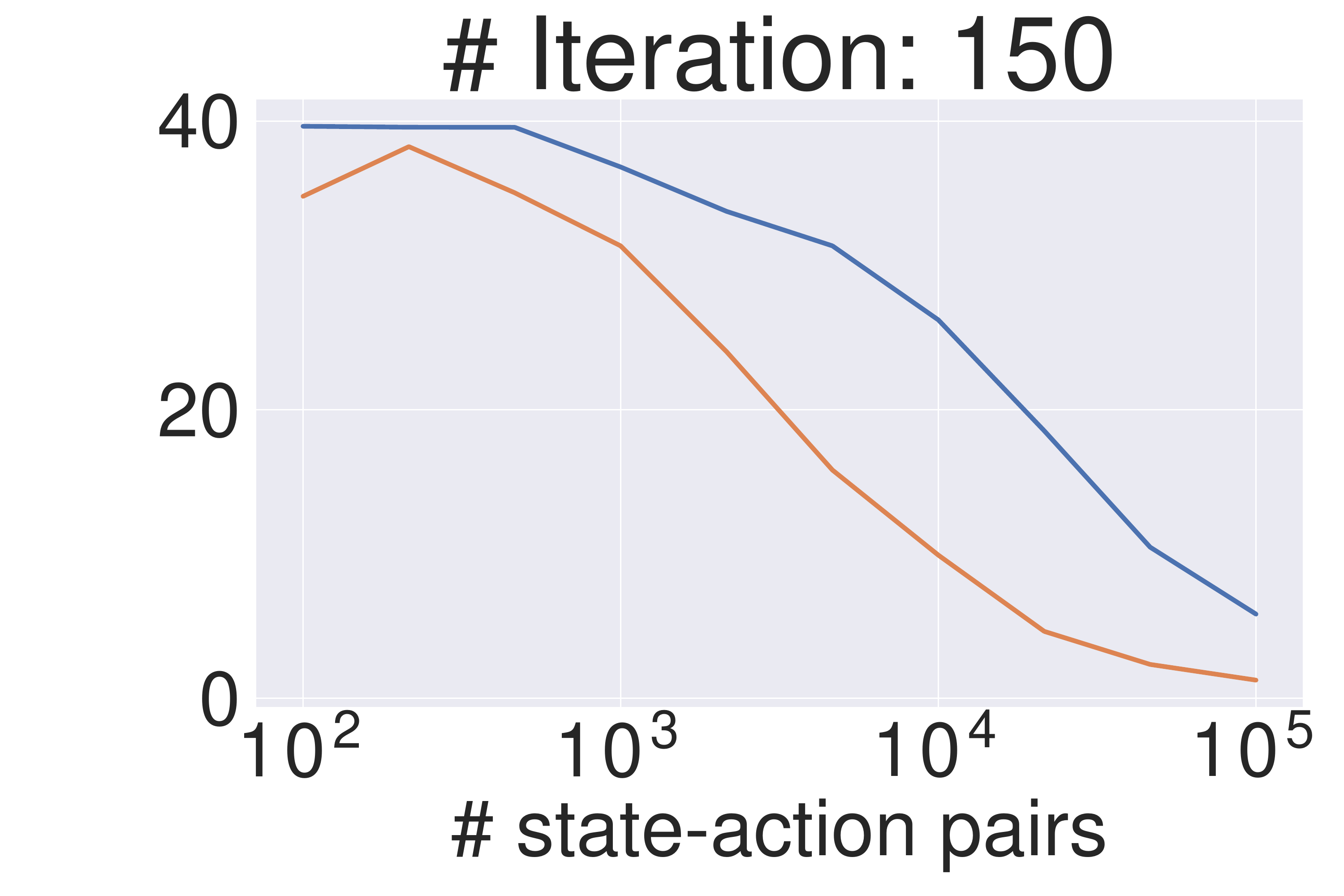}
    \end{subfigure}
    \hspace{-0.7em}
    \begin{subfigure}
  	\centering
  	\includegraphics[scale=\lineW]{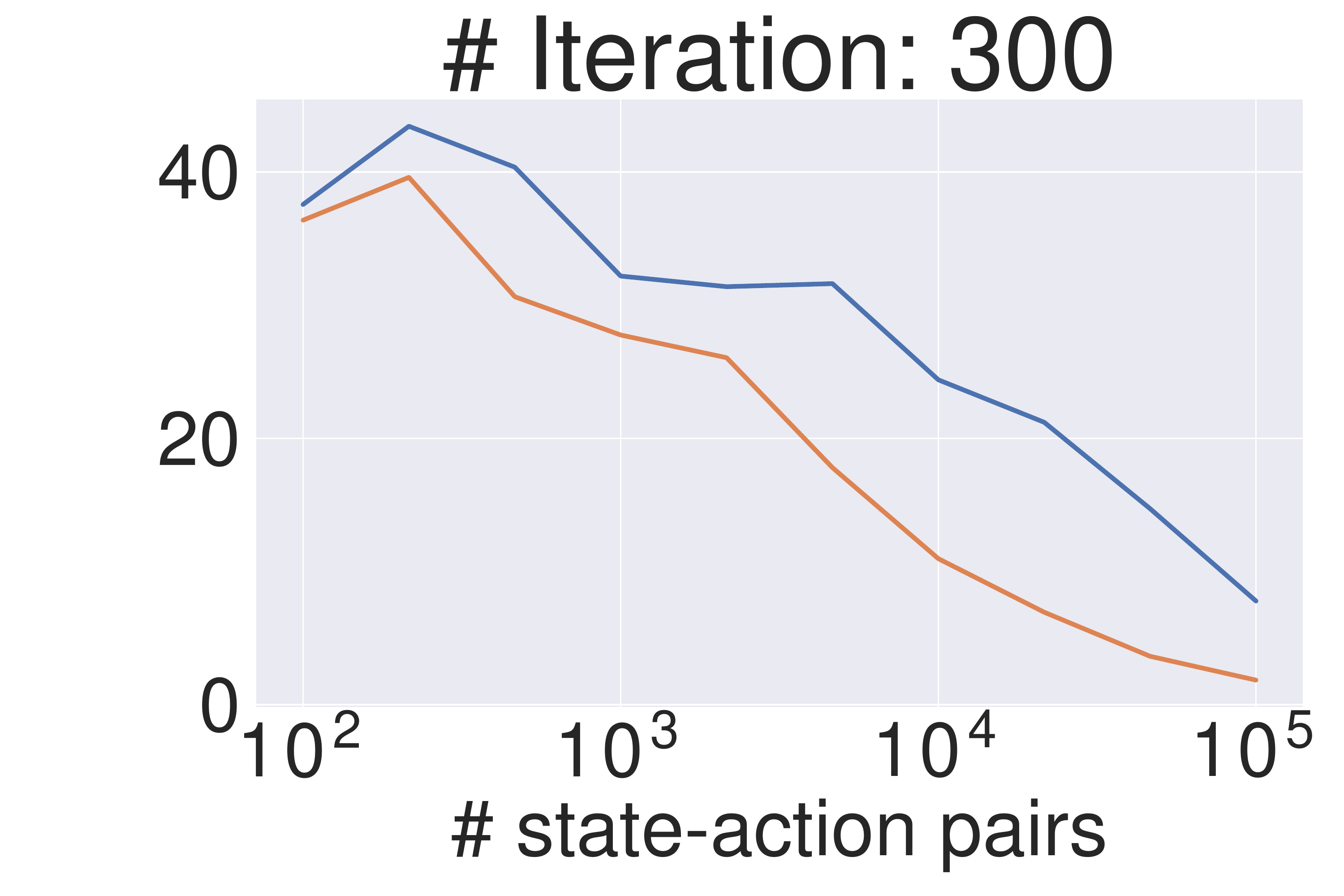}
    \end{subfigure}
    \hspace{-0.7em}
    \begin{subfigure}
  	\centering
  	\includegraphics[scale=\lineW]{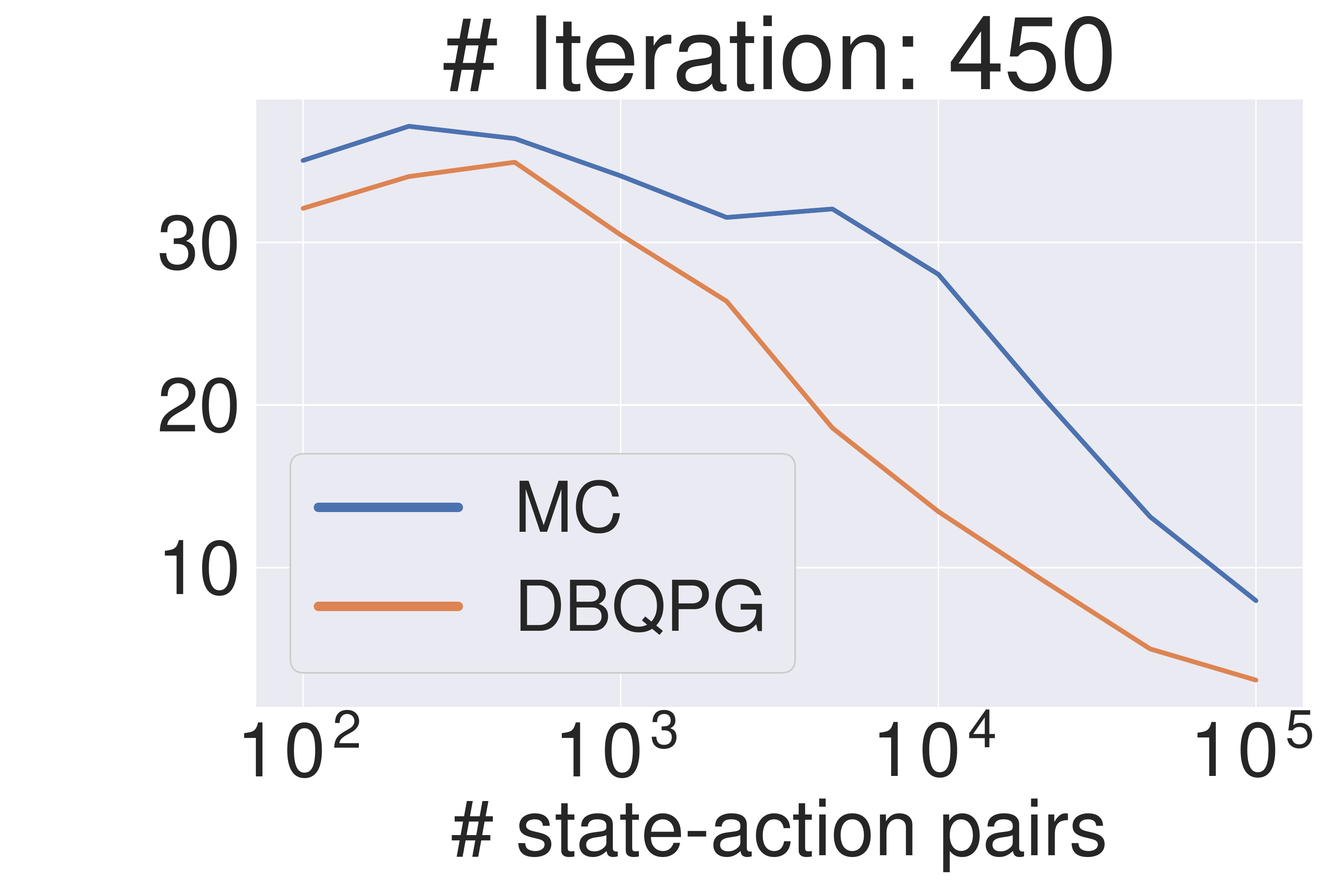}
    \end{subfigure}
\caption{An empirical analysis of the quality of policy gradient estimates as a function of the state-action sample size. The experiments are conducted for $0^{th}$,  $150^{th}$, $300^{th}$, and $450^{th}$ iteration along the training phase of \DBQPG (vanilla PG) algorithm in MuJoCo Swimmer-v2 environment. All the results have been averaged over $25$ repeated gradient measurements across $100$ random runs. (a) The accuracy plot results are obtained w.r.t the ``true gradient'', which is computed using \MC estimates of $10^6$ state-action pairs. (b) The normalized variance is computed using the ratio of trace of empirical gradient covariance matrix (like~\citet{grad_study}) and squared norm of gradient mean.}
  \label{fig:pg_accuracy_variance}
\end{figure*}



\newcommand{\lineWW}{0.041}
\begin{figure*}[h!]
	\centering
    \begin{subfigure}
  	\centering
  	\includegraphics[scale=\lineWW]{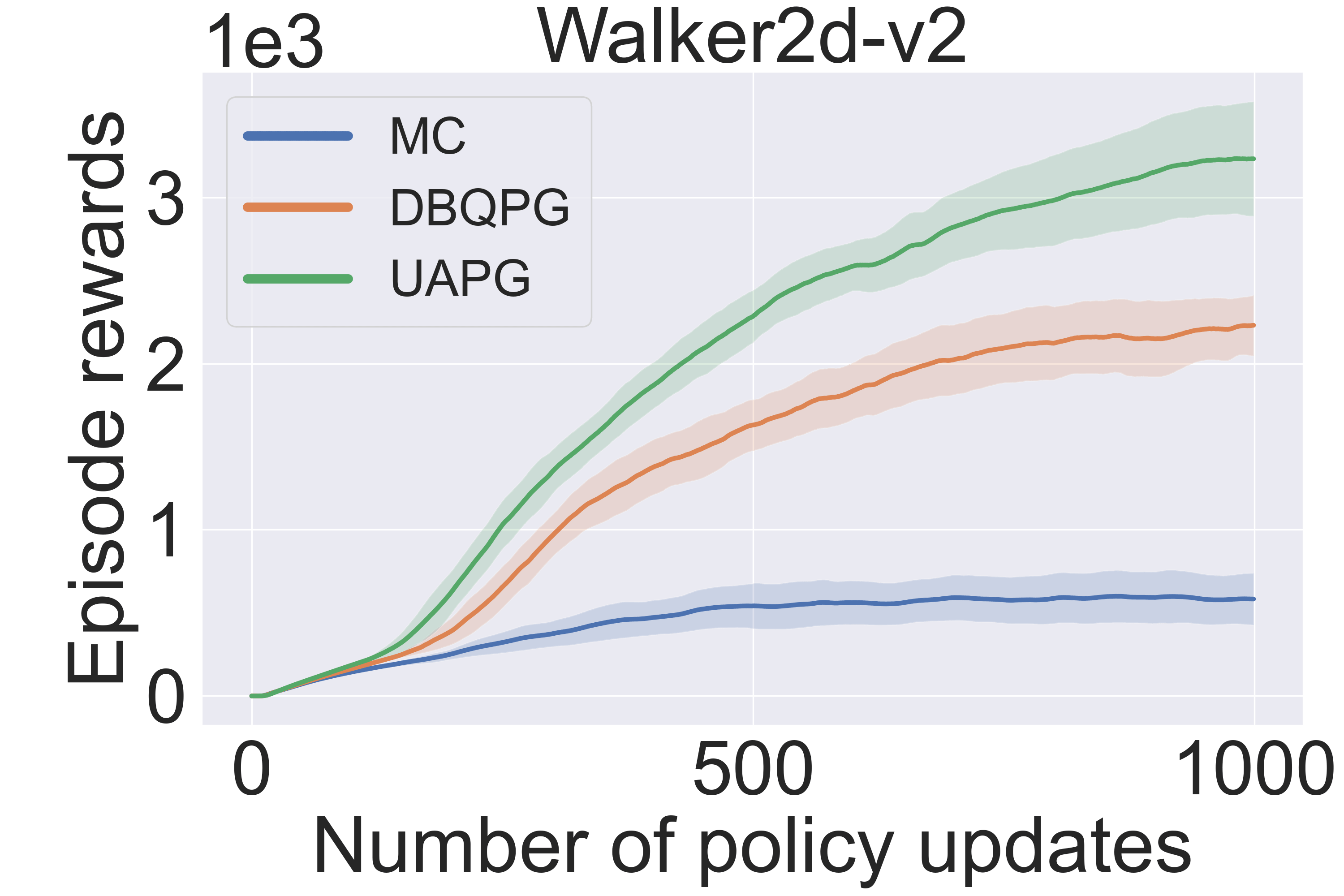}
    \end{subfigure}
    \hspace{-0.7em}
    \begin{subfigure}
  	\centering
  	\includegraphics[scale=\lineWW]{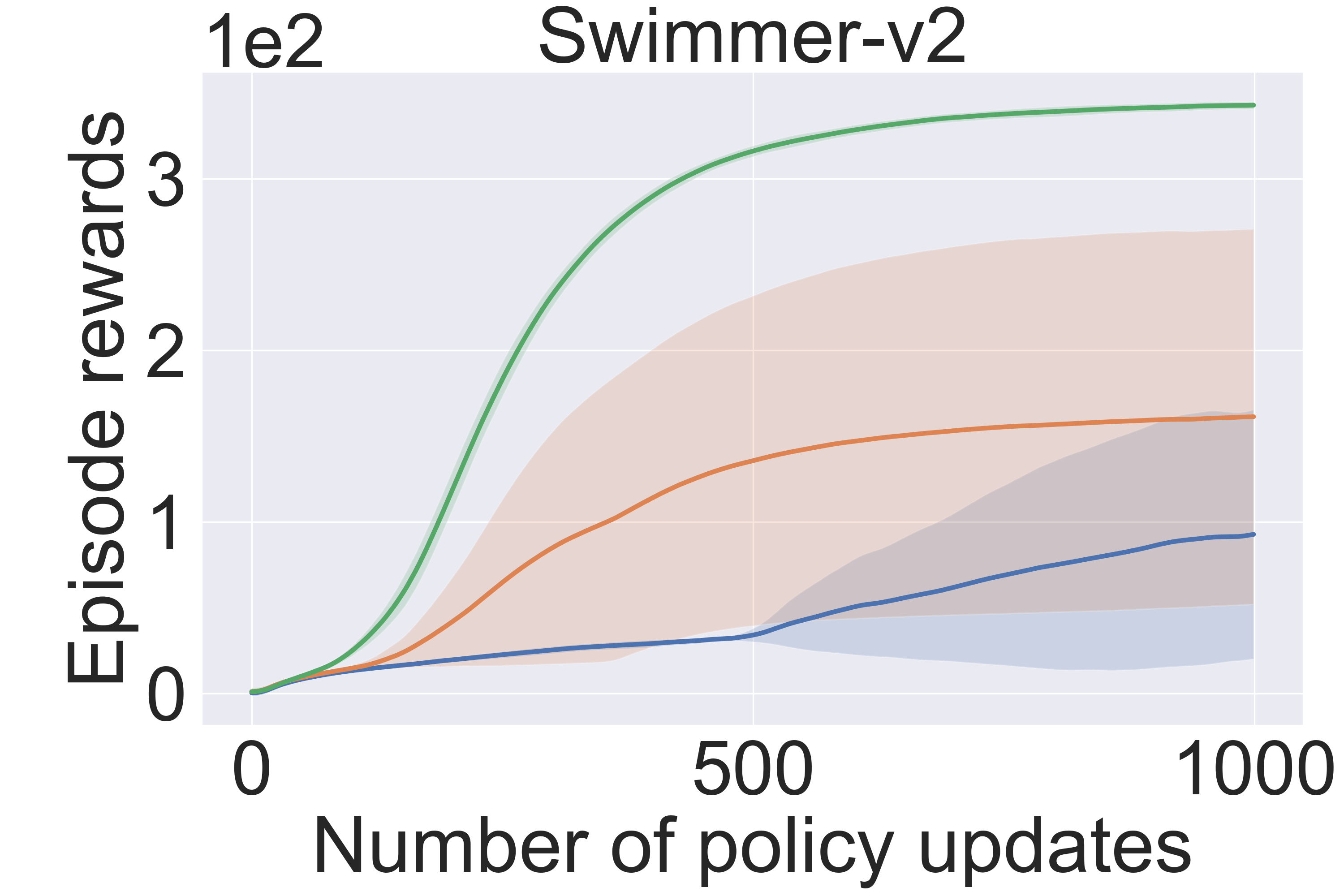}
    \end{subfigure}
    \hspace{-0.7em}
    \begin{subfigure}
  	\centering
  	\includegraphics[scale=\lineWW]{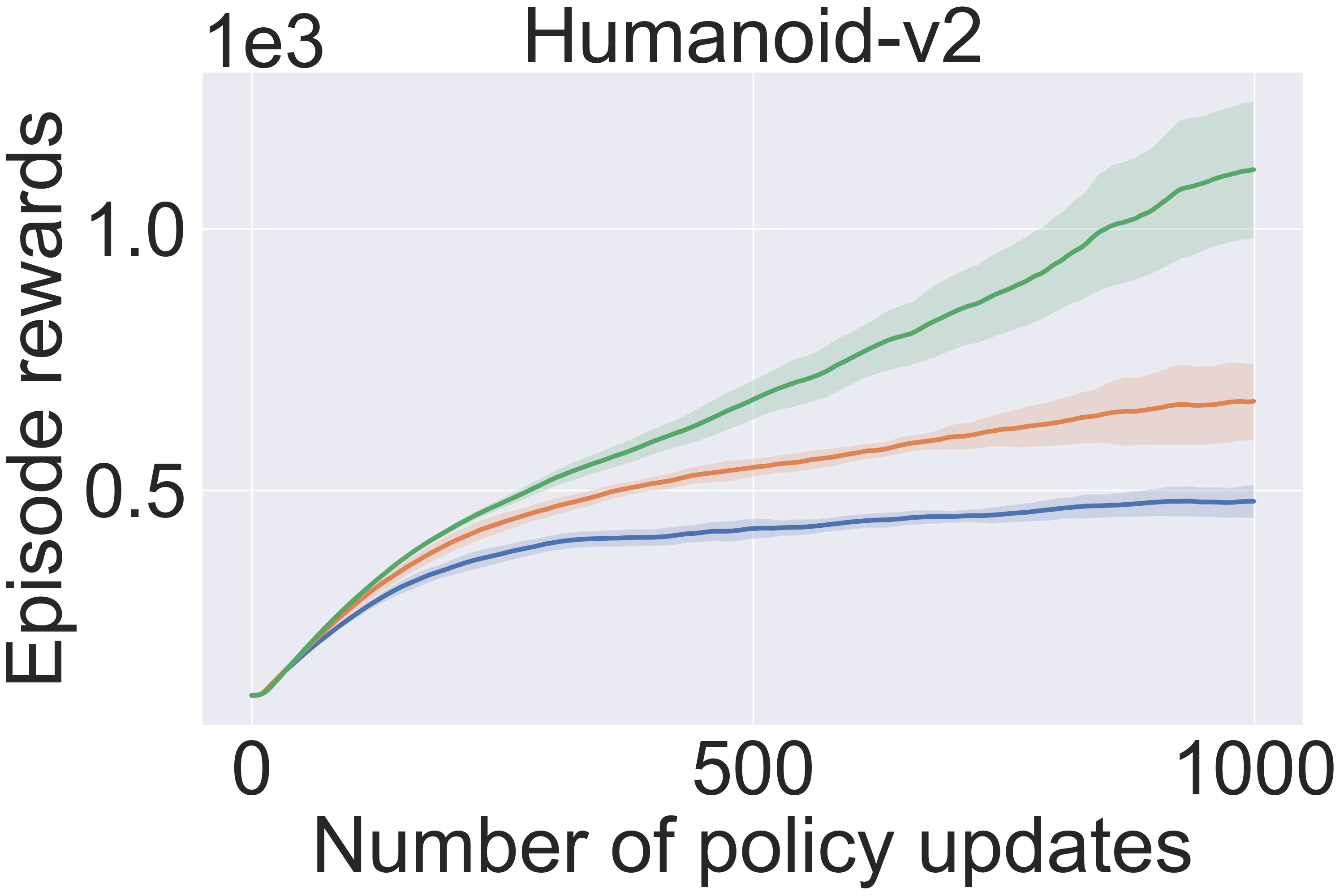}
    \end{subfigure}
    \hspace{-0.7em}
    \begin{subfigure}
  	\centering
  	\includegraphics[scale=\lineWW]{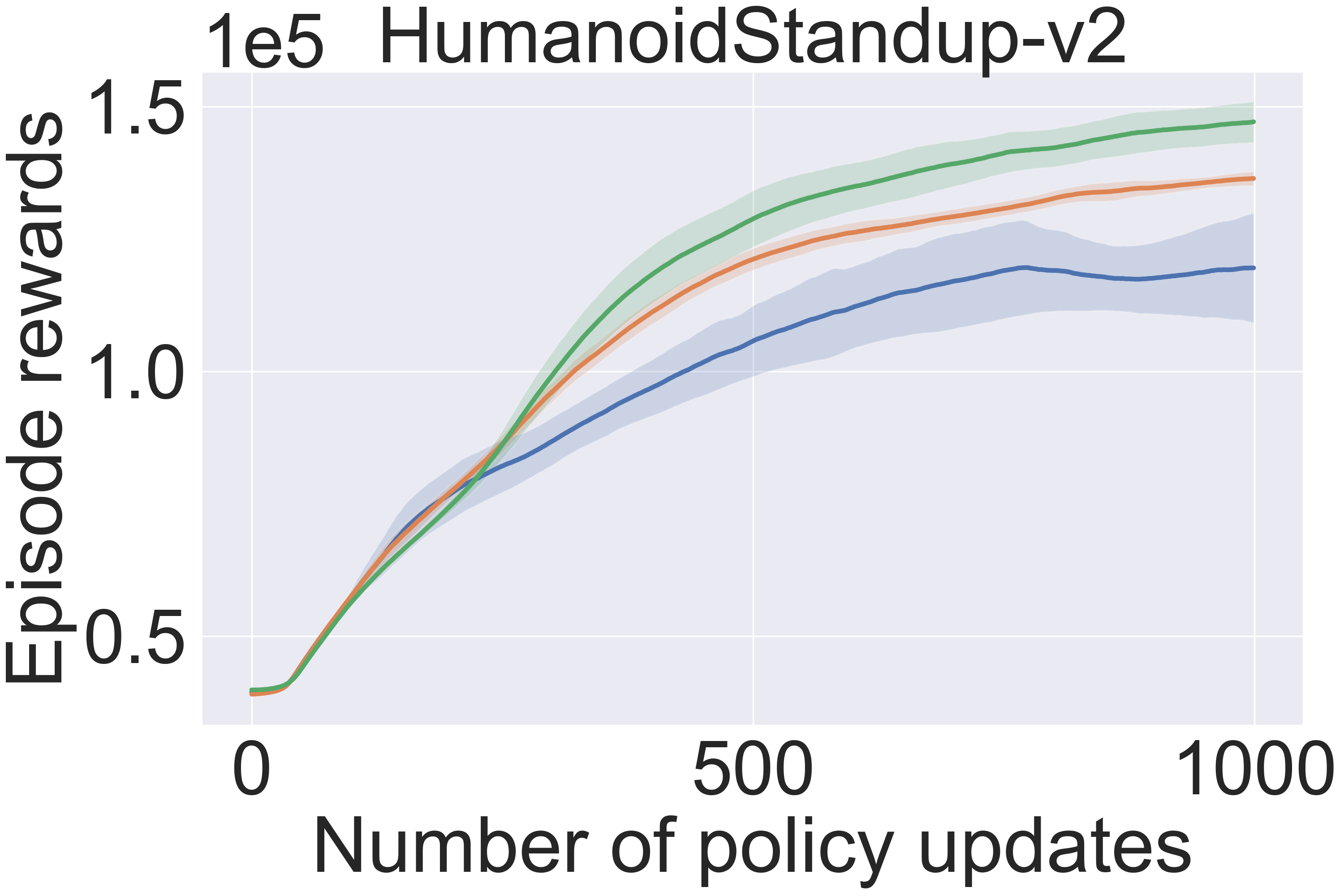}
    \end{subfigure}
    {\LARGE Vanilla \PG}

\begin{center}
\line(1,0){475}
\end{center}
	\centering
    \begin{subfigure}
  	\centering
  	\includegraphics[scale=\lineWW]{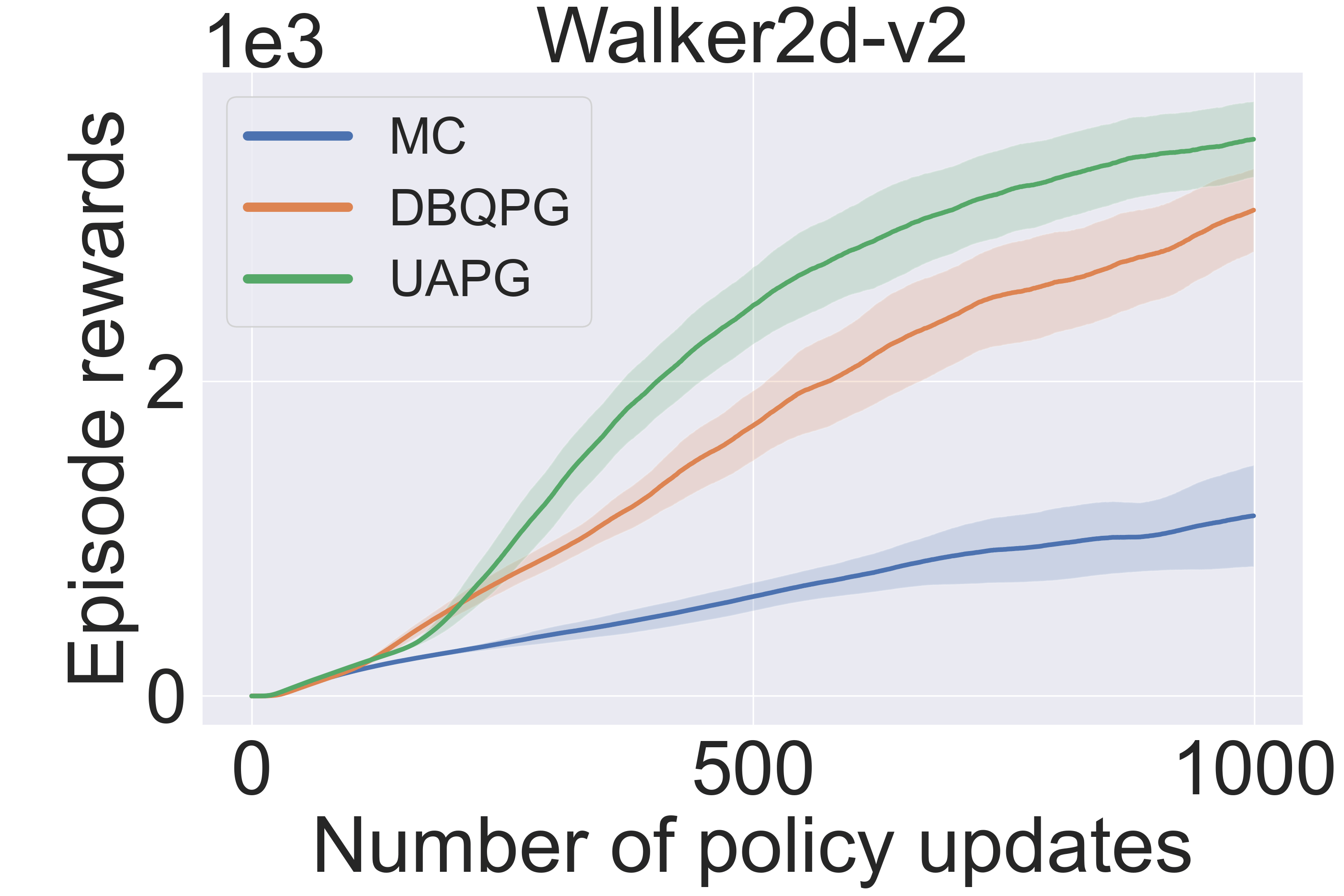}
    \end{subfigure}
    \hspace{-0.7em}
    \begin{subfigure}
  	\centering
  	\includegraphics[scale=\lineWW]{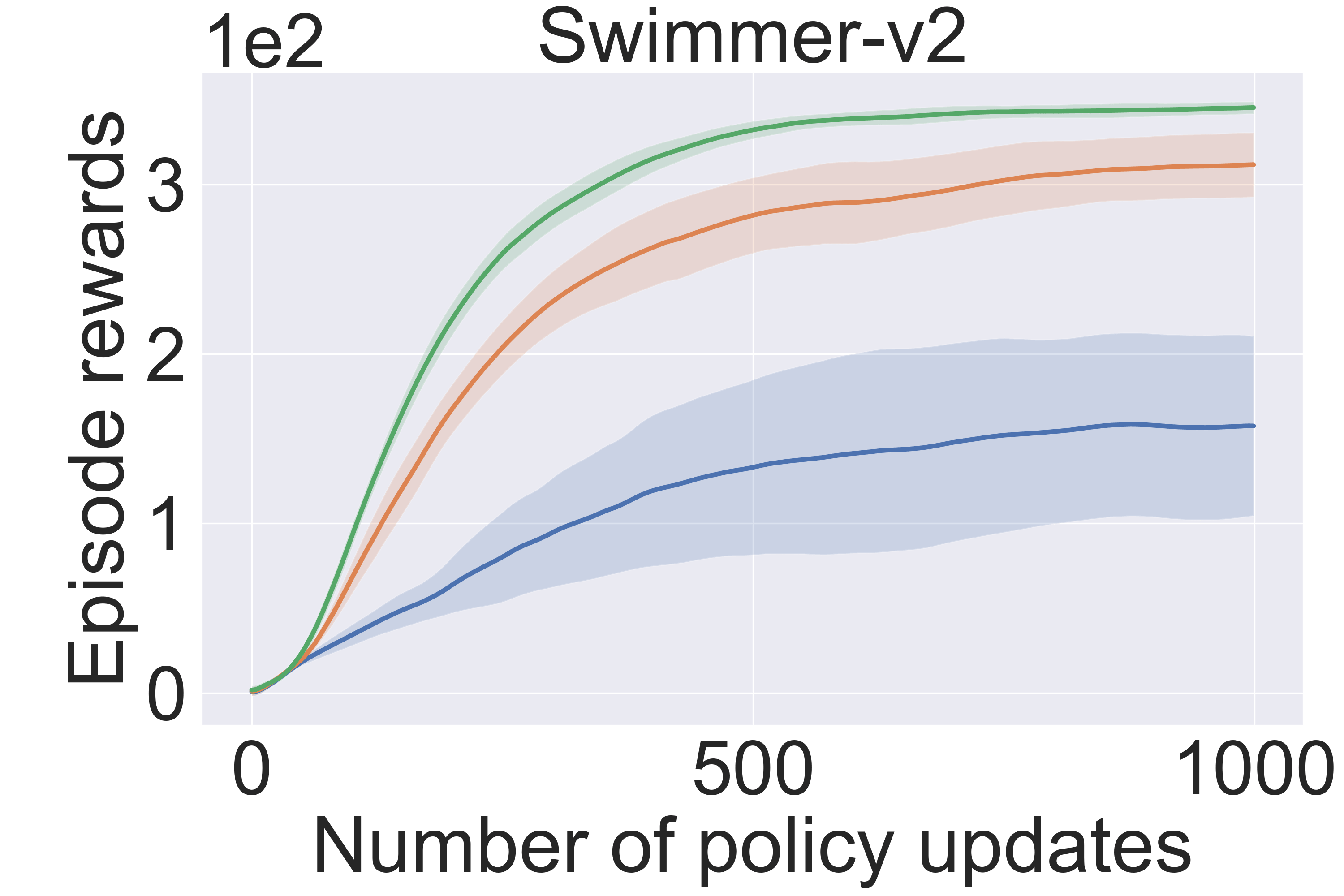}
    \end{subfigure}
    \hspace{-0.7em}
    \begin{subfigure}
  	\centering
  	\includegraphics[scale=\lineWW]{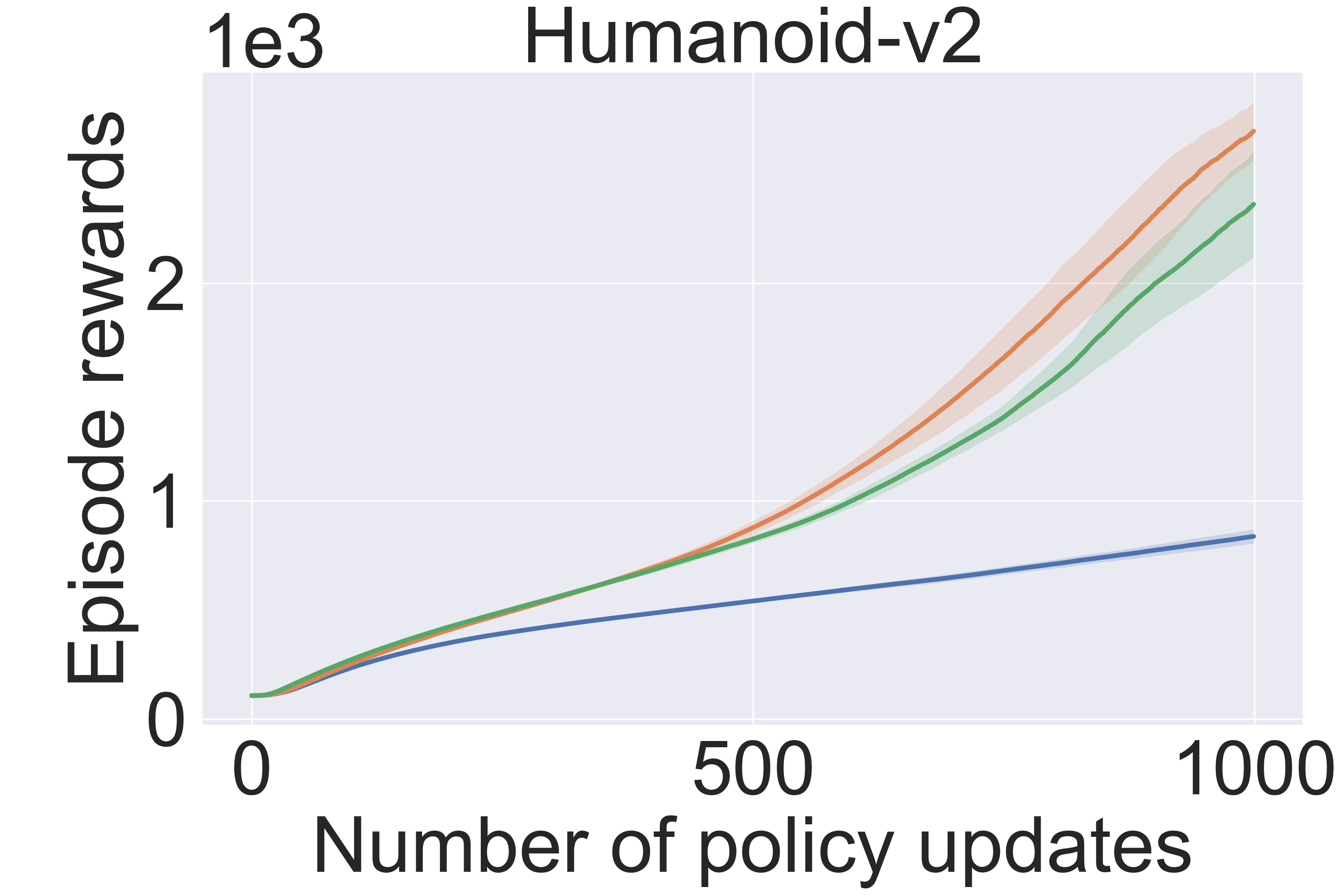}
    \end{subfigure}
    \hspace{-0.7em}
    \begin{subfigure}
  	\centering
  	\includegraphics[scale=\lineWW]{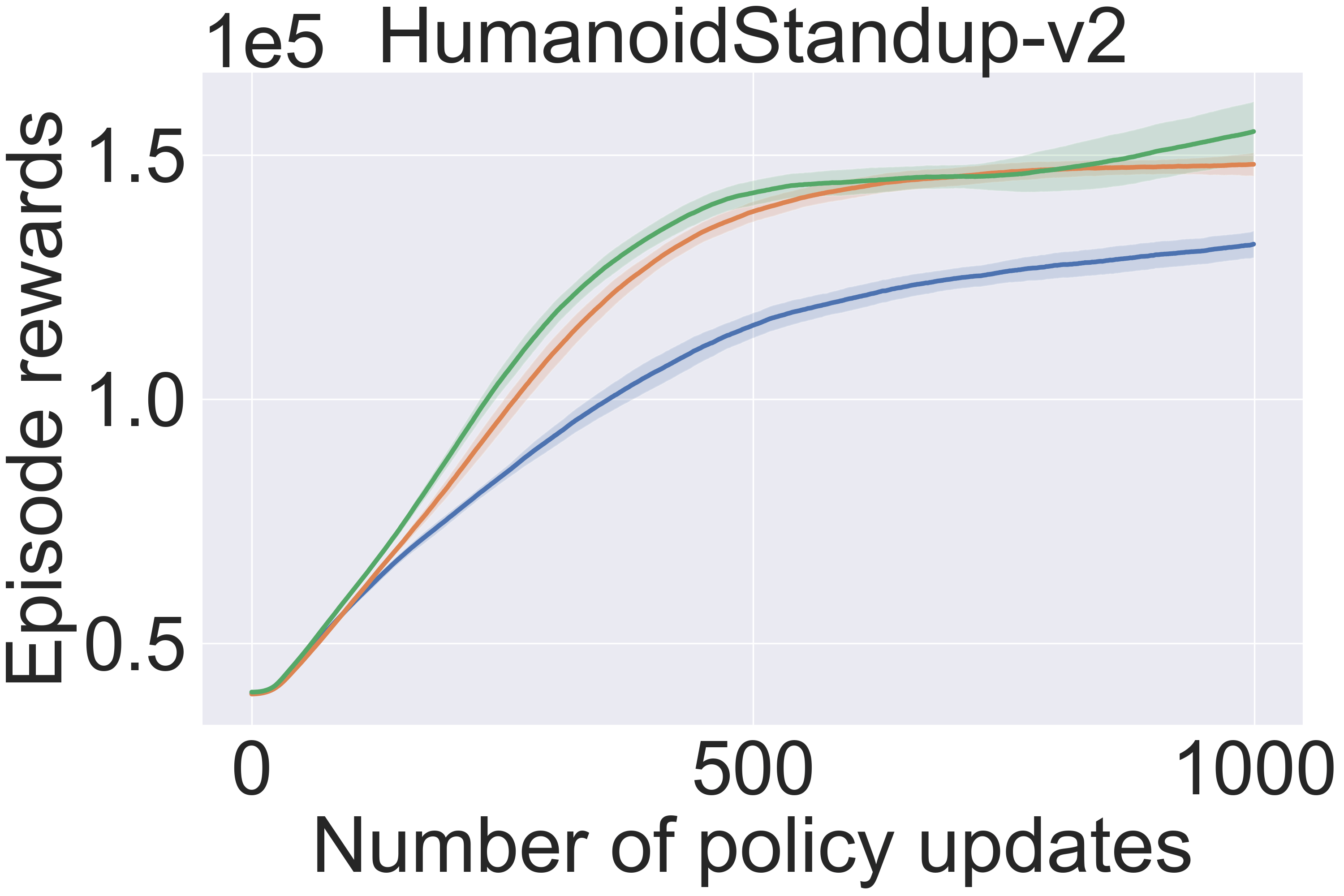}
    \end{subfigure}
    {\LARGE \NPG}
    \begin{center}
    \line(1,0){475}
    \end{center}


	\centering
    \begin{subfigure}
  	\centering
  	\includegraphics[scale=\lineWW]{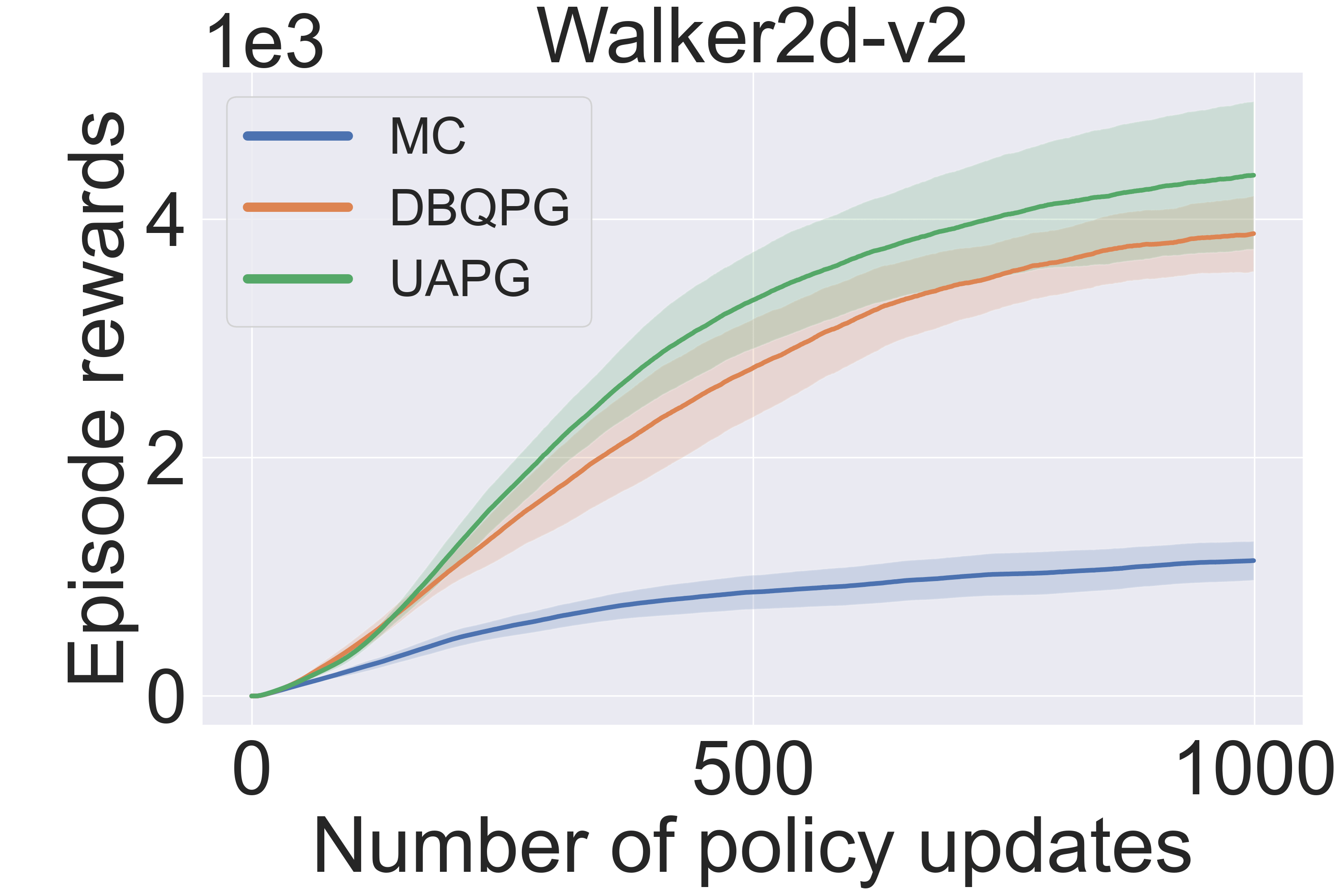}
    \end{subfigure}
    \hspace{-0.7em}
    \begin{subfigure}
  	\centering
  	\includegraphics[scale=\lineWW]{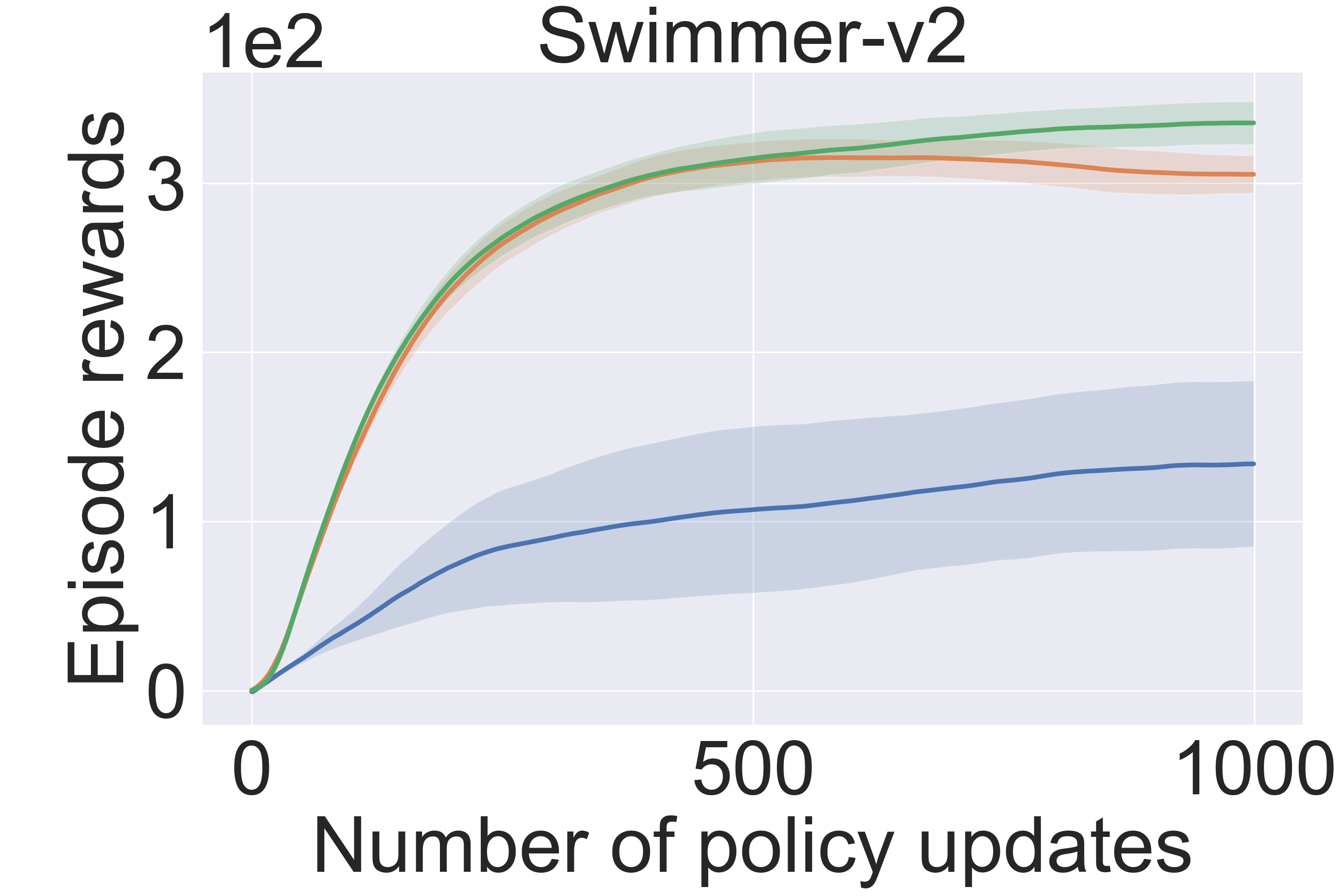}
    \end{subfigure}
    \hspace{-0.7em}
    \begin{subfigure}
  	\centering
  	\includegraphics[scale=\lineWW]{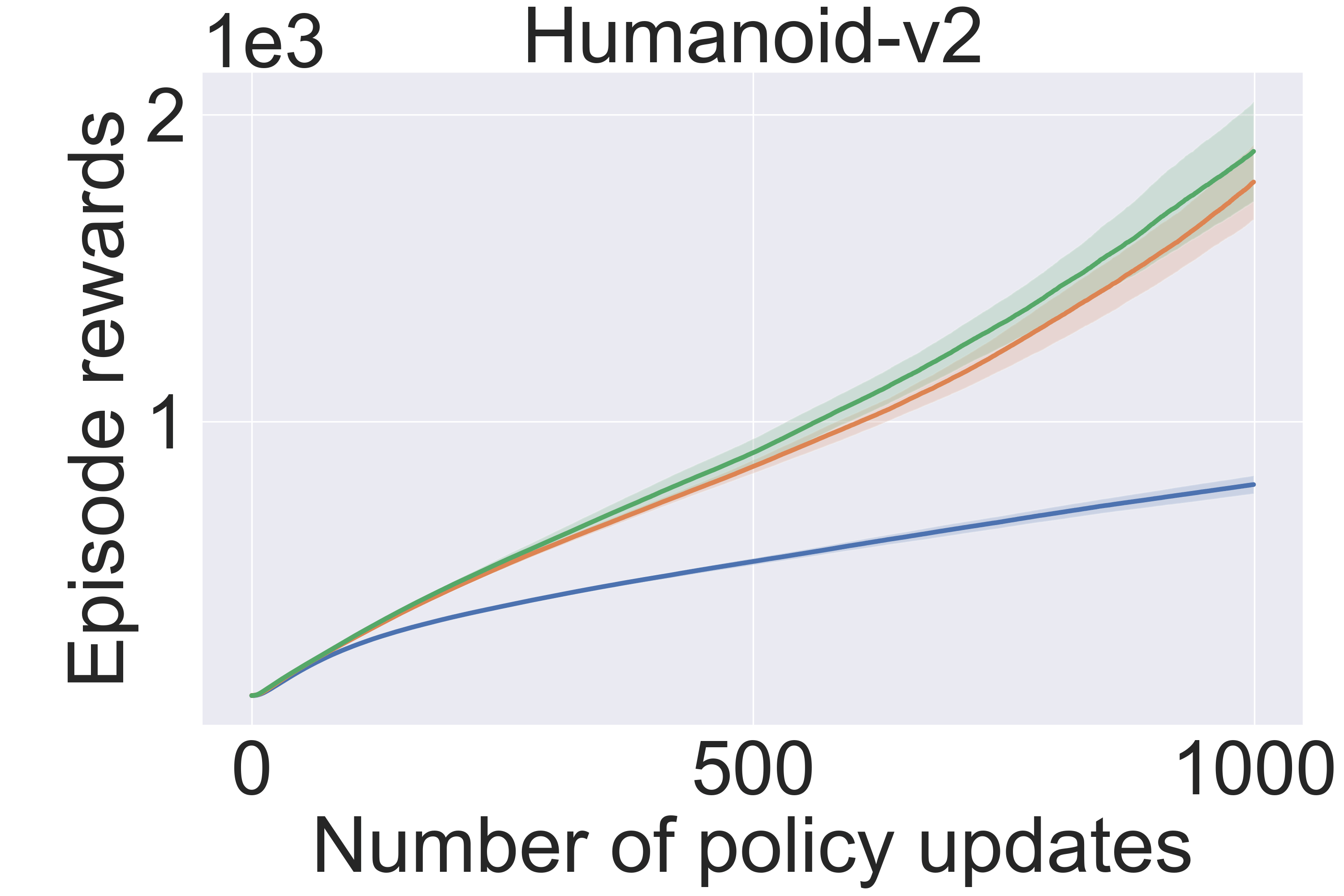}
    \end{subfigure}
    \hspace{-0.7em}
    \begin{subfigure}
  	\centering
  	\includegraphics[scale=\lineWW]{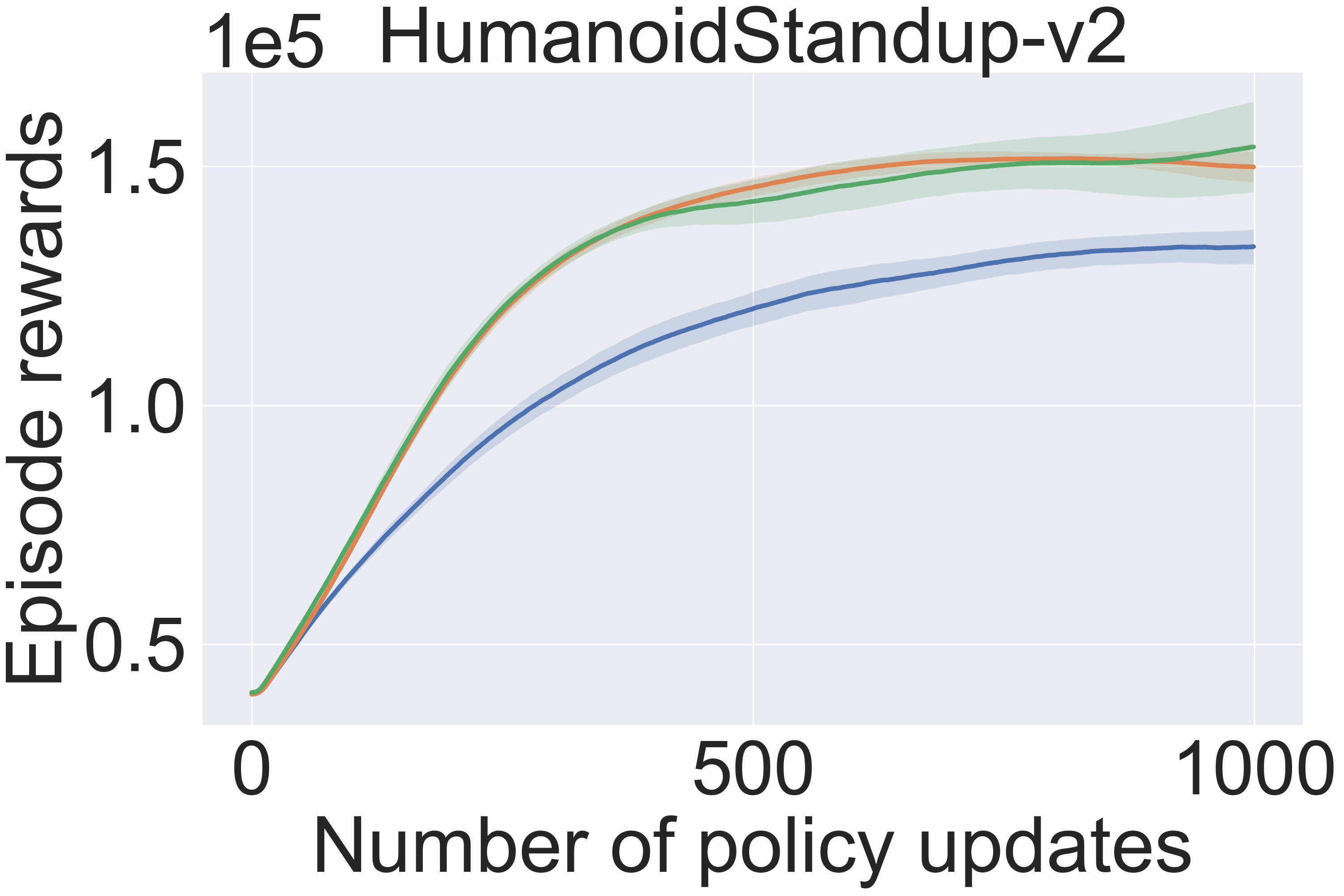}
    \end{subfigure}
    {\LARGE \TRPO}
    \begin{center}
    \line(1,0){475}
    \end{center}
\caption{The average performance ($10$ runs) of \BQ-based methods and \MC estimation in vanilla \PG, \NPG, and \TRPO frameworks across $4$ MuJoCo environments (refer Fig.~\ref{fig:appendix_all_comp_plots} in the supplement for comparison across all $7$ MuJoCo environments).
}
  \label{fig:all_comp_plots}
\end{figure*}
%
\textbf{Compatibility with Deep \PG Algorithms:}
We examine the compatibility of \BQ-based methods with the following on-policy deep policy gradient algorithms: (i) Vanilla policy gradient, (ii) natural policy gradient (NPG), and (iii) trust region policy optimization (\TRPO), as shown in~Fig.~\ref{fig:all_comp_plots}. In these experiments, only the \MC estimation subroutine is replaced with \BQ-based methods, keeping the rest of the algorithm unchanged. Note that for \TRPO, we use the \UAPG update of \NPG to compute the step direction.
We observe that \DBQPG consistently outperforms \MC estimation, both in final performance and sample complexity across all the deep \PG algorithms. This observation resonates with our previous finding of the superior gradient quality of \DBQPG estimates, and strongly advocates the use of \DBQPG over \MC for \PG estimation.

For \UAPG, we observe a similar trend as \DBQPG. The advantage of \UAPG estimates is more pronounced in the vanilla \PG, and \NPG experiments since the performance on these algorithms is highly sensitive to the choice of learning rates.
\UAPG adjusts the stepsize of each gradient component based on its uncertainty, resulting in a robust update in the face of uncertainty, a better sample complexity, and average return. We observe that \UAPG performs at least as good as, if not considerably better than, \DBQPG on most experiments. Since \TRPO updates are less sensitive to the learning rate, \UAPG adjustment does not provide a significant improvement over \DBQPG.
\section{Related Work}

The high sample complexity of \MC methods has been a long-standing problem in the \PG literature \citep{Rubinstein69}. Previous approaches that address this issue broadly focus on two aspects: (i) improving the quality of \PG estimation using a value function approximation \citep{Konda_2003}, or (ii) attaining faster convergence by robustly taking larger steps in the right direction. The former class of approaches trade a tolerable level of bias for designing a lower variance \PG estimator \citep{GAE,stone_BAC_kernel_learning}. Following the latter research direction, \citet{Kakade_2001_NPG} and \citet{kakade2002approximately} 
suggest replacing the vanilla \PG with natural policy gradient (\NPG), the steepest descent direction in the policy distribution space. While \NPG improves over vanilla \PG methods in terms of sample complexity \citep{Peters_2008,agarwal2019optimality}, it is just as vulnerable to catastrophic policy updates. Trust region policy optimization (\TRPO) \citep{TRPO} extends the \NPG algorithm with a robust stepsize selection mechanism that guarantees monotonic improvements for expected (true) policy updates.
However, the practical \TRPO algorithm loses its improvement guarantees for stochastic \PG estimates, thereby necessitating a large sample size for computing reliable policy updates. The advantages of these approaches, in terms of sample efficiency, are orthogonal to the benefits of \DBQPG and \UAPG methods.

Another line of research focuses on using Gaussian processes (\GP) to directly approximate the \PG integral \citep{BPG}. This work was followed by the Bayesian Actor-Critic (\BAC) algorithm \citep{BAC_1}, which exploits the MDP framework for improving the statistical efficiency of \PG estimation. Like \DBQPG, \BAC is a \BQ-based \PG method that uses a \GP to approximate the action-value function.
However, \BAC is an online algorithm that uses Gaussian process temporal difference (\GPTD) \citep{Engel_2005}, a sequential kernel sparsification method, for learning the value function. \BAC's sequential nature and a prohibitive $\mathcal{O}(m^2n + m^3)$ time and $\mathcal{O}(mn + m^2)$ storage complexity ($m$ is the dictionary size, i.e., the number of inducing points) prevents if from scaling to large non-linear policies and high-dimensional continuous domains.
Further, a recent extension of \BAC \citep{BAC_2} uses the uncertainty to adjust the learning rate of policy parameters, $(\mI - \frac{1}{\nu} \mC_{\vtheta}^{BQ}) \mL_{\vtheta}^{BQ}$ ($\nu$ is an upper bound for $\mC_{\vtheta}^{BQ}$), much like the \UAPG method. While this update reduces the step-size of more uncertain directions, it does not provide gradient estimates with uniform uncertainty, making it less robust to bad policy updates relative to \UAPG. Moreover, this method would not work for \NPG and \TRPO algorithms as their covariance is the inverse of an ill-conditioned matrix. Refer supplement Sec.~\ref{sec:appendix_BACvsDBQPG} for a more detailed comparison of \DBQPG and \UAPG with prior \BQ-\PG works.

\section{Discussion}
We study the problem of estimating accurate policy gradient (PG) estimates from a finite number of samples. This problem becomes relevant in numerous \RL applications where an agent needs to estimate the \PG using samples gathered through interaction with the environment. Monte-Carlo (\MC) methods are widely used for \PG estimation despite offering high-variance gradient estimates and a slow convergence rate. We propose \DBQPG, a high-dimensional generalization of Bayesian quadrature that, like \MC method, estimates \PG in linear-time. We empirically study \DBQPG and demonstrate its effectiveness over \MC methods.
We show that \DBQPG provides more accurate gradient estimates, along with a significantly smaller variance in the gradient estimation. Next, we show that replacing the \MC method with \DBQPG in the gradient estimation subroutine of three deep \PG algorithms, viz., Vanilla \PG, \NPG, and \TRPO, offers consistent gains in sample complexity and average return.
These results strongly recommend the substitution of \MC estimator with \DBQPG in deep \PG algorithms.

To obtain reliable policy updates from stochastic gradient estimates, one needs to estimate the uncertainty in gradient estimation, in addition to the gradient estimation itself. The proposed \DBQPG method additionally provides this uncertainty along with the \PG estimate. We propose \UAPG, a \PG method that uses \DBQPG's uncertainty to normalize the gradient components by their uncertainties, returning a uniformly uncertain gradient estimate. Such a normalization will lower the step size of gradient components with high uncertainties, resulting in reliable updates with robust step sizes. We show that \UAPG further improves the sample complexity over \DBQPG on Vanilla \PG, \NPG, and \TRPO algorithms. Overall, our study shows that it is possible to scale Bayesian quadrature to high-dimensional settings, while its better gradient estimation and well-calibrated gradient uncertainty can significantly boost the performance of deep \PG algorithms.



\section*{Ethics Statement}

When deploying deep policy gradient (\PG) algorithms for learning control policies in physical systems, sample efficiency becomes an important design criteria. In the past, numerous works have focused on improving the sample efficiency of \PG estimation through variance reduction, robust stepsize selection, etc. In this paper, we propose deep Bayesian quadrature policy gradient (\DBQPG), a statistically efficient policy gradient estimator that offers orthogonal benefits for improving the sample efficiency. In comparison to Monte-Carlo estimation, the default choice for \PG estimation, \DBQPG returns more accurate gradient estimates with much lower empirical variance. Since \DBQPG is a general gradient estimation subroutine, it can directly replace Monte-Carlo estimation in most policy gradient algorithms, as already demonstrated in our paper. Therefore, we think that the \DBQPG method directly benefits most policy gradient algorithms and is indirectly beneficial for several downstream reinforcement learning applications.

We also propose uncertainty aware policy gradient (\UAPG), a principled approach for incorporating the uncertainty in gradient estimation (also quantified by the \DBQPG method) to obtain reliable \PG estimates. \UAPG lowers the risk of catastrophic performance degradation with stochastic policy updates, and empirically performs at least as good as, if not better than, the \DBQPG method.
Hence, we believe that the \UAPG method is more relevant to reinforcement learning applications with safety considerations, such as robotics.



\bibliography{Ref}
\onecolumn
\newcommand{\lineWAppendix}{0.20}
\appendix
\section{Useful Identities}
Expectation of the score vector $\vu(z) = \nabla_{\vtheta} \log \pi_{\vtheta}(a|s)$ under the policy distribution $\pi_{\vtheta}(a|s)$ is $\mathbf{0}$:
\begin{align}
\label{eqn:appendix_logprob_expectiation}
\begin{split}
\E_{a \sim \pi_\vtheta(.|s)} \left[ \vu(z) \right] &= \E_{a \sim \pi_{\vtheta}(.|s)} \left[ \nabla_{\vtheta} \log \pi_{\vtheta}(a|s) \right] = \int \pi_{\vtheta}(a|s) \nabla_{\vtheta} \log \pi_{\vtheta}(a|s) da
\\
&= \int \pi_{\vtheta}(a|s) \frac{\nabla_{\vtheta} \pi_{\vtheta}(a|s)}{\pi_{\vtheta}(a|s)} da = \int \nabla_{\vtheta} \pi_{\vtheta}(a|s) da
\\
&= \nabla_{\vtheta} \left( \int \pi_{\vtheta}(a|s) da \right) = \nabla_{\vtheta} (1) = \mathbf{0}
\end{split}
\end{align}
From Eq.~\ref{eqn:appendix_logprob_expectiation}, the expectation of the Fisher kernel $k_f$ under the policy distribution $\pi_{\vtheta}(a|s)$ is also $0$:
\begin{align}
\label{eqn:appendix_fisher_expectation}
\begin{split}
\E_{a \sim \pi_\vtheta(.|s)} \left[ k_f(z,z') \right] &= \E_{a \sim \pi_\vtheta(.|s)} \left[ \vu(z)^\top \mG^{-1} \vu(z') \right] = \E_{a \sim \pi_\vtheta(.|s)} \left[ \vu(z)^\top \right]  \mG^{-1} \vu(z') = 0
\end{split}
\end{align}

\section{Bayesian Quadrature for Estimating Policy Gradient Integral}
\label{appendix_bq_pg_closedform}
Bayesian quadrature (\BQ) \citep{O'Hagan1991} provides the required machinery for estimating the numerical integration in \PG (Eq.~\ref{eqn:policy_gradient_theorem}), by using a Gaussian process (\GP) function approximation for the action-value function $Q_{\pi_\vtheta}$. More specifically, we choose a zero mean \GP, i.e., $\Egauss\left[ Q_{\pi_\vtheta}(z) \right] = 0$,
with a prior covariance function $k(z_p,z_q) = \Cov [Q_{\pi_\vtheta}(z_p),Q_{\pi_\vtheta}(z_q)]$ and an additive Gaussian noise with variance $\sigma^2$.
One benefit of this prior is that the joint distribution over any finite number of action-values (indexed by the state-action inputs, $z \in \mathcal{Z}$) is also Gaussian:
\begin{align}
\label{eqn:appendix_Q_function_prior}
\begin{split}
\mQ_{\pi_\vtheta} &= [Q_{\pi_\vtheta}(z_1),...,Q_{\pi_\vtheta}(z_n)] \sim \mathcal{N}(0,\mK),
\end{split}
\end{align}
where $\mK$ is the Gram matrix with entries $\mK_{p,q} = k(z_p,z_q)$. This \GP prior is conditioned on the observed samples $\mathcal{D} = \{z_i\}_{i=1}^{n}$ drawn from $\rho^{\pi_\theta}$ to obtain the posterior moments of $Q_{\pi_\vtheta}$: 
\begin{align}
\label{eqn:appendix_Q_function_posterior}
\begin{split}
\Egauss\left[ Q_{\pi_\vtheta}(z)| \mathcal{D} \right] &= \vk(z)^\top (\mK +\sigma^2\mI)^{-1} \Qmc,
\\
\Cov\left[ Q_{\pi_\vtheta}(z_1), Q_{\pi_\vtheta}(z_2)| \mathcal{D} \right] &= k  (z_1,z_2) - \vk(z_1)^\top (\mK + \sigma^2\mI)^{-1} \vk(z_2),
\\
\text{where } \vk(z) = [k (z_1,z &), ..., k(z_n, z)], \; \mK = [\vk(z_1), ..., \vk(z_n)].
\end{split}
\end{align}
%
%
%
%
%
%
Since the transformation from $Q_{\pi_\vtheta}(z)$ to $\nabla_{\vtheta} J(\vtheta)$ happens through a linear integral operator (Eq.~\ref{eqn:policy_gradient_theorem}), the posterior moments of $Q_{\pi_\vtheta}$ can be used to compute a Gaussian approximation of $\nabla_{\vtheta} J(\vtheta)$:
\begin{align}
\label{eqn:appendix_BayesianQuadratureIntegral}
    & \quad \qquad \displaystyle \mL_{\vtheta}^{BQ} = \Egauss \left[ \nabla_{\vtheta} J(\vtheta) | \mathcal{D} \right] =\displaystyle \int \rho^{\pi_\vtheta}(z) \vu(z) \Egauss \left[ Q_{\pi_\vtheta}(z) | \mathcal{D} \right] dz \nonumber
    \\
    &  \quad \qquad \qquad = \displaystyle \left( \int \rho^{\pi_\vtheta}(z) \vu(z) \vk(z)^\top dz\right) (\mK +\sigma^2\mI)^{-1}\Qmc
    \\
    \displaystyle \mC_{\vtheta}^{BQ} &= \Cov [ \nabla_{\vtheta} J( \vtheta) | \mathcal{D}] = \displaystyle \int \rho^{\pi_\vtheta}(z_1) \rho^{\pi_\vtheta}(z_2) \vu(z_1) \Cov[ Q_{\pi_\vtheta}(z_1), Q_{\pi_\vtheta}(z_2)| \mathcal{D}] \vu(z_2)^\top dz_1 dz_2,\nonumber
    \\
    &= \displaystyle \int \rho^{\pi_\vtheta}(z_1) \rho^{\pi_\vtheta}(z_2) \vu(z_1) \Big( k(z_1,z_2) - \vk(z_1)^\top (\mK +\sigma^2\mI)^{-1} \vk(z_2) \Big) \vu(z_2)^\top dz_1 dz_2,\nonumber
\end{align}
where the posterior mean $\mL_{\vtheta}^{BQ}$ can be interpreted as the \PG estimate and the posterior covariance $\mC_{\vtheta}^{BQ}$ quantifies the uncertainty in the \PG estimate. However, the integrals in  Eq.~\ref{eqn:appendix_BayesianQuadratureIntegral} cannot be solved analytically for an arbitrary \GP prior covariance function $k(z_1, z_2)$. \citet{BAC_1} showed that these integrals have analytical solutions when the \GP kernel $k$ is the additive composition of an arbitrary state kernel $k_s$ and the (indispensable) Fisher kernel $k_f$:
\begin{align}
\label{eqn:appendix_kernel_definitions}
    k(z_1,z_2) = c_1 k_s(s_1,s_2) + c_2 k_f(z_1,z_2), \quad k_f(z_1,z_2) &= \vu(z_1)^\top \mG^{-1} \vu(z_2),
\end{align}

where $c_1, c_2$ are hyperparameters and $\mG$ is the Fisher information matrix of the policy $\pi_\vtheta$. Using the following definitions,
\begin{small}
\begin{eqnarray}
\label{eqn:appendix_kernel_equations}
\vk_f(z) = \mU^\top\mG^{-1} \vu(z), \ \ \mK_f = \mU^\top \mG^{-1} \mU, \ \ \mK = c_1 \mK_s + c_2 \mK_f, \ \ 
\mG = \E_{z \sim \rho^{\pi_\vtheta}} [\vu(z) \vu(z)^\top] \approx \frac{1}{n} \mU\mU^\top,
\end{eqnarray}
\end{small}
\hspace{-8pt}
the closed-form expressions for \PG estimate $\mL_{\vtheta}^{BQ}$ and its uncertainty $\mC_{\vtheta}^{BQ}$ are obtained as follows,

\DeclareFontFamily{U}{stixscr}{}
\DeclareFontShape{U}{stixscr}{m}{n}{<-> s*[0.9] stix-mathscr}{}
\DeclareRobustCommand{\pointright}{%
  \mathrel{\text{\usefont{U}{stixscr}{m}{n}\symbol{"D2}}}%
}

\begin{align}
\label{eqn:appendix_pg_mean_closedform}
\begin{split}
    \displaystyle \mL_{\vtheta}^{BQ} &= \Egauss \left[ \nabla_{\vtheta} J(\vtheta) | \mathcal{D} \right] = \displaystyle \int \rho^{\pi_\vtheta}(z) \vu(z) \Egauss \left[ Q_{\pi_\vtheta}(z) | \mathcal{D} \right] dz
    \\
    &= \E_{z \sim \rho^{\pi_\vtheta}} \left[ \vu(z) \Egauss [Q_{\pi_\vtheta}(z) | \mathcal{D}] \right] = \E_{z \sim \rho^{\pi_\vtheta}} \left[ \vu(z) \vk(z)^\top \right] (\mK +\sigma^2\mI)^{-1}\Qmc
    \\
    &= \left( c_1 \E_{z \sim \rho^{\pi_\vtheta}} \left[ \vu(z) \vk_s(s)^\top \right] + c_2 \E_{z \sim \rho^{\pi_\vtheta}} \left[ \vu(z) \vk_f(z)^\top \right] \right) (\mK +\sigma^2\mI)^{-1}\Qmc
    \\
    & \;\;\;\;\; \text{$\pointright$ $\vk_s$ term disappears since the integral of $\vu(z)$ over action-dims is 0 from Eq.~\ref{eqn:appendix_logprob_expectiation}}
    \\
    &= c_2 \E_{z \sim \rho^{\pi_\vtheta}} \left[ \vu(z) \vk_f(z)^\top \right] (\mK +\sigma^2\mI)^{-1}\Qmc
    \\
    &= c_2 \E_{z \sim \rho^{\pi_\vtheta}} \left[ \vu(z) \vu(z)^\top \right] \mG^{-1} \mU (\mK +\sigma^2\mI)^{-1}\Qmc \quad \text{$\pointright$ from $\vk_f$ definition in Eq.~\ref{eqn:appendix_kernel_equations}}
    \\
    &= c_2 \mG \: \mG^{-1} \mU (c_1\mK_s + c_2\mK_f +\sigma^2\mI)^{-1}\Qmc \qquad \qquad \text{$\pointright$ from $\mG$ \;definition in Eq.~\ref{eqn:appendix_kernel_equations}}
    \\
    &= c_2 \mU (c_1\mK_s + c_2\mK_f +\sigma^2\mI)^{-1}\Qmc
\end{split}
\end{align}\footnotetext{In Eq.~\ref{eqn:appendix_pg_mean_closedform} and \ref{eqn:appendix_pg_cov_closedform}, the following state kernel $k_s$ terms vanish, as an extension to the identity in Eq.~\ref{eqn:appendix_logprob_expectiation}: $\E_{a_1 \sim \pi_\vtheta(.|s_1)} \left[ k_s(s_1,s_2) \vu(z_1) \right] = \mathbf{0}$ and $\E_{a_1 \sim \pi_\vtheta(.|s_1)} \left[ \vu(z_1) \vk_s^\top(s_1) \right] = \mathbf{0}$.}

\begin{align}
\label{eqn:appendix_pg_cov_closedform}
\begin{split}
    \displaystyle \mC_{\vtheta}^{BQ} &= \Cov \left[ \nabla_{\vtheta} J(\vtheta) | \mathcal{D} \right] = \displaystyle \int dz_1 dz_2 \rho^{\pi_\vtheta}(z_1) \rho^{\pi_\vtheta}(z_2) \vu(z_1) \Cov\left[ Q_{\pi_\vtheta}(z_1), Q_{\pi_\vtheta}(z_2)| \mathcal{D} \right] \vu(z_2)^\top
    \\
    &= \E_{z_1,z_2 \sim \rho^{\pi_\vtheta}} \left[ \vu(z_1) \Cov\left[ Q_{\pi_\vtheta}(z_1), Q_{\pi_\vtheta}(z_2)| \mathcal{D} \right] \vu(z_2)^\top \right]
    \\
    &= \E_{z_1,z_2 \sim \rho^{\pi_\vtheta}} \left[ \vu(z_1) \Big( k(z_1,z_2) - \vk(z_1)^\top (\mK +\sigma^2\mI)^{-1} \vk(z_2) \Big) \vu(z_2)^\top \right]
    \\
    & \qquad \qquad \;\;\;\; \text{$\pointright$ $k_s$ terms disappear since the integral of $\vu(z)$ over action-dims is 0 (Eq.~\ref{eqn:appendix_logprob_expectiation})}
    \\
    &= \E_{z_1,z_2 \sim \rho^{\pi_\vtheta}} \left[ \vu(z_1) \Big( c_2 k_f(z_1,z_2) - c_2^2 \vk_f(z_1)^\top (\mK +\sigma^2\mI)^{-1} \vk_f(z_2) \Big) \vu(z_2)^\top \right]
    \\
    & \qquad \qquad \qquad \qquad \qquad \; \text{$\pointright$ from the definitions of $k_f(z_1,z_2)$ (Eq.~\ref{eqn:appendix_kernel_definitions}) and $\vk_f(z)$ (Eq.~\ref{eqn:appendix_kernel_equations})}
    \\
    &= \E_{z_1,z_2 \sim \rho^{\pi_\vtheta}} \left[ \vu(z_1) \vu(z_1)^\top \Big( c_2 \mG^{-1} - c_2^2 \mG^{-1}\mU (\mK +\sigma^2\mI)^{-1} \mU^\top\mG^{-1} \Big) \vu(z_2) \vu(z_2)^\top \right]
    \\
    &= \E_{z_1 \sim \rho^{\pi_\vtheta}} [\vu(z_1) \vu(z_1)^\top] \Big( c_2 \mG^{-1}
    - c_2^2 \mG^{-1}\mU (\mK +\sigma^2\mI)^{-1} \mU^\top\mG^{-1} \Big) \E_{z_2 \sim \rho^{\pi_\vtheta}} [\vu(z_2) \vu(z_2)^\top]
    \\
    &= \mG \Big( c_2 \mG^{-1} - c_2^2 \mG^{-1}\mU (\mK +\sigma^2\mI)^{-1} \mU^\top\mG^{-1} \Big) \mG
    \\
    &= c_2 \mG - c_2^2\mU (c_1\mK_s + c_2\mK_f +\sigma^2\mI)^{-1} \mU^\top
\end{split}
\end{align}
Further, the inverse of $\mC_{\vtheta}^{BQ}$ can also be found analytically using the \citet{woodbury} identity:
\begin{align}
\begin{split}
    \displaystyle \left(\mC_{\vtheta}^{BQ}\right)^{-1} &= \left( c_2 \mG - c_2^2\mU (c_1\mK_s + c_2\mK_f +\sigma^2\mI)^{-1} \mU^\top \right)^{-1}
    \\
    &= \frac{1}{c_2}\mG^{-1} + \mG^{-1}\mU \left(c_1\mK_s + c_2\mK_f +\sigma^2\mI -c_2 \mU^\top \mG^{-1} \mU \right)^{-1}  \mU^\top \mG^{-1}
    \\
    &= \frac{1}{c_2}\mG^{-1} + \mG^{-1}\mU \left(c_1\mK_s + \sigma^2\mI \right)^{-1} \mU^\top \mG^{-1}
\end{split}
\end{align}
Thus, by choosing the overall kernel as a additive composition of the Fisher kernel $k_f$ and an arbitrary state kernel $k_s$, the \BQ approach\footnote{Different from the proof provided in \citet{BAC_1}, our derivation of \BQ-\PG substitutes the sequential computation of \GP posterior \citep{Engel_2003} with parallel computation over a batch of samples. This puts \BQ-\PG in the context of contemporary deep \PG algorithms.} has a closed-form expression for the gradient mean $\mL_{\vtheta}^{BQ}$ and its uncertainty $\mC_{\vtheta}^{BQ}$ (gradient covariance).

\section{More Intuition Behind the \GP Kernel Choice}
\label{sec:appendix_intuition_gp_kernel_choice}
In this section, we analyze the implications of the specified \GP kernel choice that provides an analytical solution to the \PG integral (shown in the previous section).
Particularly, we study the affect of the additive composition of a state kernel $k_s$ and the Fisher kernel $k_f$ on (i) modeling the $Q_{\pi_{\vtheta}}$ function, and consequently (ii) the policy gradient.

\subsection{Posterior Moments of the Value Functions}
\label{appendix_state_value_advantage}
Here, we split $Q_{\pi_{\vtheta}}$ function as the sum of a state-value function $V_{\pi_{\vtheta}}$ and advantage function $A_{\pi_{\vtheta}}$:
\begin{equation}
    Q_{\pi_{\vtheta}}(z_t) =  V_{\pi_{\vtheta}}(s_t) + A_{\pi_{\vtheta}}(z_t), \quad V_{\pi_{\vtheta}}(s_t) = \E_{a_t \sim \pi_{\vtheta}(\cdot|s_t)} \big[Q_{\pi_{\vtheta}}(z_t) \big],
\end{equation}
where $V_{\pi_{\vtheta}}(s)$ is the expected return under $\pi_{\vtheta}$ from the initial state $s$, and $A_{\pi_{\vtheta}}(z)$ denotes the advantage (or disadvantage) of picking a particular initial action $a$ relative to the policy's prediction $a \sim \pi_{\vtheta}$.
The specified \GP kernel choice for modeling the $Q_{\pi_\vtheta}$ function approximates $V_{\pi_\vtheta}$ and $A_{\pi_\vtheta}$ as follows:
\begin{align}
\begin{split}
\label{eqn:appendix_value_posterior_1}
\Egauss\left[ V_{\pi_\vtheta}(s)|\mathcal{D} \right] &= \E_{a \sim \pi_\vtheta(.|s)}\left[ \Egauss\left[ Q_{\pi_\vtheta}(z)|\mathcal{D} \right] \right] = \E_{a \sim \pi_\vtheta(.|s)}\left[ \vk(z)^\top \right] (\mK +\sigma^2\mI)^{-1} \Qmc
\\
&= \E_{a \sim \pi_\vtheta(.|s)}\left[ \left( c_1 \vk_s(s) + c_2 \vk_f(z)\right)^\top \right] (\mK +\sigma^2\mI)^{-1} \Qmc
\\
&= \left( c_1 \vk_s(s) + c_2 \E_{a \sim \pi_\vtheta(.|s)}\left[ \vk_f(z) \right] \right)^\top (\mK +\sigma^2\mI)^{-1} \Qmc
\\
&= c_1 \vk_s(s)^\top (c_1\mK_s + c_2\mK_f +\sigma^2\mI)^{-1} \Qmc
\end{split}
\end{align}
\begin{align}
\begin{split}
\label{eqn:appendix_adv_posterior_1}
\Egauss\left[ A_{\pi_\vtheta}(z)|\mathcal{D} \right] &= \Egauss\left[ \left(Q_{\pi_\vtheta}(z) - V_{\pi_\vtheta}(s) \right) |\mathcal{D} \right] = \Egauss\left[ Q_{\pi_\vtheta}(z)|\mathcal{D} \right] - \Egauss\left[ V_{\pi_\vtheta}(s)|\mathcal{D} \right]
\\
&= \left(\vk(z) - c_1 \vk_s(s) \right)^\top (c_1 \mK_s + c_2 \mK_f +\sigma^2\mI)^{-1} \Qmc
\\
&= c_2 \vk_f(z)^\top (c_1 \mK_s + c_2 \mK_f +\sigma^2\mI)^{-1} \Qmc
\end{split}
\end{align}
\begin{align}
\begin{split}
\Cov\left[ V_{\pi_\vtheta}(s_1), Q_{\pi_\vtheta}(z_2)| \mathcal{D} \right] &= \E_{a_1 \sim \pi_\vtheta(.|s_1)}\left[ \Cov\left[ Q_{\pi_\vtheta}(z_1), Q_{\pi_\vtheta}(z_2)| \mathcal{D} \right] \right],
\\
&=  \E_{a_1 \sim \pi_\vtheta(.|s_1)}\left[ k(z_1,z_2) - \vk(z_1)^\top(\mK+\sigma^2\mI)^{-1}\vk(z_2) \right]
\\
&= c_1 k_s(s_1,s_2) - c_1 \vk_s(s_1)^\top (c_1\mK_s + c_2\mK_f +\sigma^2\mI)^{-1} \vk(z_2)
\end{split}
\end{align}
\begin{align}
\begin{split}
\label{eqn:appendix_value_posterior_2}
\Cov\left[ V_{\pi_\vtheta}(s_1), V_{\pi_\vtheta}(s_2)| \mathcal{D} \right] &= \E_{a_2 \sim \pi_\vtheta(.|s_2)}\left[ \Cov\left[ V_{\pi_\vtheta}(s_1), Q_{\pi_\vtheta}(z_2)| \mathcal{D} \right] \right]
\\
&= c_1 k_s(s_1,s_2) - c_1 \vk_s(s_1)^\top (\mK +\sigma^2\mI)^{-1} \E_{a_2 \sim \pi_\vtheta(.|s_2)}\left[\vk(z_2) \right]
\\
&= c_1 k_s(s_1,s_2) - c_1^2 \vk_s(s_1)^\top (c_1\mK_s + c_2\mK_f +\sigma^2\mI)^{-1} \vk_s(s_2)
\end{split}
\end{align}
\begin{align}
\begin{split}
\Cov\left[ A_{\pi_\vtheta}(z_1), Q_{\pi_\vtheta}(z_2)| \mathcal{D} \right] &= \Cov\left[ Q_{\pi_\vtheta}(z_1) - V_{\pi_\vtheta}(s_1), Q_{\pi_\vtheta}(z_2)| \mathcal{D} \right]
\\
&= c_2 k_f(z_1,z_2) - c_2 \vk_f(z_1)^\top (c_1\mK_s + c_2\mK_f +\sigma^2\mI)^{-1} \vk(z_2)
\end{split}
\end{align}
\begin{align}
\begin{split}
\label{eqn:appendix_adv_posterior_2}
\Cov\left[ A_{\pi_\vtheta}(z_1), A_{\pi_\vtheta}(z_2)| \mathcal{D} \right] &= \Cov\left[ A_{\pi_\vtheta}(z_1), Q_{\pi_\vtheta}(z_2)| \mathcal{D} \right] - \E_{a_2 \sim \pi_\vtheta(.|s_2)}\left[ \Cov\left[ A_{\pi_\vtheta}(z_1), Q_{\pi_\vtheta}(z_2)| \mathcal{D}  \right] \right]
\\
&= c_2 k_f(z_1,z_2) - c_2^2 \vk_f(z_1)^\top (c_1\mK_s + c_2\mK_f +\sigma^2\mI)^{-1} \vk_f(z_2).
\end{split}
\end{align}
Note that the posterior moments of both $V_{\pi_\vtheta}$ and $A_{\pi_\vtheta}$ are dependent on the state $k_s$ and Fisher $k_f$ kernels. However, only the posterior moments of $V_{\pi_\vtheta}$ vanish upon removing the state kernel $k_s$ (set $c_1 = 0$ in Eq.~\ref{eqn:appendix_value_posterior_1}, \ref{eqn:appendix_value_posterior_2}), while removing the Fisher kernel $k_f$ causes the posterior moments of $A_{\pi_\vtheta}$ to vanish (set $c_2 = 0$ in Eq.~\ref{eqn:appendix_adv_posterior_1}, \ref{eqn:appendix_adv_posterior_2}). Thus, choosing the overall kernel as the additive composition of a state kernel $k_s$ and the Fisher kernel $k_f$ implicitly divides the $Q_{\pi_\vtheta}$ function into state-value function $V_{\pi_\vtheta}$ and advantage function $A_{\pi_\vtheta}$, separately modeled by $k_s$ and $k_f$ respectively.

\subsection{\MC Estimation as a Degenerate Case of \BQ}
\label{appendix:MCPG_degenerateofBQ}
The closed-form \BQ-\PG expression $\mL_{\vtheta}^{BQ}$ and the \MC-\PG expression $\mL_{\vtheta}^{MC}$ (Eq.~\ref{eqn:Monte-Carlo_PG} in the main paper) are surprisingly similar, except for subtle differences that arise from the \GP kernel $k$ choice. This observation calls for a comparative analysis of \BQ-\PG with respect to the \MC-\PG baseline.

Removing Fisher kernel ($c_2 = 0$ in Eq.~\ref{eqn:appendix_pg_mean_closedform}) suppresses the advantage function, making \PG moments  $\mL_{\vtheta}^{BQ} = 0$ and $\mC_{\vtheta}^{BQ} = 0$. Further, removing the state kernel ($c_1 = 0$ in Eq.~\ref{eqn:appendix_pg_mean_closedform}), reduces the \BQ-\PG $\mL_{\vtheta}^{BQ}$ to \MC-\PG $\mL_{\vtheta}^{MC}$:
\begin{align}
\begin{split}
    \displaystyle \mL_{\vtheta}^{BQ} \bigl\vert_{c_1 = 0} 
    &= \displaystyle c_2 \mU (c_2 \mK_f +\sigma^2\mI)^{-1}\Qmc
    \\
    &= \displaystyle c_2 \mU (c_2 \mU^\top \mG^{-1} \mU +\sigma^2\mI)^{-1}\Qmc \qquad \qquad \;\; \text{$\pointright$ using the $\mK_f$ definition from Eq.~\ref{eqn:appendix_kernel_equations}}
    \\
    &= c_2 \left(\frac{1}{\sigma^2} \mU - \frac{c_2}{\sigma^4} \mU \mU^\top \left( \mG + \frac{c_2}{\sigma^2} \mU \mU^\top \right)^{-1} \mU \right) \Qmc
    \\
    & \qquad \qquad \qquad \qquad \qquad \text{$\pointright$ applying the \citet{woodbury} matrix (inversion) identity}
    \\
    &= \frac{c_2}{\sigma^2} \left( \mU - \frac{c_2 n}{\sigma^2} \mG \left( \mG + \frac{c_2 n}{\sigma^2} \mG \right)^{-1} \mU \right) \Qmc \;\; \text{$\pointright$ using the $\mG$ definition from Eq.~\ref{eqn:appendix_kernel_equations}}
    \\
    &= \frac{c_2}{\sigma^2} \left(\mU - \frac{c_2 n}{(\sigma^2+c_2 n)} \mU \right) \Qmc
    \\
    &= \frac{c_2}{\sigma^2+c_2 n}\mU \Qmc
\end{split}
\end{align}
Thus, \MC estimation is a limiting case of \BQ when the state kernel $k_s$ vanishes, i.e., the posterior distributions over the state-value function $V_{\pi_\vtheta}$ becomes non-existent (for $c_1=0$ in Eq.~\ref{eqn:appendix_value_posterior_1}, \ref{eqn:appendix_value_posterior_2}).Looking at this observation backwards, \BQ-\PG with a prior state kernel $k_s = 0$ (or equivalently \MC-\PG) is incapable of modeling the state-value function, which vastly limits the \GP's expressive power for approximating the $Q_{\pi_{\vtheta}}$ function, and consequently the \PG estimation.
Alternatively, \BQ-\PG can offer more accurate gradient estimates than \MC-\PG when the state kernel $k_s$ captures a meaningful prior with respect to the MDP's state-value function.
This observation is consistent with previous works \citep{briol2015frank,kanagawa2016convergence,kanagawa2020convergence} that prove a strictly faster convergence rate of \BQ over \MC, under mild regularity assumptions.
Further, the posterior covariance $\mC_{\vtheta}^{BQ}$ becomes a scalar multiple of the prior covariance $c_2 \mG$ (or the F.I.M $\mG$):
\begin{align}
\begin{split}
    \displaystyle \mC_{\vtheta}^{BQ} \bigl\vert_{c_1 = 0}
    &= c_2 \mG - c_2^2 \mU \left(c_2 \mK_f + \sigma^2 \mI \right)^{-1} \mU^\top
    \\
    &= c_2 \mG - c_2^2 \mU \left(c_2 \mU^\top \mG^{-1} \mU + \sigma^2 \mI \right)^{-1} \mU^\top
    \\
    &= c_2 \mG - \frac{c_2^2}{\sigma^2} \left(\mU - \frac{c_2 n}{(\sigma^2+c_2 n)} \mU \right) \mU^\top
    \\
    & \qquad \text{$\pointright$ applying the \citet{woodbury} identity, similar to $\mL_{\vtheta}^{BQ}$ proof}
    \\
    &= c_2 \mG - \frac{c_2^2 n}{(\sigma^2+c_2 n)} \mG \qquad \;\; \text{$\pointright$ using the $\mG$ definition from Eq.~\ref{eqn:appendix_kernel_equations}}
    \\
    &= \frac{\sigma^2 c_2}{\sigma^2+c_2 n} \mG
\end{split}
\end{align}

\section{Scaling \BQ to High-Dimensional Settings}
\label{appendix:efficient_DBQPG_implementation}
In comparison to \MC methods, \BQ approaches have several appealing properties, such as a strictly faster convergence rate \citep{briol2015frank,kanagawa2016convergence,kanagawa2020convergence} and a logical propagation of numerical uncertainty from the action-value $Q_{\pi_\vtheta}$ function space to the policy gradient estimation. However, the complexity of estimating \BQ's posterior moments, $\mL_{\vtheta}^{BQ}$ and $\mC_{\vtheta}^{BQ}$, is largely influenced by the expensive matrix-inversion operation $(\mK +\sigma^2\mI)^{-1}$, that scales with an $\mathcal{O}(n^3)$ time and $\mathcal{O}(n^2)$ storage complexity ($n$ is the sample size).
In the following, we provide a detailed description of the \DBQPG method (Sec.~\ref{sec:DBQPG_algorithm} in the main paper) that allows us to scale \BQ to high-dimensional settings, while retaining the superior statistical efficiency over \MC methods.

While it is expensive to compute the exact matrix inversion, all we need is to compute $(\mK +\sigma^2\mI)^{-1}\Qmc$, an inverse matrix-vector multiplication (i-MVM) operation, that can be efficiently implemented using the conjugate gradient (CG) algorithm.
Using the CG algorithm, the i-MVM operation can be computed implicitly, i.e., without the explicit storage or inversion of a full-sized matrix, by simply following iterative routines of efficient MVMs. The computational complexity of solving $(\mK +\sigma^2\mI)^{-1} \mQ$ with $p$ iterations of CG is $\mathcal{O}(p \mathcal{M})$, where $\mathcal{M}$ is the computational complexity associated with the MVM computation $\mK \mQ$. One of the appealing properties of the CG algorithm is that it often converges within machine precision after $p \ll n$ iterations. However, naively computing $\mK \mQ = c_1 \mK_s \mQ + c_2 \mK_f \mQ$ still has a prohibitive $\mathcal{O}(n^2)$ time and storage complexity. We propose separate strategies for efficiently computing $\mK_s \mQ$ and $\mK_f \mQ$ MVMs.

\subsection{Efficient MVM Computation with Fisher Covariance Matrix}
\label{sec:appendix_Fisher_mvm}
The Fisher covariance matrix $\mK_f$ of a policy $\pi_\vtheta$ can be factorized as the product of three matrices, $\mU^\top \mG^{-1} \mU$, which leads to two distinct ways of efficiently implementing $\mK_f \mQ$. 

\textbf{Approach 1:} Observe that the $\mU$ matrix is the Jacobian (transposed) of the log-probabilities and $\mG$ matrix is the hessian of the KL divergence, with respect to policy parameters $\vtheta$. As a result, $\mK_f \mQ$ can be directly computed sequentially using three MVM routines: (i) a vector-Jacobian product ($vJp$) involving the $\mU$ matrix, followed by (ii) an inverse-Hessian-vector product (i-$Hvp$) involving the $\mG$ matrix, and finally (iii) a Jacobian-vector product ($Jvp$) involving the $\mU$ matrix.
\begin{align}
\label{eqn:appendix_fisher_kernel_vector_product}
\begin{split}
    \mK_f \mQ = \Big( \mU^\top \big( \mG^{-1} &( \mU \mQ ) \big)\Big) = \left( \frac{\partial \mathcal{L}}{\partial \vtheta} \left( \mG^{-1} \left( \left( \frac{\partial \mathcal{L}}{\partial \vtheta} \right)^\top \mQ \right)\right)\right),
    \\
    \text{where } \mathcal{L} = [\log \pi_{\vtheta}(a_1 | & s_1), ..., \log \pi_{\vtheta}(a_n | s_n)], \text{ } (s_i,a_i) \sim \rho^{\pi_\vtheta} \text{ } \forall i \in [1,n].
\end{split}
\end{align}
Most standard automatic differentiation (AD) and neural network packages support $vJp$
and $Hvp$. CG algorithm can be used to compute the i-$Hvp$ from $Hvp$ \citep{TRPO}.
A $Jvp$ can be computed using a single forward-mode AD operation or two reverse-mode AD operations \citep{jvp}.
While $\mK_f \mQ$ has a linear complexity in sample size $n$, the numerous AD calls makes this procedure slightly slower.

\textbf{Approach 2 (faster):} Observe that the $n \times n$ dimensional matrix $\mK_f = \mU^\top \mG^{-1} \mU$ has a rank $|\Theta| < n$ (since $\mU$ has the dimensions $|\Theta| \times n$). To efficiently compute $\mK_f \mQ$, it helps to first visualize the $\mU$ matrix in terms of its singular value decomposition (SVD), $\mU = \mP \mLambda \mR^\top$, where $\mP$ and $\mR$ are orthogonal matrices with dimensions $|\Theta| \times |\Theta|$ and $n \times |\Theta|$ respectively, and $\mLambda$ is an $|\Theta| \times |\Theta|$ diagonal matrix of singular values. Accordingly, $\mG$ and $\mK_f$ expressions can be simplified as:
\begin{align}
\begin{split}
    \mG &= \frac{1}{n} \mU\mU^\top = \frac{1}{n} \mP \mLambda^2 \mP^\top, \;\;\; \mK_f = \mU^\top \mG^{-1} \mU = n \mR \mLambda \mP^\top \left( \mP \mLambda^{-2} \mP^\top \right) \mP \mLambda \mR^\top = n \mR \mR^\top.
\end{split}
\end{align}
In practice, we avoid the computational overhead of a full-rank SVD by using randomized truncated SVD \cite{randomizedSVD} to compute the rank $\delta \ll |\Theta|$ approximations for $\mP$, $\mLambda$ and $\mR$, i.e. $|\Theta| \times \delta$, $\delta \times \delta$ and $n \times \delta$ dimensional matrices respectively. Further, the fast SVD of the $\mU$ matrix can be computed using an iterative routine of implicit MVM computations, thus, avoiding the explicit formation and storage of the $\mU$ matrix at any point of time.
The implicit low-rank nature of the linear kernel provides an efficient MVM for $\mK_f$ in $\mathcal{O}(n \delta)$ time and space complexity.

\subsection{Efficient MVM Computation with State Covariance Matrix}
\label{sec:appendix_state_mvm}

Unlike the fixed Fisher kernel, the choice of the state kernel $k_s$ is arbitrary, and thus, requires a general method for fast $\mK_s\mQ$ computation. We rely on \textit{structured kernel interpolation (SKI)} \citep{kissgp}, a general inducing point framework for linear-time MVM computation with arbitrary kernels. Using $m$ inducing points $\{\hat{s}_i\}_{i=1}^{m}$, SKI replaces the $\mK_s$ matrix with a rank $m$ approximation $\hat{\mK}_s = \mW \mK_s^m \mW^\top$, where $\mK_s^m$ is an $m \times m$ Gram matrix with entries ${\mK_s^m}_{(p,q)} = k_s(\hat{s}_p, \hat{s}_q)$, and $\mW$ is an $n\times m$ interpolation matrix whose entries depend on the relative placement of sample points $\{s_i\}_{i=1}^{n}$ and inducing points $\{\hat{s}_i\}_{i=1}^{m}$. Thus, $\hat{\mK}_s \mQ$ can be computed using three successive MVMs: (i) an MVM with $\mW^\top$, followed by (ii) an MVM with $\mK_s^m$, and finally (iii) an MVM with $\mW$. To compute the MVM with $\mW$ matrix in linear time, \citet{kissgp} suggests a sparse $\mW$ matrix derived from local cubic interpolation (only $4$ non-zero entries per row). Thus, even a naive $\mathcal{O}(m^2)$ implementation of an MVM with $\mK_s^m$ substantially reduces the complexity of $\hat{\mK}_s \mQ$ to $\mathcal{O}(n+m^2)$ time and storage. Additionally, the SKI framework offers flexibility with the choice of inducing point locations to further exploit the structure in \GP's kernel functions, e.g., (i) using the Kronecker method \citep{kronecker_Saatchi} with a product kernel offers an $\mathcal{O}(n+Ym^{1+1/Y})$ time and $\mathcal{O}(n+Ym^{2/Y})$ storage complexity, or (ii) the Topelitz method \citep{toeplitz_turner:2010} with a stationary kernel offers an $\mathcal{O}(n + Y m\log m)$ time and $\mathcal{O}(n + Y m)$ storage complexity ($Y$ is input dimensionality).

\subsection{Practical \DBQPG Algorithm}
\label{sec:appendix_practical_dbqpg}
\begin{figure}[!ht]
	\centering
  	\includegraphics[scale=0.5]{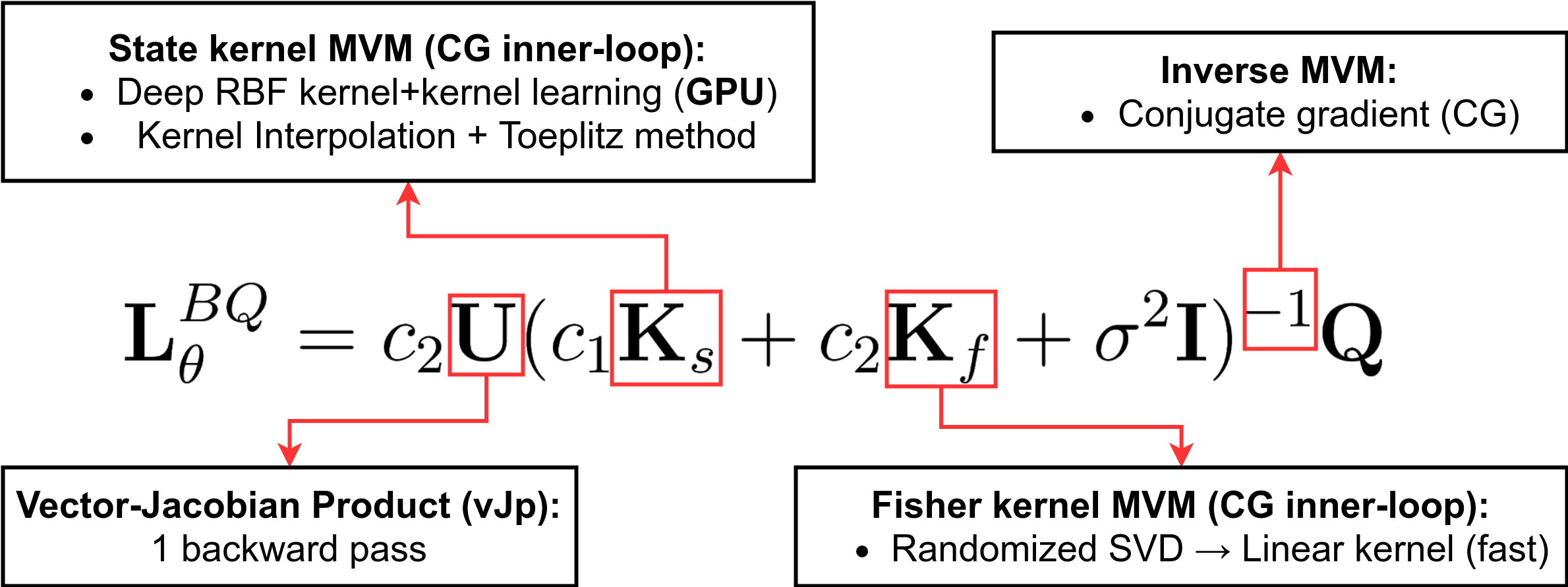}
  	\caption{A summary of fast kernel computation methods used in the practical \DBQPG algorithm.}
    \label{fig:appendix_dbqpg_algo_img}
\end{figure}

\DBQPG brings together the above-mentioned fast kernel methods under one practical algorithm, summarized in Fig.~\ref{fig:appendix_dbqpg_algo_img}.  The first step is to compute the action-value estimates $\Qmc$, an $n$ dimensional vector, either from \MC rollouts/TD(1) estimates or using an explicit critic network. The $\Qmc$ vector, along with the efficient MVM strategies for $\mK_s$ (Supplement Sec.~\ref{sec:appendix_state_mvm}) and $\mK_f$ (Supplement Sec.~\ref{sec:appendix_Fisher_mvm}) are provided to the CG algorithm, which computes $\alpha = (c_1\mK_s + c_2\mK_f +\sigma^2 \mI)^{-1} \Qmc$ in linear-time. Finally, $\mL_{\vtheta}^{BQ} = \mU \alpha$ can be computed using a $vJp$ involving the $\mU$ matrix. For natural gradient algorithms (e.g. \NPG and \TRPO), we precondition the \PG estimate with the $\mG^{-1}$ matrix, which is an i-$Hvp$ operation over KL divergence (similar to \TRPO \citep{TRPO}).

\subsection{Monte-Carlo Estimation of the Fisher Information Matrix}
\begin{figure}[!ht]
	\centering
    \begin{subfigure}
  	\centering
  	\includegraphics[scale=0.07]{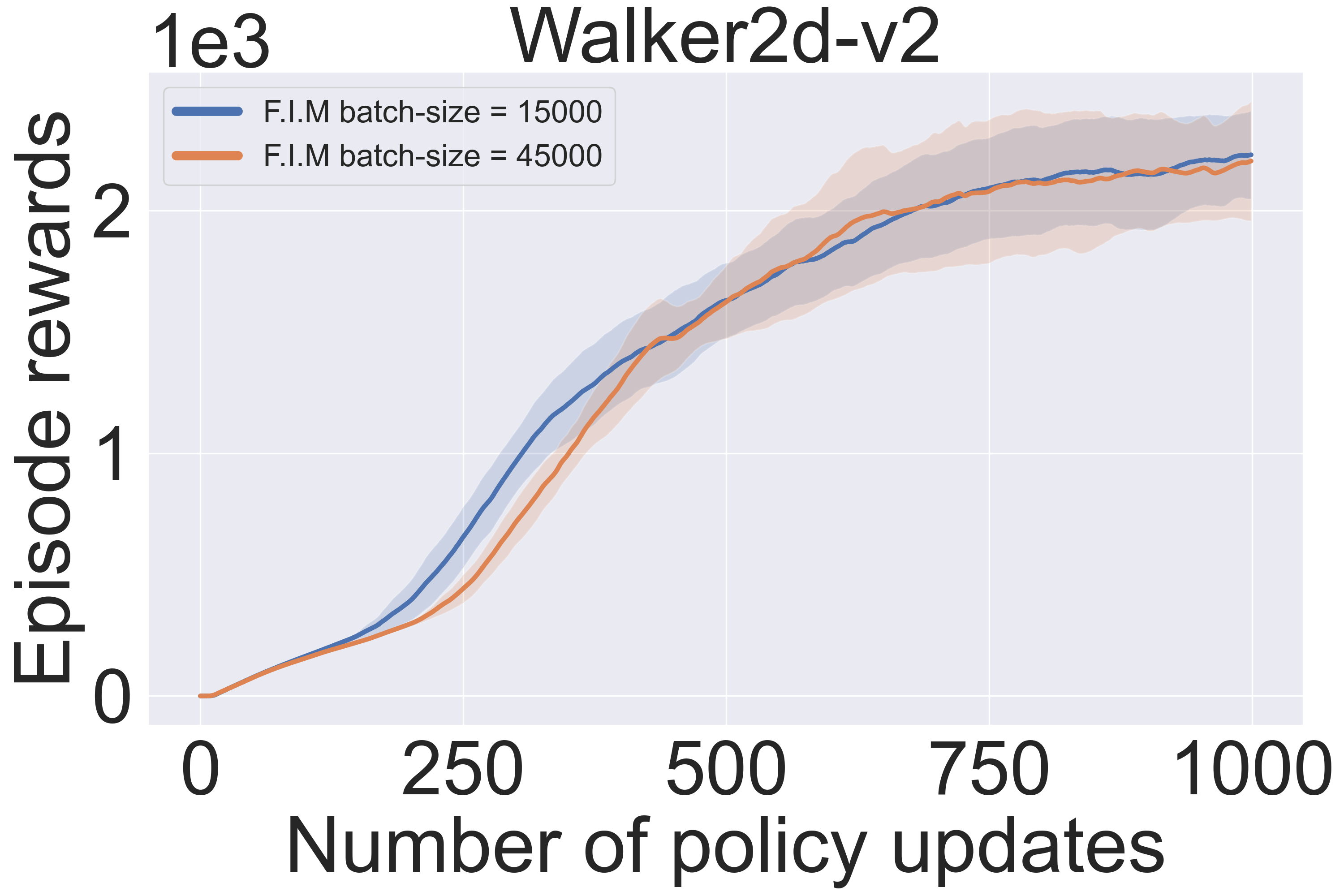}
    \end{subfigure}
    \caption{Vanilla \PG experiment with \DBQPG estimator and 15000 sample size, while estimating the F.I.M $\mG$ using (i) 15000 and (ii) 45000 samples.}
  \label{fig:appendix_fisher_estimation}
\end{figure}

The analytical solution for $\mL_{\vtheta}^{BQ}$ is obtained by assuming that the Fisher information matrix $\mG$ is estimated exactly. However, the practical \DBQPG method estimates $\mG$ (in the Fisher kernel $k_f$) from samples using the Monte-Carlo method (Eq.~\ref{eqn:appendix_kernel_equations}). Here, we investigate the affect of \MC approximation of $\mG$ on the \BQ-\PG performance, by estimating $\mG$ using $3\times$ more samples than \BQ-\PG sample size. It can be seen from Fig.~\ref{fig:appendix_fisher_estimation} that increasing the sample-size of \MC approximation of $\mG$ does not appreciably improve the performance. As a result, while it is possible to replace the \MC approximation of $\mG$ matrix with a more efficient numerical integration method (like \BQ), we do not expect to see significant improvements over the \MC estimation baseline for the $\mG$ matrix.
\vspace{-0.5em}

\section{\UAPG: From Theory to Practice}
\label{appendix:UAPG_practical_implement}

\PG estimation from samples offers stochastic gradient estimates with a non-uniform gradient uncertainty, i.e., the \PG estimator is more uncertain about the gradient's step size along some directions over others. Due to this inherent disparity in the uncertainty of different gradient components, using a constant learning rate occasionally results in large policy updates along directions of high uncertainty, and consequently a catastrophic performance degradation. Thus, \PG algorithms that use \MC-\PG or \DBQPG estimation treat stochastic gradient estimates as the true gradient, making them vulnerable to bad policy updates. \UAPG uses \DBQPG's gradient uncertainty $\mC_{\vtheta}^{BQ}$ to normalize the different components of a \DBQPG estimate based on their respective uncertainties, bring all the gradient components to the same scale. In other words, \UAPG offers gradient estimates with uniform uncertainty, i.e. gradient covariance is the identity matrix. In theory, the \UAPG update is defined as follows:
\begin{align}
    \mL_{\vtheta}^{UAPG} &= \big( \mC_{\vtheta}^{BQ} \big)^{-\frac{1}{2}}\mL_{\vtheta}^{BQ}, \quad \text{such that}
    \\
    \mC_{\vtheta}^{UAPG} &= \big( \mC_{\vtheta}^{BQ} \big)^{-\frac{1}{2}} \mC_{\vtheta}^{BQ} \big( \mC_{\vtheta}^{BQ} \big)^{-\frac{1}{2}} = \mI. \nonumber
\end{align}

However, the empirical $\mC_{\vtheta}^{BQ}$ estimates are often ill-conditioned matrices (spectrum decays quickly) with a numerically unstable inversion. Since $\mC_{\vtheta}^{BQ}$ only provides a good estimate of the top few directions of uncertainty, the \UAPG update is computed from a rank-$\delta$ singular value decomposition (SVD) approximation of $\mC_{\vtheta}^{BQ} \approx \nu_\delta \mI + \sum_{i=1}^{\delta} \vh_i (\nu_i - \nu_\delta) \vh_i^\top$ as,
\begin{align}
\label{eqn:appendix_UAPG_practical_vanilla}
   \mL_{\vtheta}^{UAPG} = \nu_{\delta}^{-\frac{1}{2}} \big(\mI + \sum\nolimits_{i=1}^{\delta} \vh_i \big(\sqrt{{\nu_{\delta}}/{\nu_i}} - 1\big)\vh_i^\top \big) \mL_{\vtheta}^{BQ}.
\end{align}
The computational complexity of the randomized SVD operation is $\mathcal{O}(2\delta\mathcal{M} + \delta^2|\Theta| + \delta^2 n)$, where $\mathcal{M}$ is the computational complexity of the MVM operation. Using the fast kernel computation methods mentioned earlier, the complexity of MVM operation $\mathcal{M}$ and, subsequently, the randomized SVD algorithm reduces to linear in the number of policy parameters $|\Theta|$ and the sample-size $n$ (see Fig.~\ref{fig:appendix_wall_clock} for empirical evidence).

On the other hand, for the natural policy gradient update $\mL^{NBQ}_{\vtheta} = \mG^{-1} \mL^{BQ}_{\vtheta}$, the gradient uncertainty $\mC^{NBQ}_{\vtheta}$ is,
\begin{align}
\label{eqn:appendix_UAPG_cov_natural}
\mC_{\vtheta}^{NBQ} &= \mG^{-1} \mC_{\vtheta}^{BQ} \mG^{-1}\nonumber
    \\
    &= c_2 (\mG^{-1} - c_2 \mG^{-1} \mU \Big( c_1 \mK_s + c_2 \mK_f + \sigma^2 \mI \Big)^{-1} \mU^\top \mG^{-1})\nonumber
    \\
    &= c_2 (\mG + c_2 \mU \left( c_1 \mK_s + \sigma^2 \mI \right)^{-1} \mU^\top)^{-1},
\end{align}
and the ideal \UAPG update is $\mL^{NUAPG}_{\vtheta} = \left( \mC^{NBQ}_{\vtheta}\right)^{-\frac{1}{2}} \mG^{-1}\mL^{BQ}_{\vtheta}$. However, since $\mC_{\vtheta}^{NBQ}$ is the inverse of an ill-conditioned matrix, the singular values corresponding to the high-uncertainty directions of $\mC_{\vtheta}^{NBQ}$ show very little variation in uncertainty. Thus, we instead apply a low-rank approximation on ${\mC_{\vtheta}^{NBQ}}^{-1} \approx \nu_\delta \mI + \sum_{i=1}^{\delta} \vh_i (\nu_i - \nu_\delta) \vh_i^\top$ for the \UAPG update of \NPG:
\begin{equation}
\label{eqn:appendix_UAPG_npg}
    \mL_{\vtheta}^{NUAPG} = \nu_{\delta}^{\frac{1}{2}} \Big(\mI + \sum\nolimits_{i=1}^{\delta} \vh_i \big( \min \big(\sqrt{{\nu_i}/{\nu_{\delta}}}, \eps \big) - 1\big)\vh_i^\top \Big) \mG^{-1} \mL_{\vtheta}^{BQ}, \qquad \eps > 1
\end{equation}
Note that ${\mC^{NBQ}_{\vtheta}}^{-1}$ is an ill-conditioned matrix whose top $\delta$ PCs denote the least uncertain/most confident directions of natural gradient estimation (equivalent to the bottom $\delta$ PCs of $\mC^{NBQ}_{\vtheta}$). Further, we replace $\sqrt{{\nu_i}/{\nu_{\delta}}}$ with $\min(\sqrt{{\nu_i}/{\nu_\delta}},\eps)$ in Eq.~\ref{eqn:appendix_UAPG_npg} to avoid taking large steps along these directions, solely on the basis of their uncertainty. Thus, for \NPG estimates, \UAPG offers a more reliable policy update direction by relatively increasing the step size along the most confident directions (i.e., the top $\delta$ PCs of ${\mC_{\vtheta}^{NBQ}}^{-1}$), as opposed to lowering the stepsize for the most uncertain directions like in the case of \UAPG over Vanilla \PG estimates.

\subsection{Relation between the Gradient Uncertainties of Vanilla \PG and \NPG}
\label{sec:appendix_vpg_npg_uapg}
Empirically, we found that the optimal value of $c_2 \ll 1$, at which point, the gradient uncertainty of Vanilla \PG and Natural \PG algorithms approximate to:
\begin{align}
\mC_{\vtheta}^{BQ}&= c_2 \mG - c_2^2\mU (c_1\mK_s + c_2\mK_f +\sigma^2\mI)^{-1} \mU^\top \approx c_2 \mG,
\\
\mC_{\vtheta}^{NBQ} &= c_2 (\mG + c_2 \mU \left( c_1 \mK_s + \sigma^2 \mI \right)^{-1} \mU^\top)^{-1} \approx c_2 {\mG}^{-1}.
\end{align}
This observation is particularly interesting because for $c_2 \ll 1$, most uncertain gradient directions for vanilla \PG approximately correspond to the most confident (least uncertain) directions for \NPG. Crudely speaking, the natural gradient takes the step size along each direction and divides it by the estimated variance (from the gradient covariance matrix), which results in an inversion of the uncertainty. In contrast, \UAPG divides the stepsize along each direction by the estimated standard deviation, which results in uniform uncertainty along all the directions. Moreover, for $c_2 \ll 1$, the ideal UAPG update for both vanilla \PG and \NPG converges along the $\mG^{-\frac{1}{2}}\mL_{\vtheta}^{BQ}$ direction.

\section{Wall-Clock Performance of \DBQPG and \UAPG}
\label{sec:appendix_wall_clock_performance}

\begin{figure}[!ht]
	\centering
  	\includegraphics[scale=0.28]{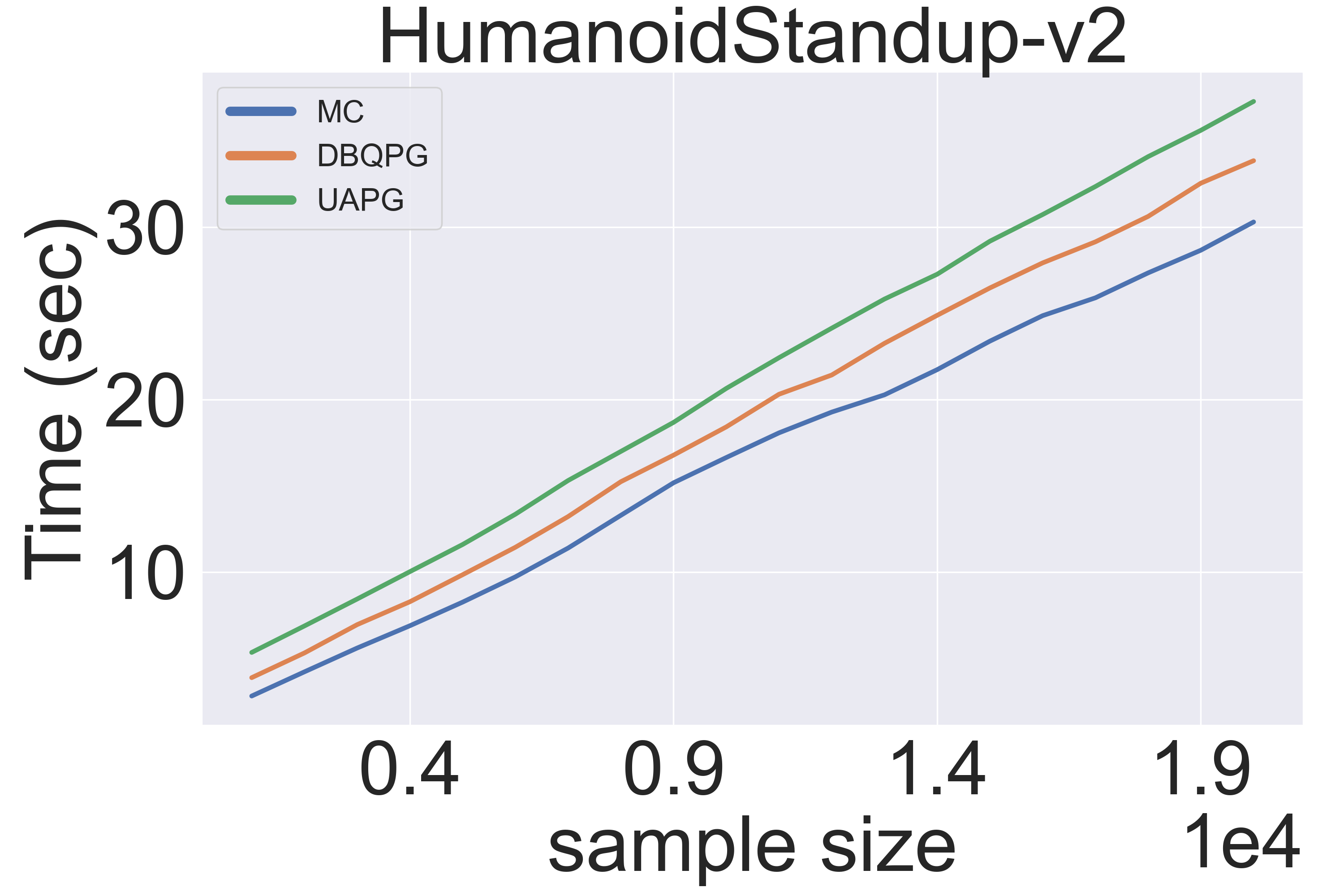}
  	\caption{Wall-Clock time for \PG estimation using \MC, \DBQPG and \UAPG methods on HumanoidStandup-v2 environment.}
    \label{fig:appendix_wall_clock}
\end{figure}

\begin{figure*}[h!]
	\centering
    \begin{subfigure}
  	\centering
  	\includegraphics[scale=\lineWW]{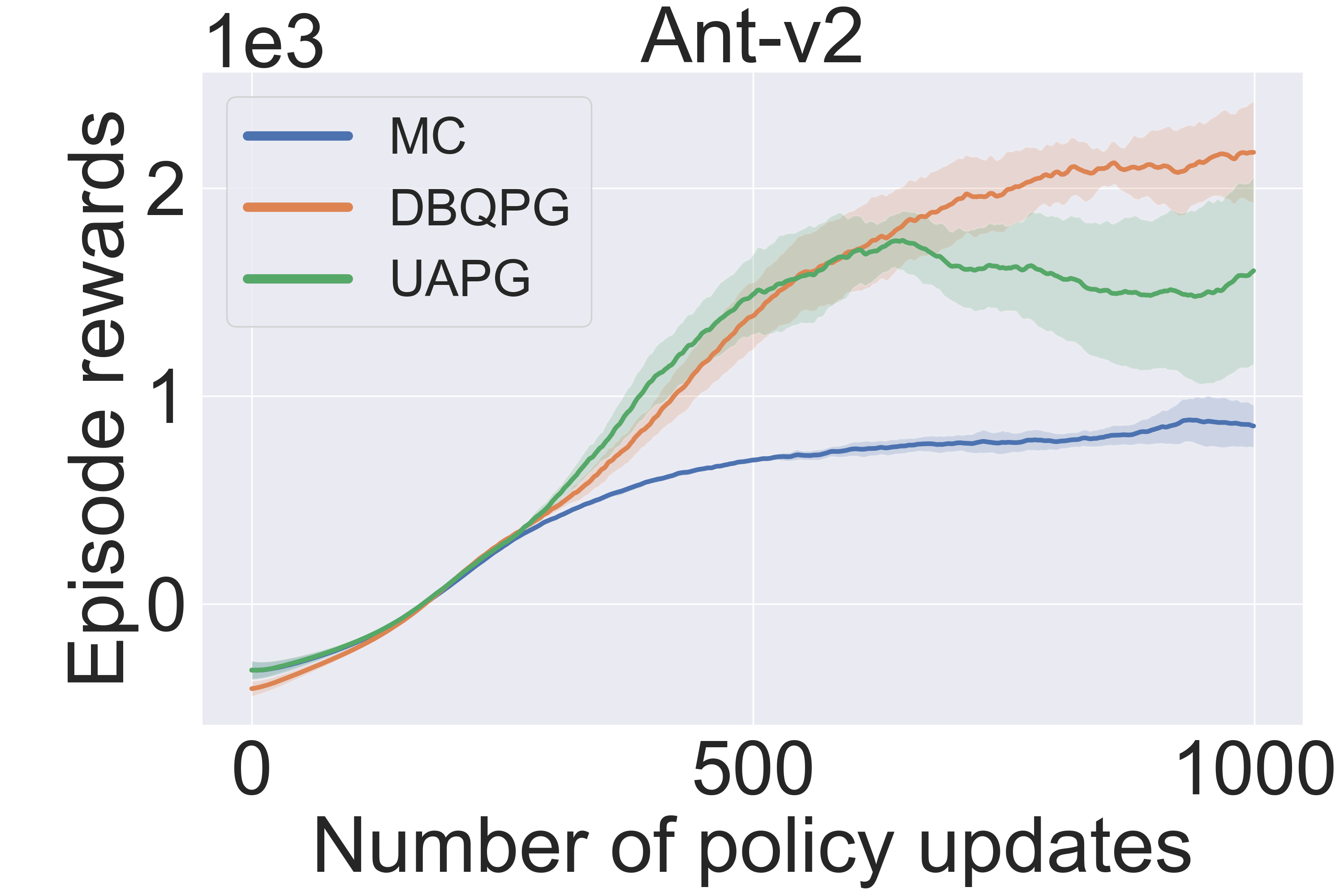}
    \end{subfigure}
    \hspace{-0.7em}
    \begin{subfigure}
  	\centering
  	\includegraphics[scale=\lineWW]{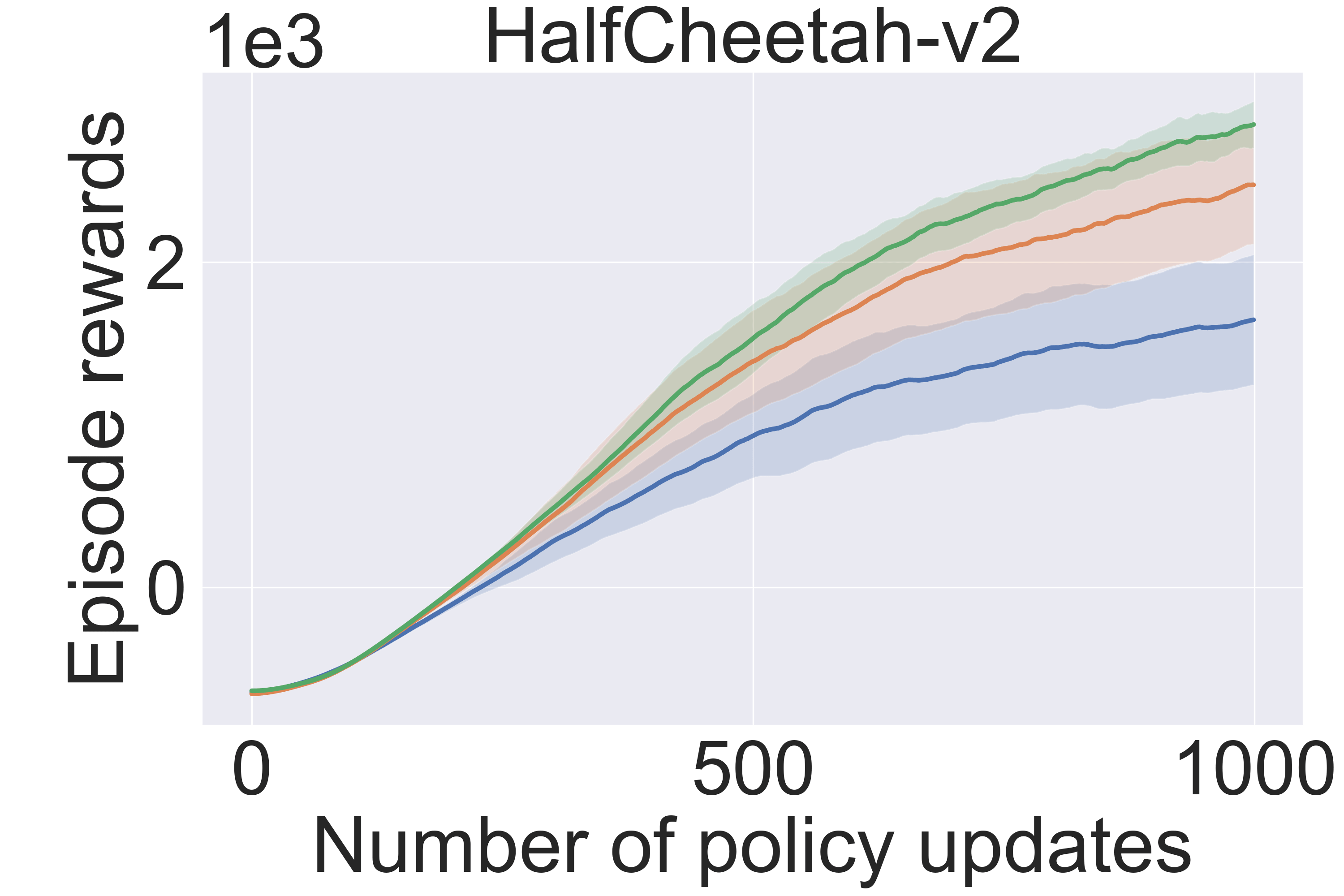}
    \end{subfigure}
    \hspace{-0.7em}
    \begin{subfigure}
  	\centering
  	\includegraphics[scale=\lineWW]{Plots/VanillaPG/Humanoid_plot.png}
    \end{subfigure}
    \hspace{-0.7em}
    \begin{subfigure}
  	\centering
  	\includegraphics[scale=\lineWW]{Plots/VanillaPG/HumanoidStandup_plot.png}
    \end{subfigure}
    \begin{subfigure}
  	\centering
  	\hspace*{0.1cm}
  	\includegraphics[scale=\lineWW]{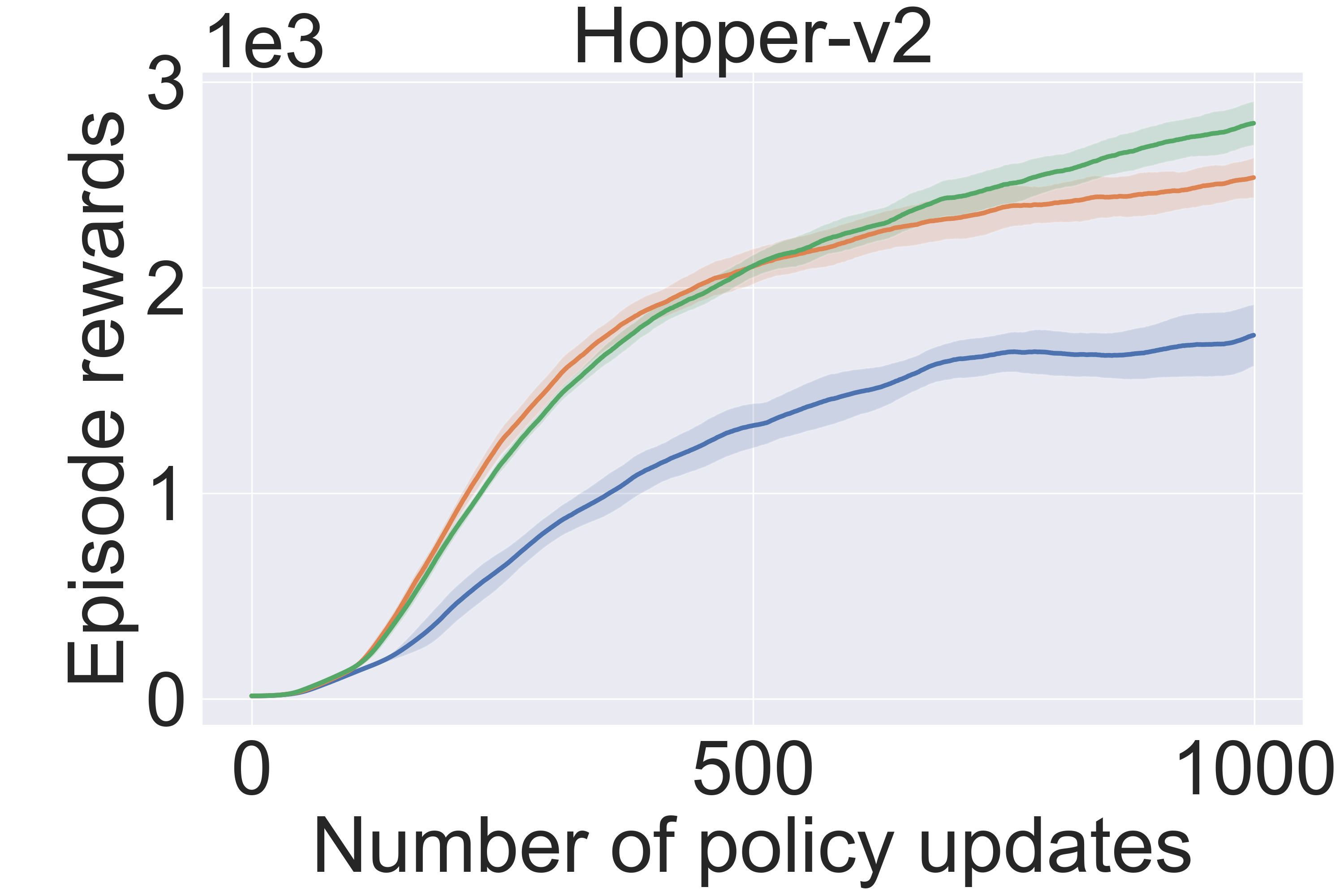}
    \end{subfigure}
    \hspace{-0.7em}
    \begin{subfigure}
  	\centering
  	\includegraphics[scale=\lineWW]{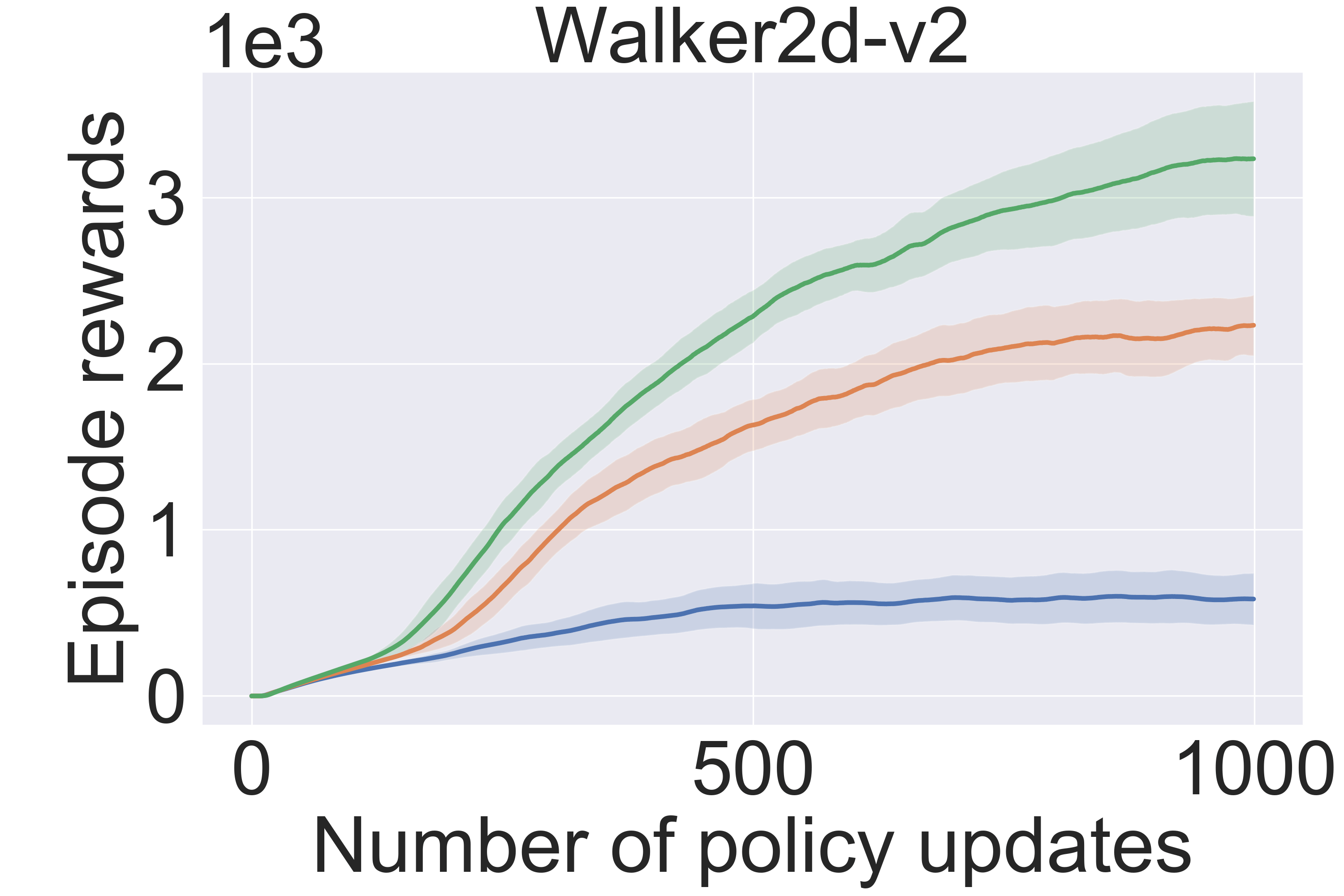}
    \end{subfigure}
    \hspace{-0.7em}
    \begin{subfigure}
  	\centering
  	\includegraphics[scale=\lineWW]{Plots/VanillaPG/Swimmer_plot.png}
    \end{subfigure}
    {\LARGE \:\:\:\:\:\:Vanilla \PG}

\begin{center}
\line(1,0){475}
\end{center}
	\centering
    \begin{subfigure}
  	\centering
  	\includegraphics[scale=\lineWW]{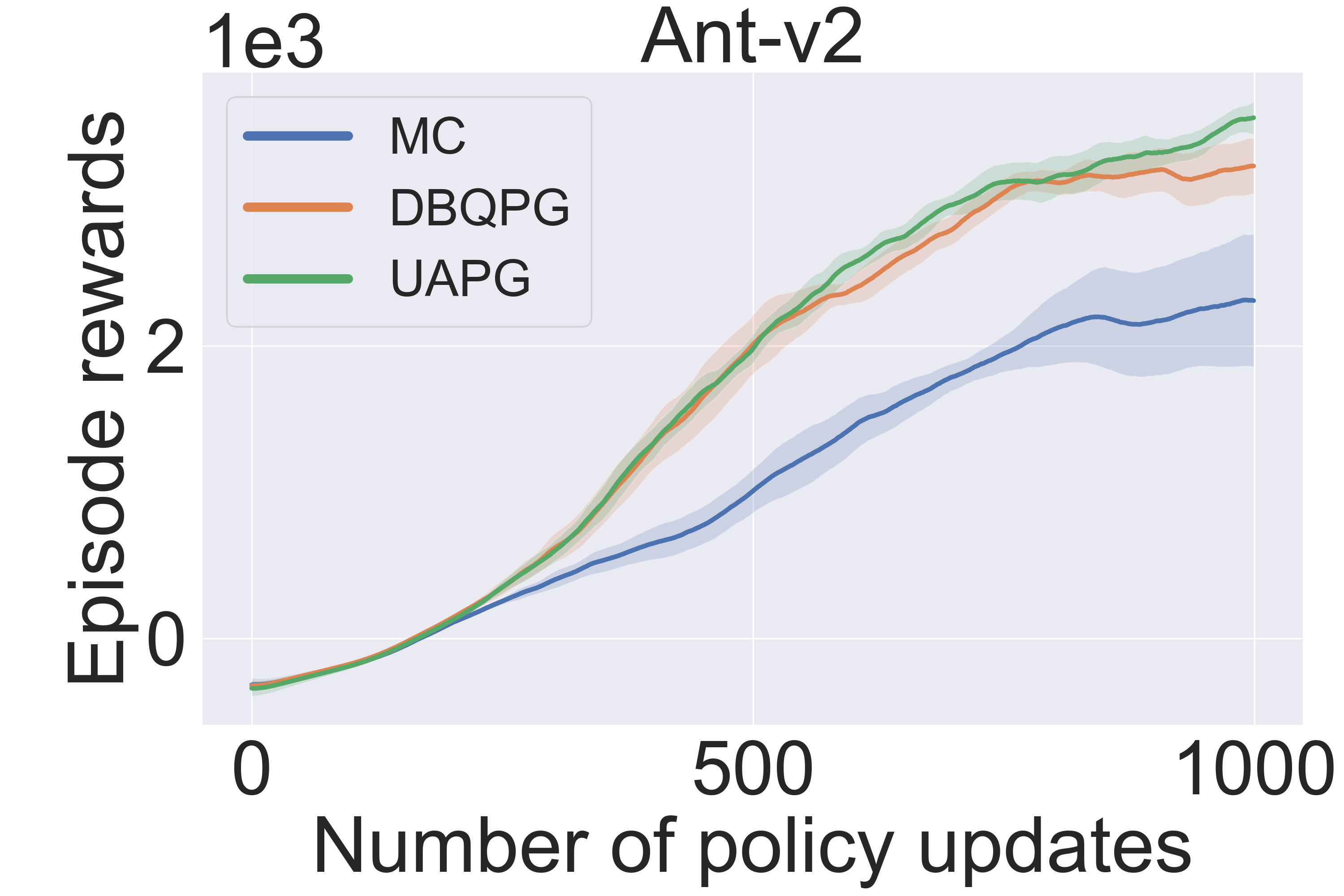}
    \end{subfigure}
    \hspace{-0.7em}
    \begin{subfigure}
  	\centering
  	\includegraphics[scale=\lineWW]{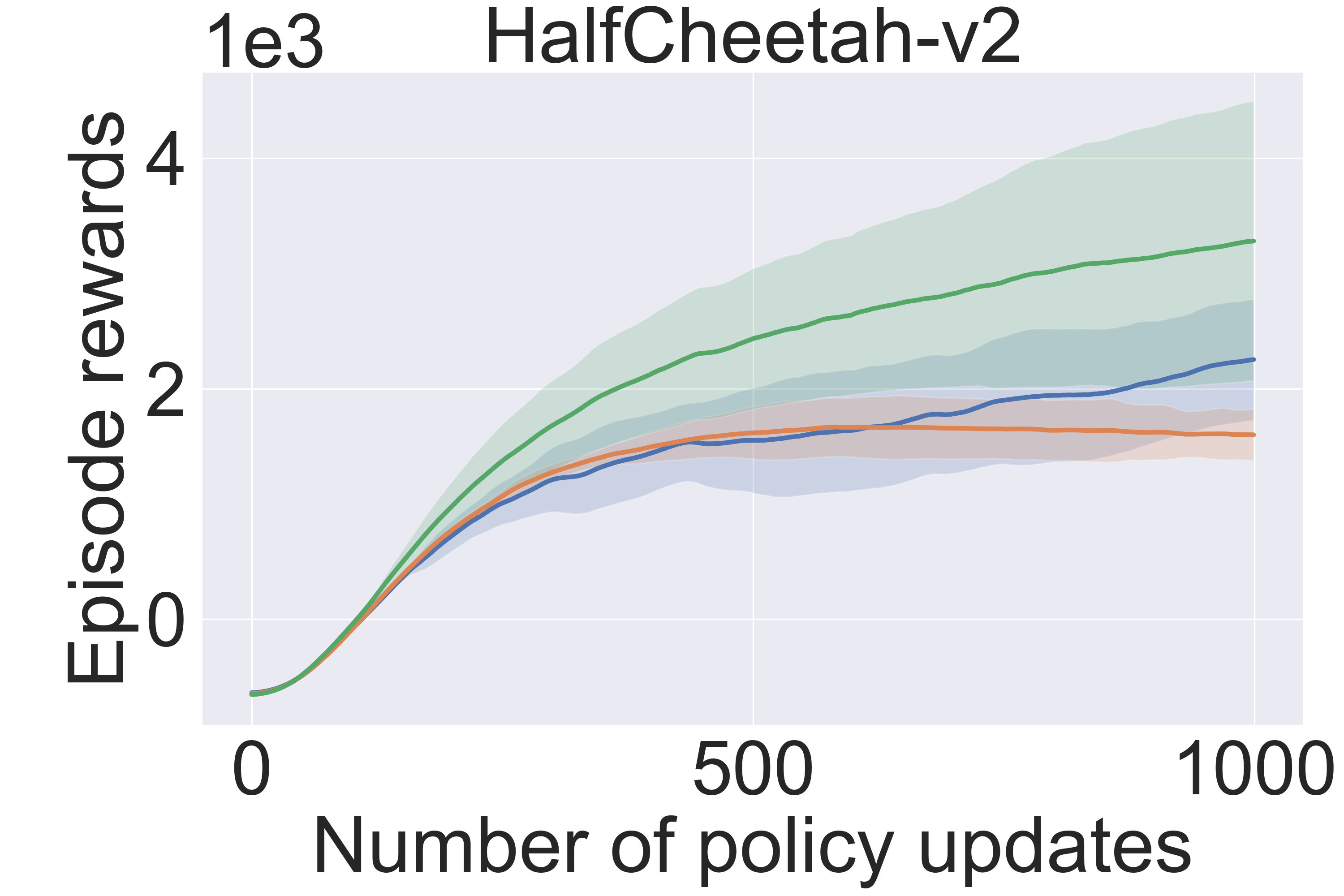}
    \end{subfigure}
    \hspace{-0.7em}
    \begin{subfigure}
  	\centering
  	\includegraphics[scale=\lineWW]{Plots/NPG/Humanoid_plot.png}
    \end{subfigure}
    \hspace{-0.7em}
    \begin{subfigure}
  	\centering
  	\includegraphics[scale=\lineWW]{Plots/NPG/HumanoidStandup_plot.png}
    \end{subfigure}
    \begin{subfigure}
  	\centering
  	\hspace*{0.1cm}
  	\includegraphics[scale=\lineWW]{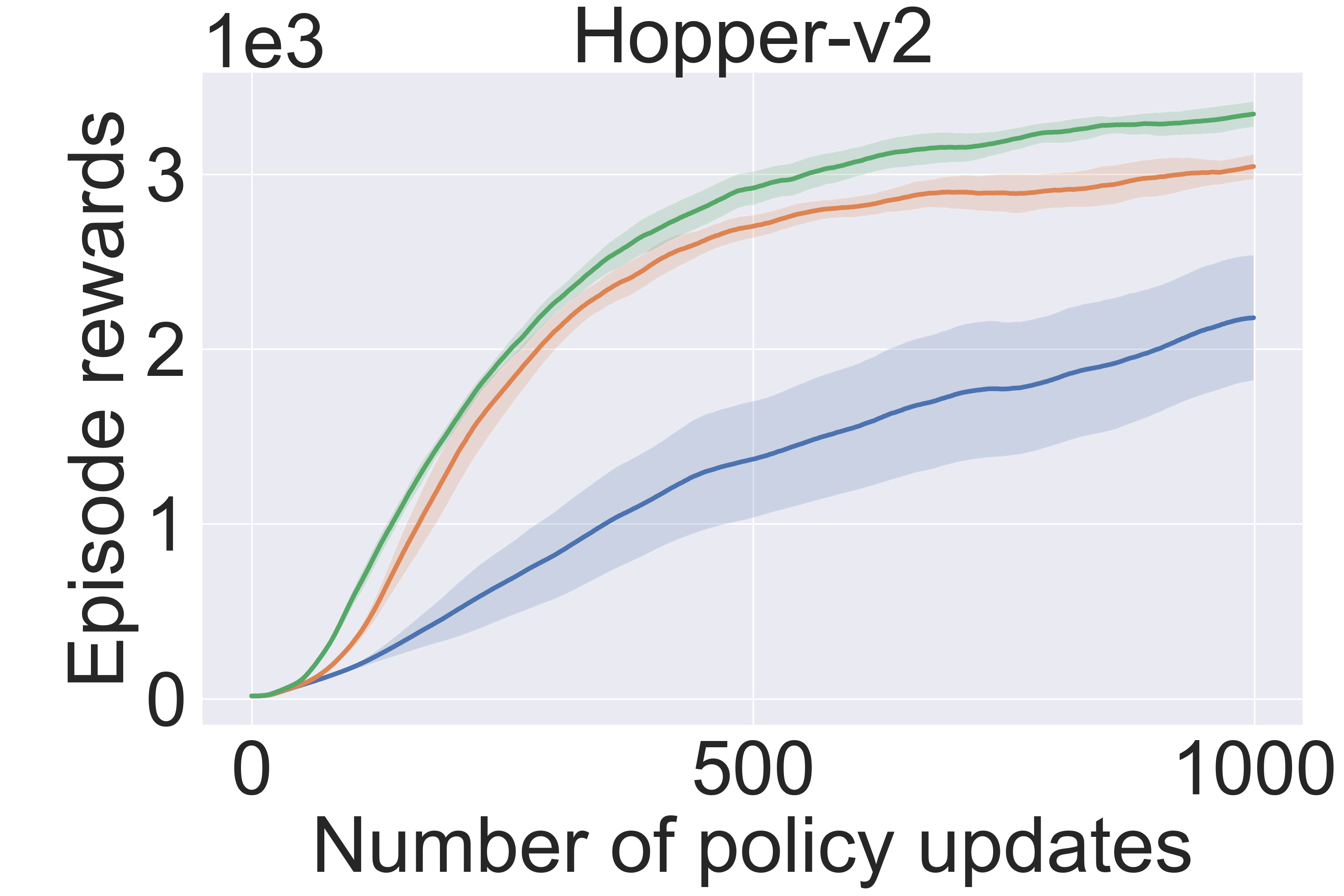}
    \end{subfigure}
    \hspace{-0.7em}
    \begin{subfigure}
  	\centering
  	\includegraphics[scale=\lineWW]{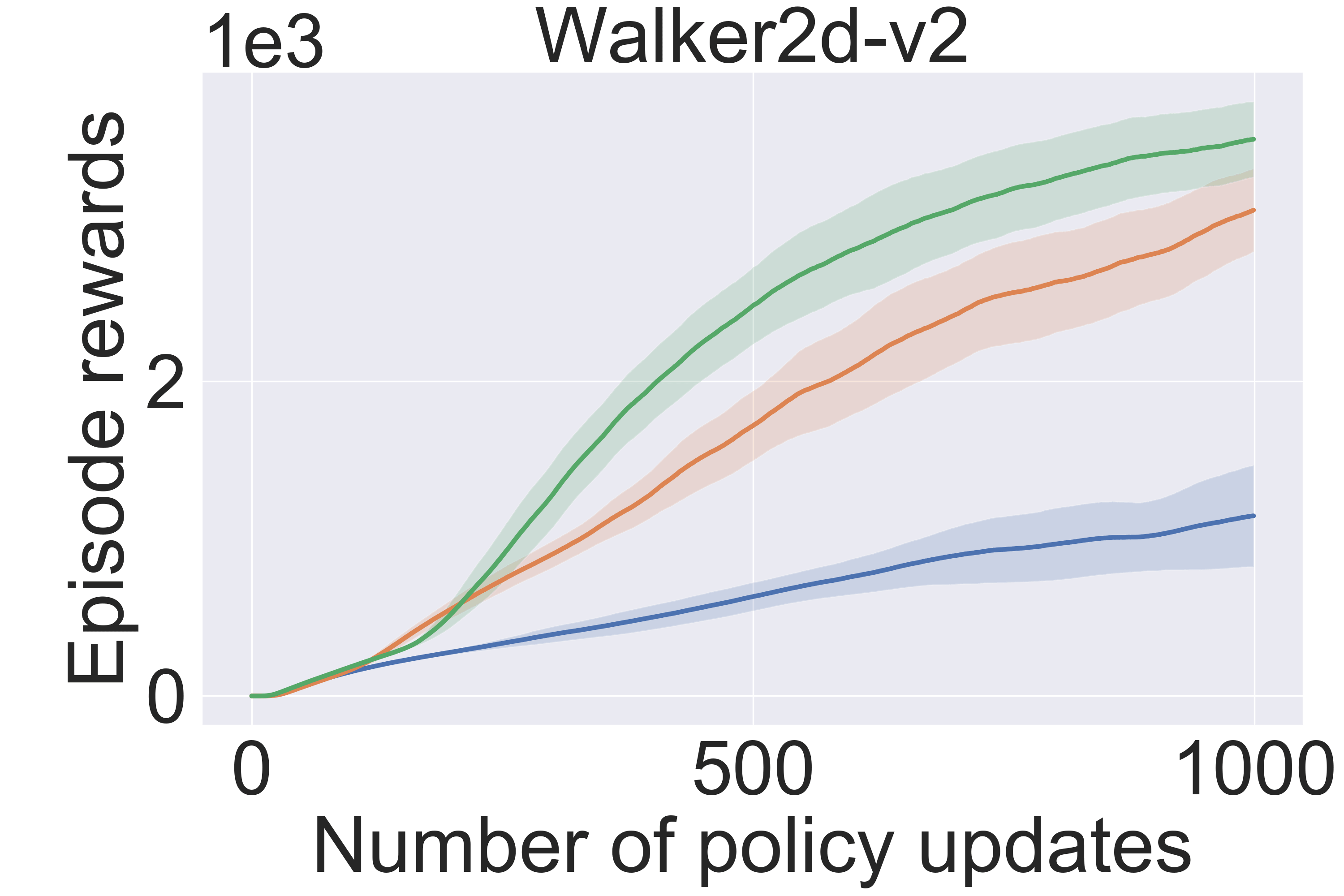}
    \end{subfigure}
    \hspace{-0.7em}
    \begin{subfigure}
  	\centering
  	\includegraphics[scale=\lineWW]{Plots/NPG/Swimmer_plot.png}
    \end{subfigure}
    {\LARGE \:\:\:\:\:\:\:\:\:\:\NPG\:\:\:\:}
    \begin{center}
    \line(1,0){475}
    \end{center}


	\centering
    \begin{subfigure}
  	\centering
  	\includegraphics[scale=\lineWW]{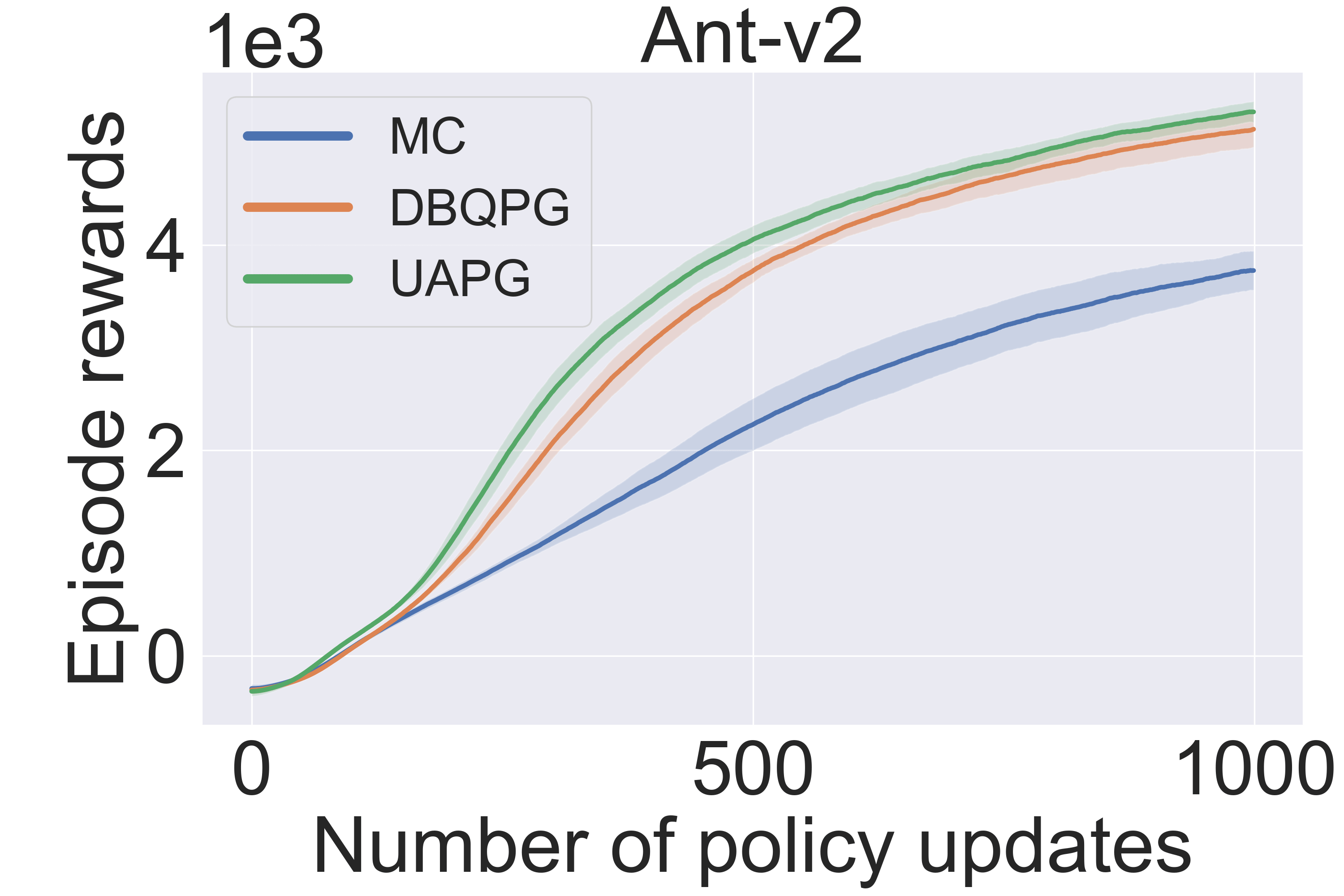}
    \end{subfigure}
    \hspace{-0.7em}
    \begin{subfigure}
  	\centering
  	\includegraphics[scale=\lineWW]{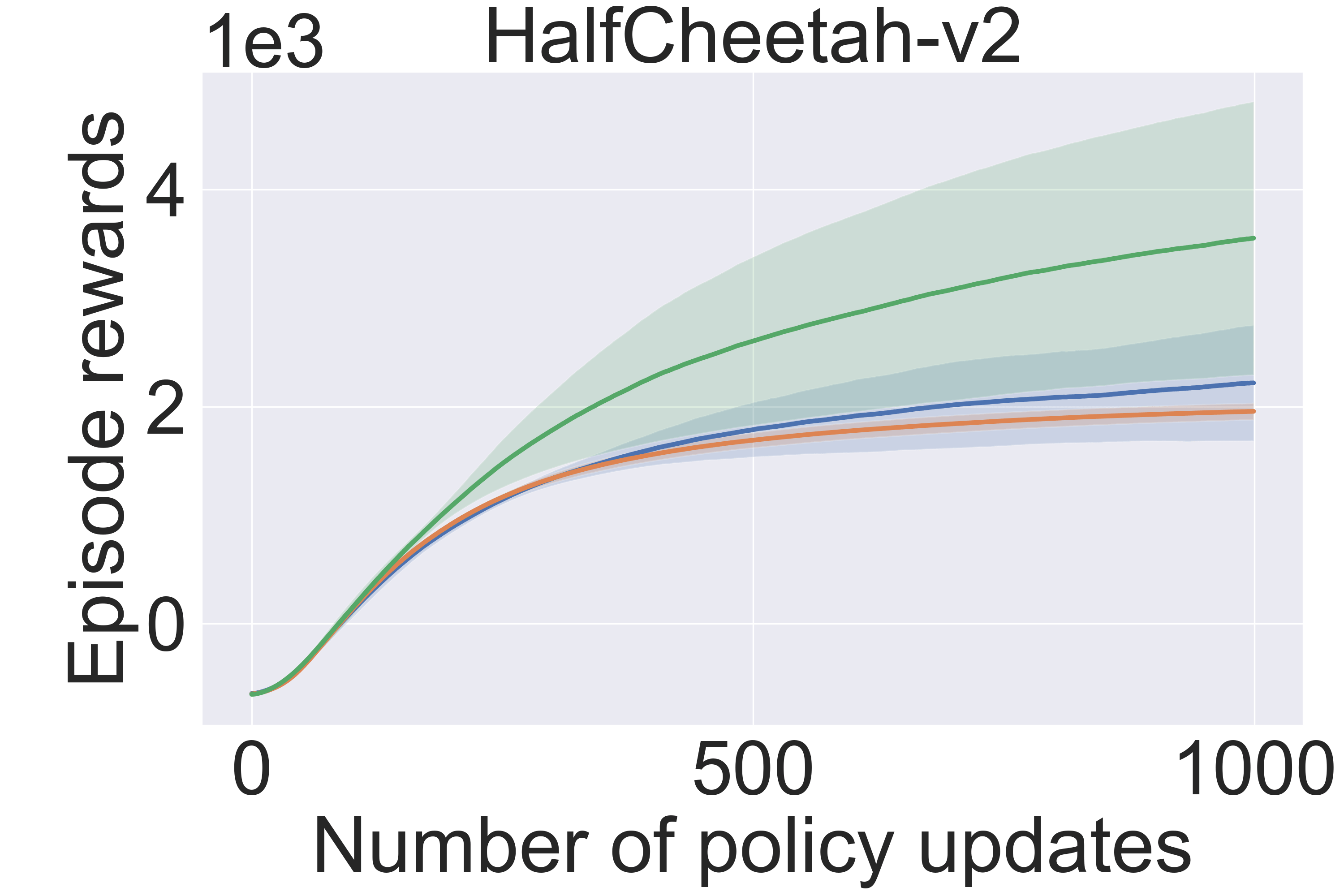}
    \end{subfigure}
    \hspace{-0.7em}
    \begin{subfigure}
  	\centering
  	\includegraphics[scale=\lineWW]{Plots/TRPO/Humanoid_plot.png}
    \end{subfigure}
    \hspace{-0.7em}
    \begin{subfigure}
  	\centering
  	\includegraphics[scale=\lineWW]{Plots/TRPO/HumanoidStandup_plot.png}
    \end{subfigure}
    \begin{subfigure}
  	\centering
  	\hspace*{0.1cm}
  	\includegraphics[scale=\lineWW]{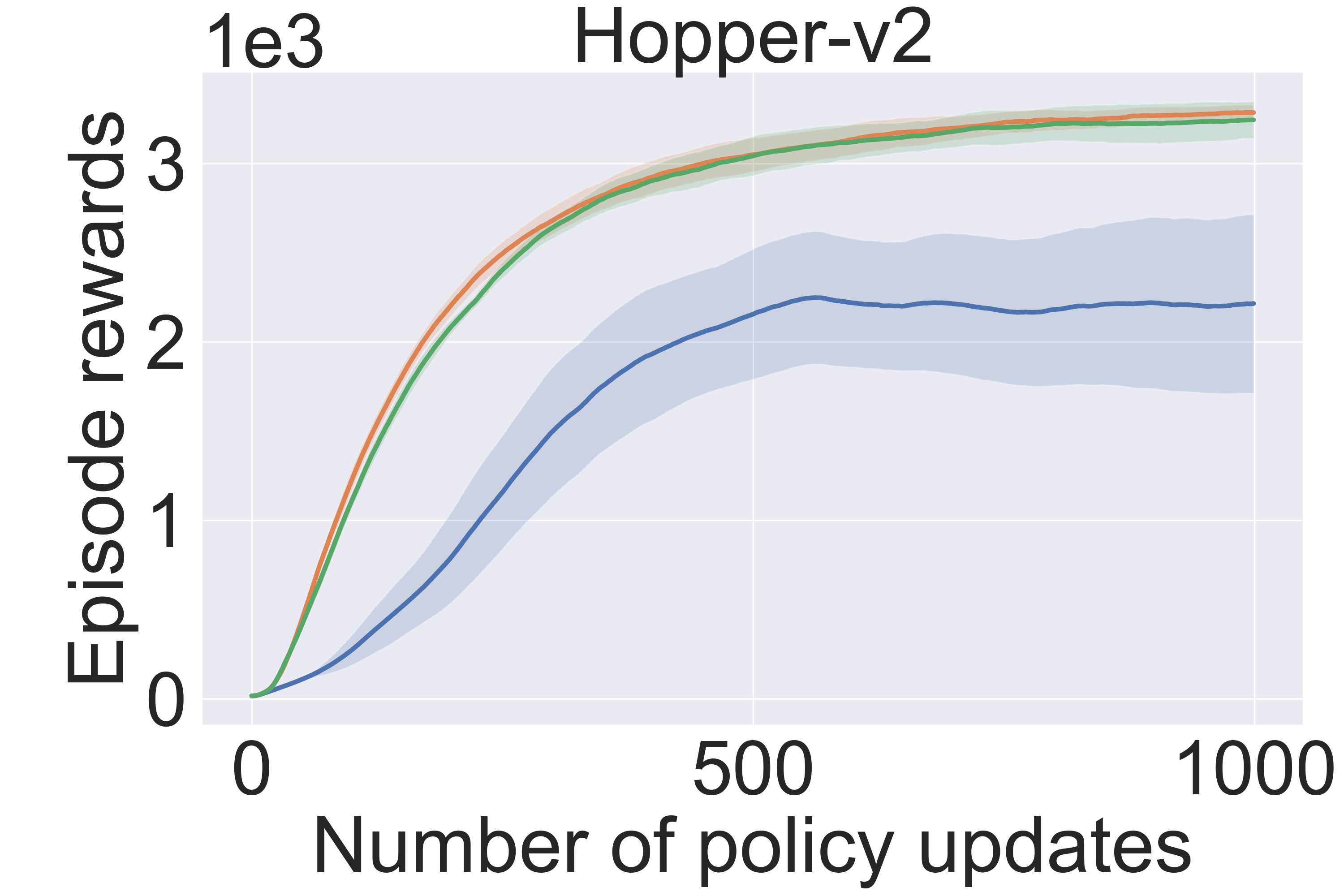}
    \end{subfigure}
    \hspace{-0.7em}
    \begin{subfigure}
  	\centering
  	\includegraphics[scale=\lineWW]{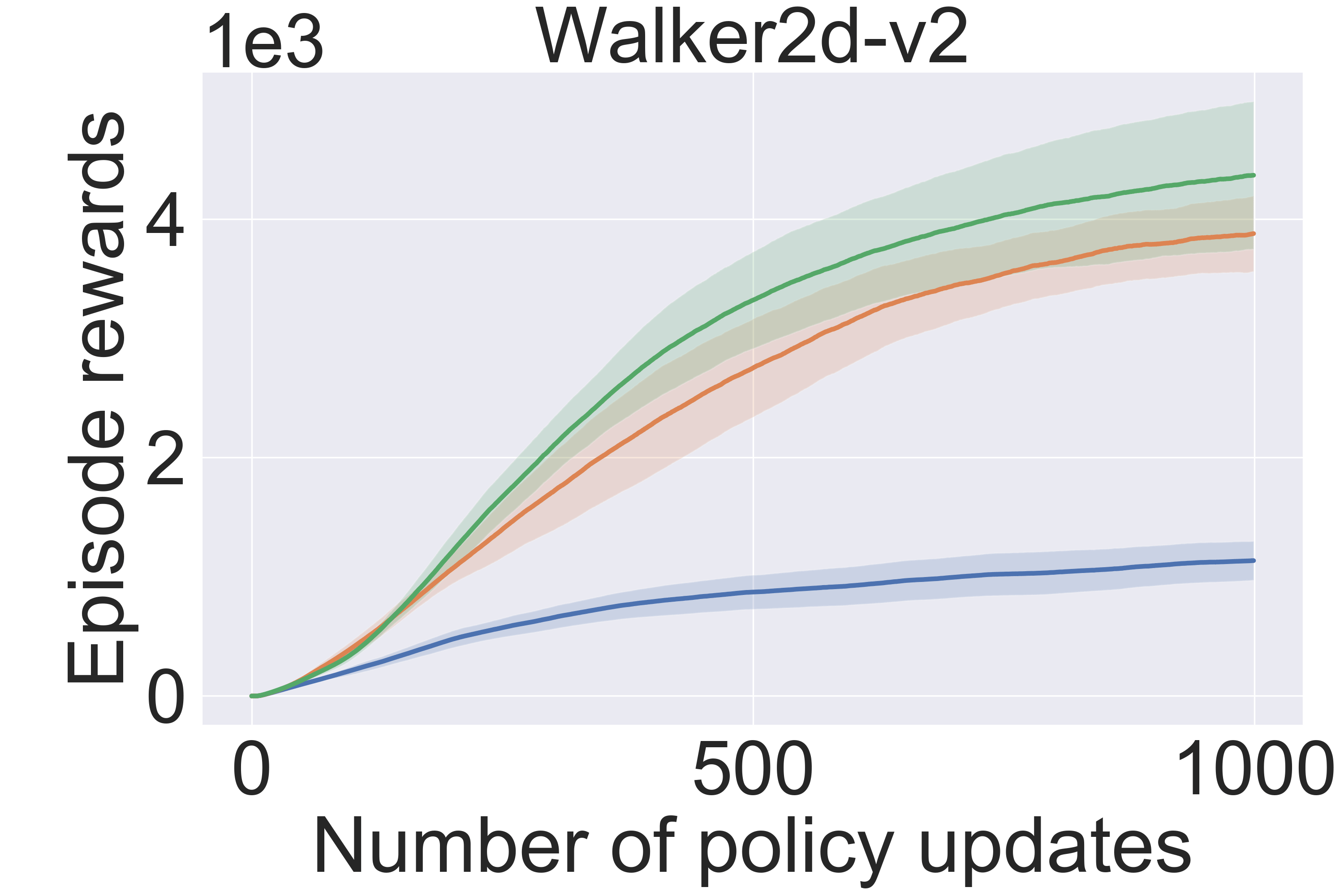}
    \end{subfigure}
    \hspace{-0.7em}
    \begin{subfigure}
  	\centering
  	\includegraphics[scale=\lineWW]{Plots/TRPO/Swimmer_plot.png}
    \end{subfigure}
    {\LARGE \:\:\:\:\:\:\:\:\:\TRPO\:\:\:}
    \begin{center}
    \line(1,0){475}
    \end{center}
\caption{Full comparison of \BQ-based methods and \MC estimation in vanilla \PG, \NPG, and \TRPO frameworks across $7$ MuJoCo environments. The agent's performance is averaged over $10$ runs.
}
  \label{fig:appendix_all_comp_plots}
\end{figure*}

The wall-clock time for \PG estimation using \MC, \DBQPG and \UAPG, on ``HumanoidStandup-v2'' environment (376-d state and 17-d action space) is reported in Fig.~\ref{fig:appendix_wall_clock}. Clearly, \DBQPG and \UAPG methods have a negligible computational overhead over \MC-\PG, while being significantly more sample efficient (Fig.~\ref{fig:appendix_all_comp_plots}). Further, it is also clear from the plot that the wall-clock time of \MC, \DBQPG and \UAPG methods increases linearly in sample size $n$, which agrees with our complexity analysis in Sec.~\ref{sec:DBQPG_algorithm} (main paper).

\section{Comparison with Prior \BQ-\PG works}
\label{sec:appendix_BACvsDBQPG}

\citet{Engel_2003} is one of the first RL approaches to use \GP{}s for approximating the action-value function $Q_{\pi_{\vtheta}}$. The Bayesian deep Q-networks (BDQN) \citep{azizzadenesheli2018efficient} extends this idea to  high-dimensional domains with a discrete action space. More specifically, BDQN uses Bayesian linear regression, i.e., a parametric \GP with a linear kernel, instead of linear regression to learn the last layer in a standard DQN architecture. BDQN then uses the uncertainty modeled by the \GP to perform Thompson sampling on the learned posterior distribution for efficiently balancing exploration and exploitation. Like in BDQN, the implicit \GP critic in \DBQPG can also be seen as a standard deep critic network with its final linear layer substituted by a non-parametric \GP layer.

\begin{figure*}[!ht]
	\centering
    \begin{subfigure}
  	\centering
  	\includegraphics[scale=0.07]{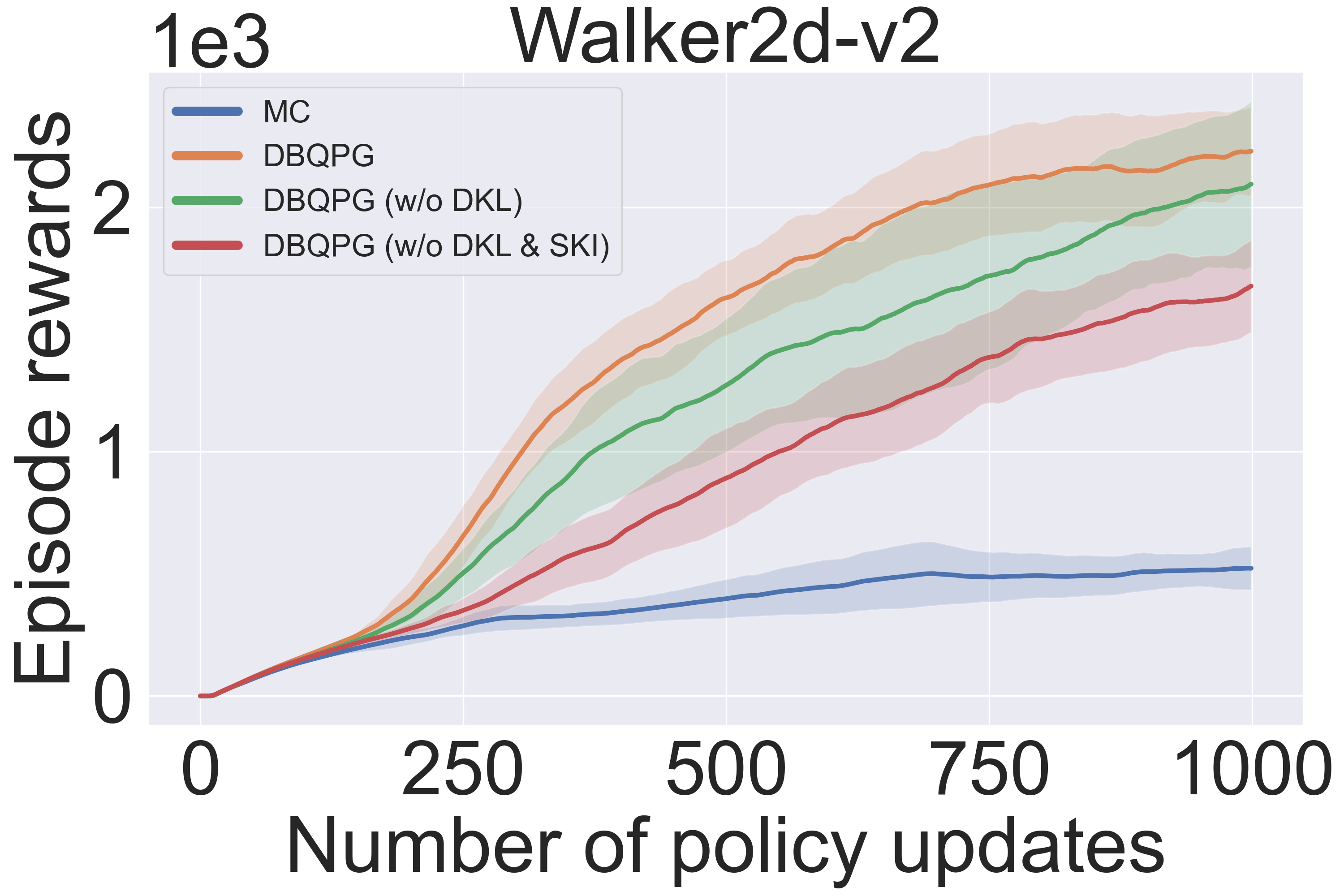}
    \end{subfigure}
    \begin{subfigure}
  	\centering
  	\includegraphics[scale=0.07]{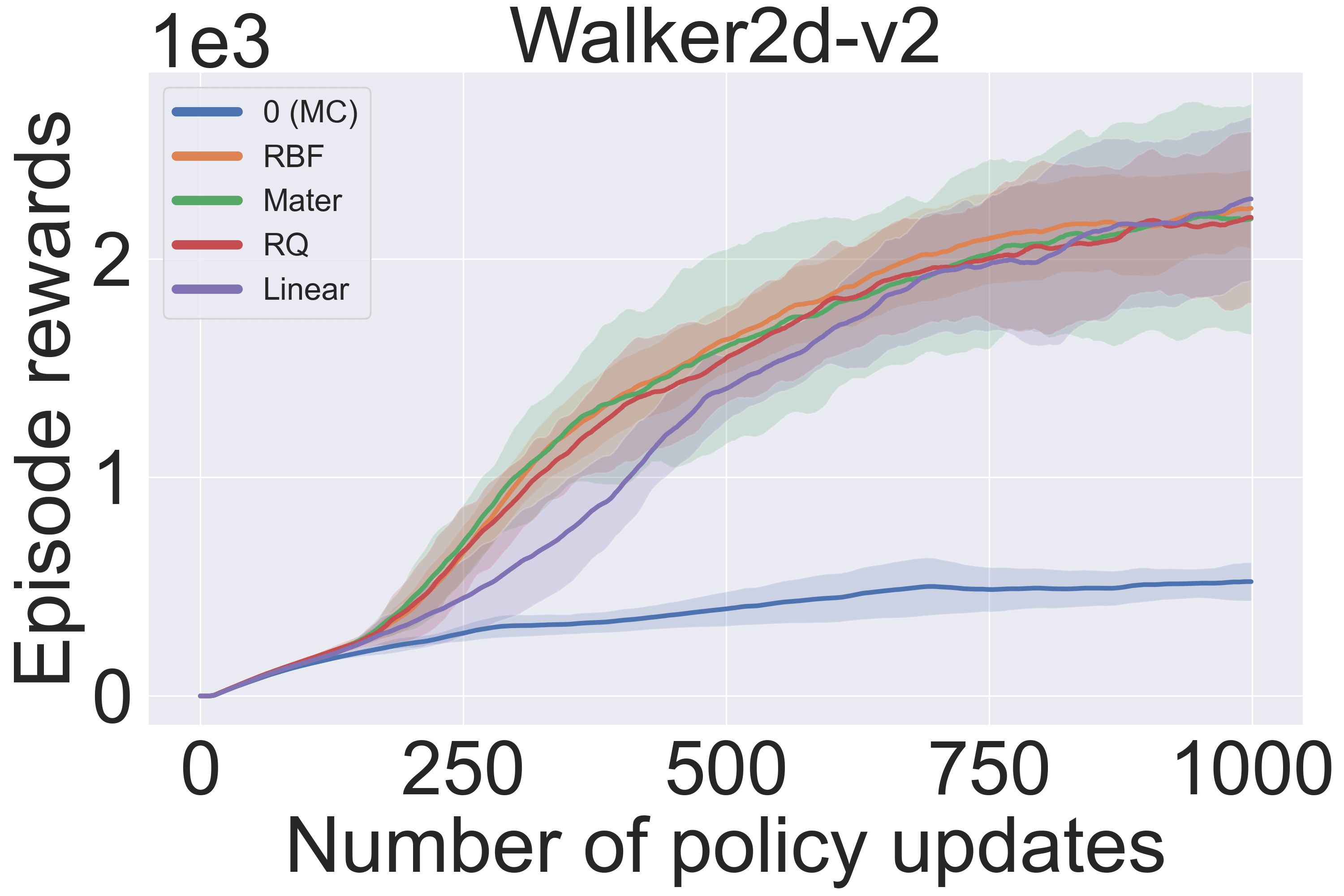}
    \end{subfigure}
    \caption{(left) Ablation study to investigate the role of different components of \DBQPG, and (right) Comparison between common state kernel $k_s$ choices.}
  \label{fig:appendix_Ablation_KernelSelection_WallClock}
\end{figure*}
The \DBQPG algorithm is an extension of the (sequential) Bayesian Actor-Critic (\BAC) algorithm \citep{BAC_1} that leverages the automatic differentiation framework and fast kernel computation methods for \PG estimation from a batch of samples (parallel), thus placing the \BQ-\PG framework in the context of contemporary \PG algorithms. Aside from this, \DBQPG also replaces the traditional inducing points framework \citep{Inducing_points_SOR} ($\mathcal{O}(m^2n+m^3)$ time and $\mathcal{O}(mn+m^2)$ storage complexity) used in \BAC, with a more computationally-efficient alternative, SKI ($\mathcal{O}(n+m^2)$ time and storage). Moreover, \DBQPG boosts the expressive power of base (state) kernel with a deep neural network, followed by kernel learning its parameters (Eq.~\ref{eqn:GP_nll}, main paper).

Switching from deep kernel back to a base kernel (\DBQPG (w/o DKL)) corresponds to a noticeable drop in performance as shown in Fig.~\ref{fig:appendix_Ablation_KernelSelection_WallClock} (left). Moreover, replacing the SKI approximation with a traditional inducing points method (\DBQPG (w/o DKL \& SKI)) further lowers the performance.
Here, ``\DBQPG (w/o DKL \& SKI)'' can be thought of as the reimplementation of \BAC in batch settings (parallel computation). Interestingly, \BAC (ours), i.e. \DBQPG (w/o DKL \& SKI), still manages to outperform the \MC-\PG baseline. Further, Fig.~\ref{fig:appendix_Ablation_KernelSelection_WallClock} (middle) suggests that most popular base kernels, when coupled with a DNN feature extractor, provide nearly similar performance, while easily outperforming the \MC-\PG baseline. This observation suggests that a deep (state) kernel with any of the popular base kernel choices is often a better prior that the trivial state kernel $k_s = 0$, which corresponds to  \MC-\PG.

\begin{figure*}[!ht]
	\centering
    \begin{subfigure}
  	\centering
  	\includegraphics[scale=0.07]{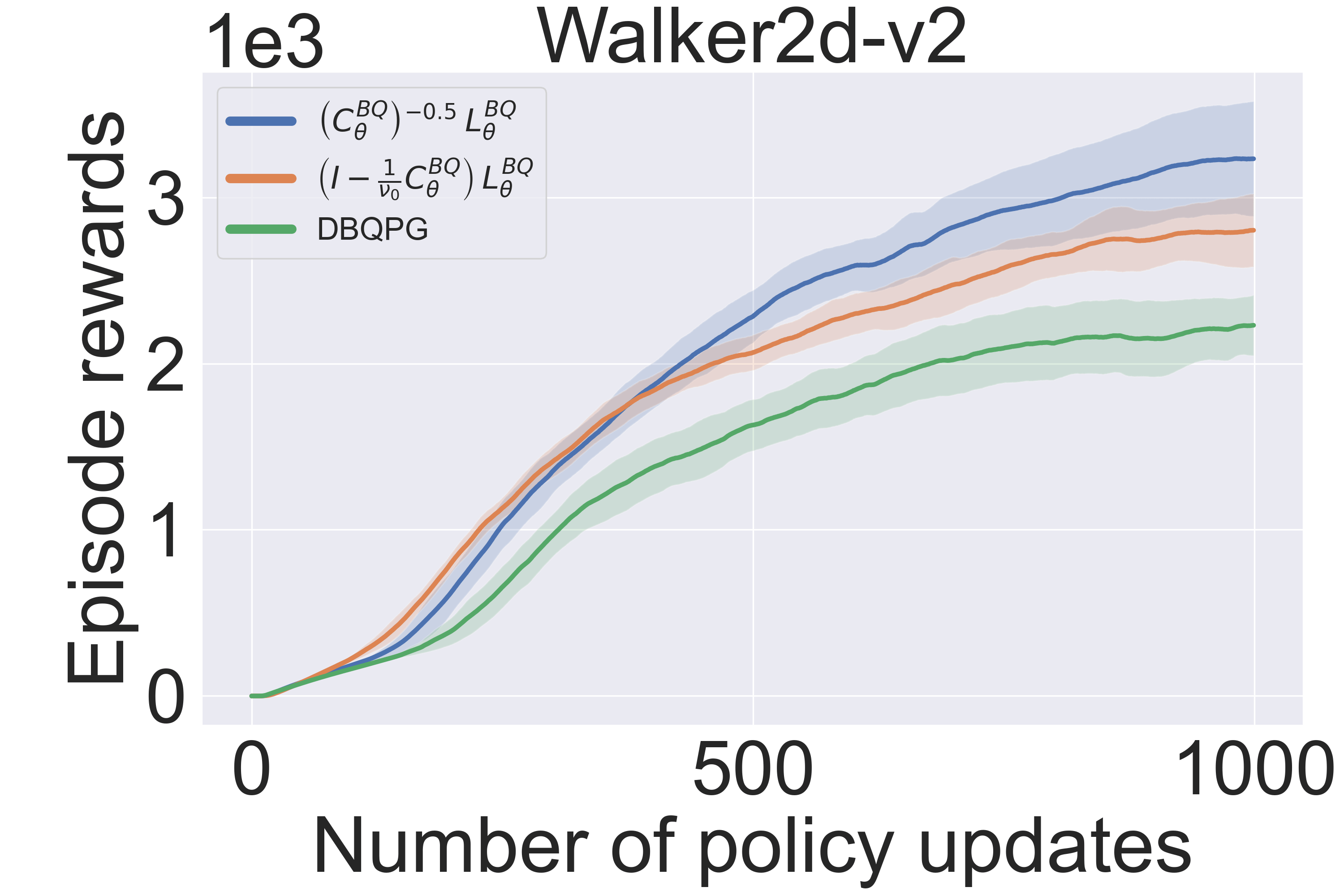}
    \end{subfigure}
    \begin{subfigure}
  	\centering
  	\includegraphics[scale=0.07]{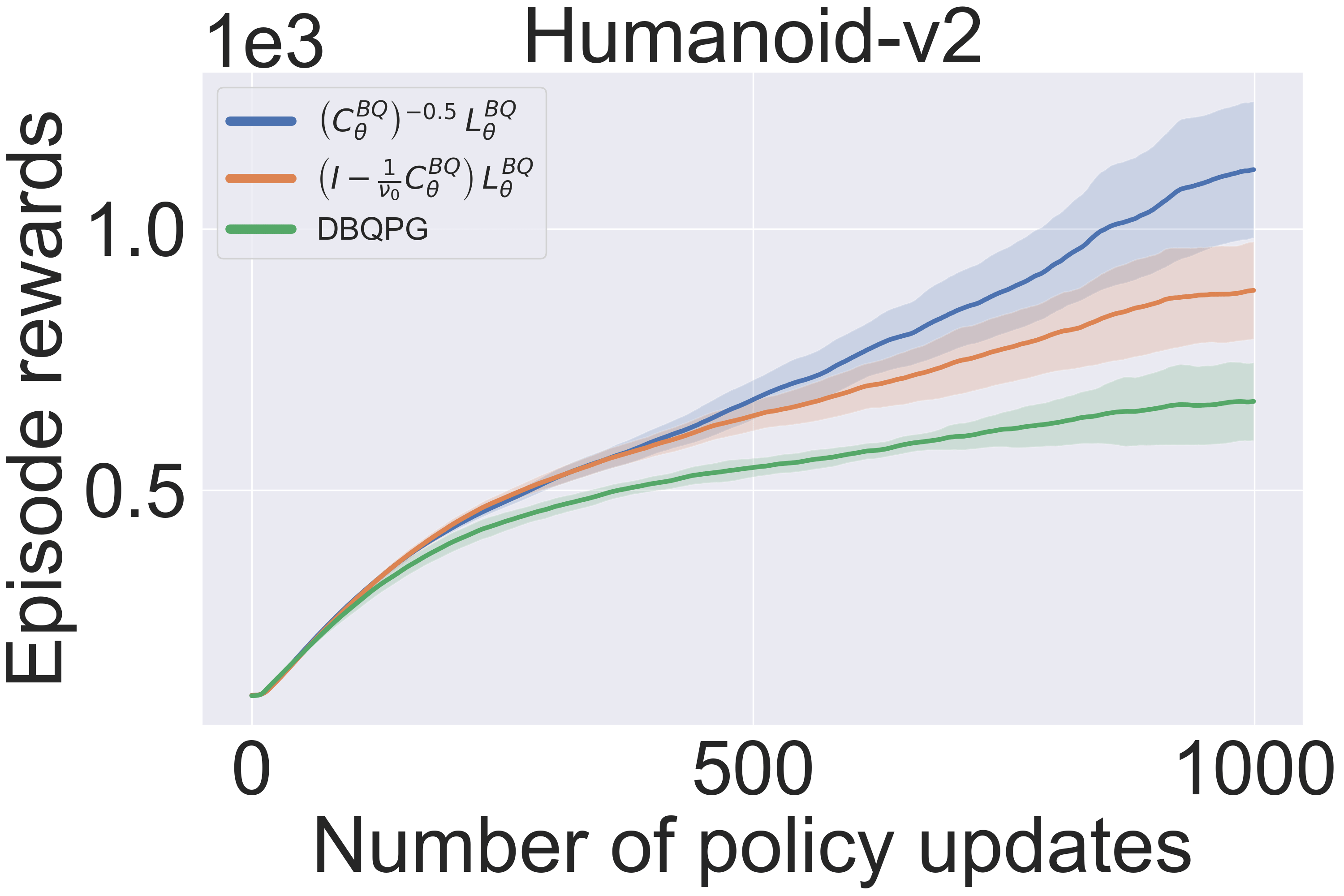}
    \end{subfigure}
    \caption{\UAPG versus the uncertainty-based policy update proposed in \citet{BAC_2}.}
  \label{fig:appendix_uncertainty_expt}
\end{figure*}
Further, \citet{BAC_2} introduces an extension of \BAC that uses the gradient covariance $\mC_{\vtheta}^{BQ}$ for adjusting the learning rate of policy parameters:
\begin{equation}
    \mL_{\vtheta}^{UBAC} = \left(\mI - \frac{1}{\nu_0}\mC_{\vtheta}^{BQ}\right)\mL_{\vtheta}^{BQ},
\end{equation}
where $\nu_0$ is the largest singular value\footnote{The original uncertainty-based update was proposed for the Bayesian policy gradient method \citep{BPG}, which has been adapted for \textit{our} BAC \citep{BAC_1} implementation and compared against \DBQPG.} of $\mC_{\vtheta}^{BQ}$. This gradient preconditioning is one of the many possible solutions to lower the step-size of gradient components with high uncertainties, with another solution being \UAPG. However, \UAPG update satisfies a more stricter condition of providing policy updates with uniform uncertainty.
Since, $\mL_{\vtheta}^{UBAC}$ does not provide \PG estimates with uniform uncertainty, it is more vulnerable to bad policy updates than \UAPG. This is also demonstrated in practice, where $\mL_{\vtheta}^{UAPG}$ outperforms $\mL_{\vtheta}^{UBAC}$, while both the uncertainty-based methods outperform \DBQPG (see Fig.~\ref{fig:appendix_uncertainty_expt}). Moreover, $\mL_{\vtheta}^{UBAC}$ update would not work for \NPG and \TRPO algorithms as their covariance is the inverse of an ill-conditioned matrix.

\section{Implementation Details}
\label{appendix_implementation_details}

We follow the architecture from Fig.~\ref{fig:DBQPG_UAPG_diagram} (left; main paper), where the policy $\pi_\vtheta$ comprises of a deep neural network that takes the observed state as input and estimates a normal distribution in the output action-space, characterized by the mean and standard deviation. We use the standard policy network architecture \citep{TRPO,Duan_2016} that comprises of a $2$-layered fully-connected MLP with $64$ hidden units and a tanh non-linearity in each layer. Our state feature extractor $\phi(s)$ is another MLP with hidden dimensions $64$, $48$, and $10$, and tanh non-linearity. The explicit critic network is simply a linear layer with $1$ output unit, on top of the $10$-d output of $\phi(s)$. The explicit critic models the state-value function $V_{\pi_\vtheta}$ and its parameters are optimized for TD error. The $n$ dimensional vector $\Qmc$ is formed from generalized advantage estimates \citep{GAE}, which are computed using \MC estimates of return and the state-value predictions offered by the explicit critic network.

The implicit \GP critic implicitly models the generalized advantage function using the samples $\Qmc$, a fisher kernel (no additional hyperparameters; derived directly from the policy parameters) and a deep RBF state kernel (lengthscale + $\phi(s)$ parameters). For structured kernel interpolation \citep{kissgp}, we use a grid size of $128$ and impose an additive structure (i.e. the overall kernel is the sum of individual RBF kernels along each input dimension) on the deep RBF kernel. Additive structure allows us to take advantage of the Toeplitz method \citep{toeplitz_turner:2010} for fast and scalable MVM computation. The \GP's noise variance $\sigma^2$ is set to $10^{-4}$. In all the experiments of \BQ-based methods, we fixed the hyperparameters  $c_1 = 1$ and $c_2= 5 \times 10^{-5}$ (tuned values). The parameters of deep RBF kernel are optimized for \GP-MLL objective (Eq.~\ref{eqn:GP_nll}). Since $\phi(s)$ is shared between the implicit critic (deep RBF kernel) and explicit critic, its parameters are updated for both \GP-MLL and TD error objectives. Our implementation is based on \textit{GPyTorch} \citep{gpytorch}, a software package for scalable \GP inference with GPU acceleration support.

For Vanilla \PG and \NPG experiments, we used the ADAM optimizer \citep{adam} to update the policy network. In the \UAPG experiments, increasing the SVD rank $\delta$ pushes the practical \UAPG estimate closer to the ideal \UAPG estimate, while, also increases the GPU memory requirement. We balance this trade-off for each environment by choosing a $\delta$ that closely approximates the initial spectrum of gradient uncertainty $\mC^{BQ}_{\vtheta}$ and also has a favorable GPU memory consumption. Lastly, we set $\eps=3$ for \UAPG's natural gradient update. Our implementation of \DBQPG and \UAPG methods is made publicly available at {\color{blue}{\url{https://github.com/Akella17/Deep-Bayesian-Quadrature-Policy-Optimization}}}.
\begin{table*}[ht]
\centering
\begin{adjustbox}{max width=0.6\linewidth}
\begin{tabular}{c|c|c}
\multicolumn{2}{c|}{\textit{\textbf{Parameter}}} & \textit{\textbf{Value}} \\ \hline
\multicolumn{2}{c|}{Batch size} & $15000$ \\
\multicolumn{2}{c|}{Discount factor $\gamma$} & $0.995$ \\
\multicolumn{2}{c|}{GAE coefficient $\tau$} & $0.97$ \\
\multicolumn{2}{c|}{Trust region constraint / step size} & $0.01$ \\ \hline
\multirow{3}{*}{\textit{Conjugate Gradient}} & Max. CG iterations & $50$ \\
 & Residue (i.e., CG Threshold) & $10^{-10}$ \\
 & Damping (stability) factor & $0.1$
\end{tabular}
\end{adjustbox}
    \caption{Common hyperparameter setting across all the experiments.}
\end{table*}

\end{document}